\documentclass[letterpaper]{article} 

\usepackage{times}  
\usepackage{helvet}  
\usepackage{courier}  
\usepackage{url}  
\usepackage{graphicx}  
\usepackage[table,xcdraw,dvipsnames,svgnames,x11names]{xcolor}
\usepackage[utf8]{inputenc} 
\usepackage[T1]{fontenc}    
\usepackage{hyperref}
\usepackage{booktabs}       
\usepackage{amsfonts}       
\usepackage{nicefrac}       
\usepackage{microtype}      
\usepackage{tikz}
\usetikzlibrary{shapes.misc, positioning}
\usepackage{amsthm}
\usepackage{thmtools,thm-restate}
\usepackage{parskip}
\usepackage{authblk}
\usepackage{amsmath}  
\usepackage[numbers,sort]{natbib}

\usepackage{floatrow}
\newfloatcommand{capbtabbox}{table}[][\FBwidth]

\usepackage{macros}
\usepackage{basic}

\usepackage{microtype}
\usepackage{enumitem}
\usepackage{caption}
\usepackage{multirow}
\usepackage{amssymb}
\usepackage{subcaption}  
\usepackage{float}       
\usepackage{color}     
\usepackage[most]{tcolorbox}
\usepackage{array} 
\usepackage{wrapfig}
\usepackage{xspace}  
\usepackage{tcolorbox}
\usepackage{stfloats}  
\usepackage{cleveref}

\title{COSMOS: Predictable and Cost-Effective Adaptation of LLMs}

 \author[$\dagger$]{Jiayu Wang}
  \author[$\dagger$]{Aws Albarghouthi}
 \author[$\dagger$]{Frederic~Sala}

 \affil[$\dagger$]{University of Wisconsin-Madison}
 \affil[ ]{\footnotesize{\texttt{\{milawang,aws,fredsala\}@cs.wisc.edu}}}

\begin{document}

\maketitle

\begin{abstract}
Large language models (LLMs) achieve remarkable performance across numerous tasks by using a diverse array of adaptation strategies. 
However, optimally selecting a model and adaptation strategy under resource constraints is challenging and often requires extensive experimentation. 
We investigate whether it is possible to accurately predict both performance and cost  without expensive trials. 
We formalize the strategy selection problem for LLMs and introduce \methodabb, a unified prediction framework that efficiently estimates adaptation outcomes at minimal cost. 
We instantiate and study the capability of our framework via a pair of powerful predictors: embedding-augmented lightweight proxy models to predict fine-tuning performance, and low-sample scaling laws to forecast retrieval-augmented in-context learning. 
Extensive evaluation across eight representative benchmarks demonstrates that \methodabb achieves high prediction accuracy while reducing computational costs by 92.72\% on average, and up to 98.71\% in resource-intensive scenarios. Our results show that efficient prediction of adaptation outcomes is not only feasible but can substantially reduce the computational overhead of LLM deployment while maintaining performance standards.

\end{abstract}

\section{Introduction}
Large language models (LLMs) have scaled dramatically in both capability and availability, with over 1.6 million models now shared on Hugging Face 
(as of Apr. 2025). 
Each model offers distinct performance characteristics and computational demands. The emergence of diverse adaptation techniques has further expanded the space of possible deployment configurations. 
This raises a \emph{key question}: how can we systematically identify an optimal choice of model and adaptation strategy in a cost-effective way?

\begin{figure*}
    \centering
    \includegraphics[width=0.95\textwidth]{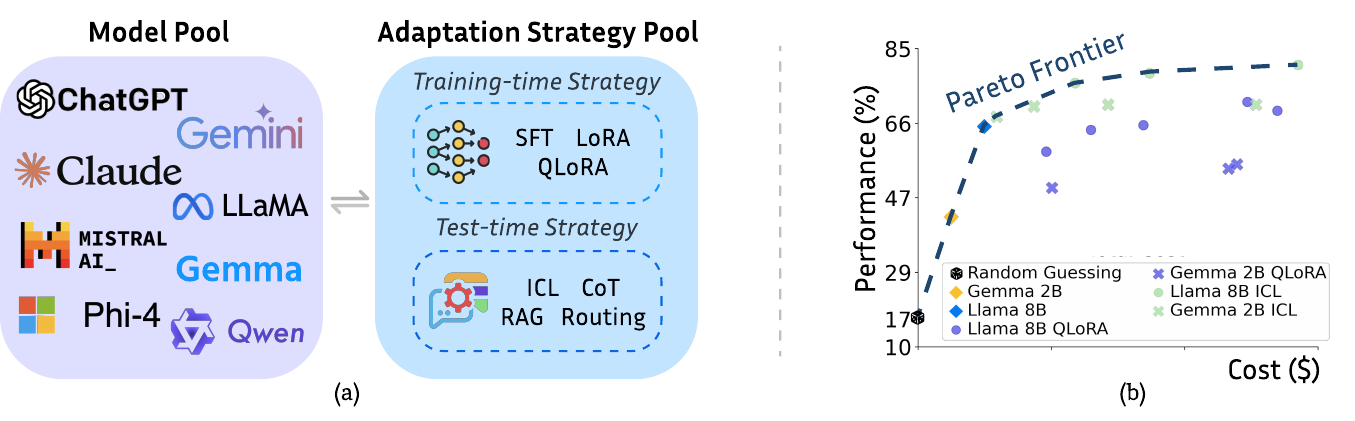}
   \vspace{-1em}
    \caption{Overview of the \problemabb problem for LLMs and performance--cost tradeoff. \textbf{(a)} Given a downstream task, practitioners select from a pool of foundation models and adaptation strategies. 
    \textbf{(b)} Each model--strategy combination results in different performance and cost.
    The challenge lies in choosing optimal combinations that balance performance and cost.}
    \label{fig:prob-def-choose-model-strategy}
\end{figure*}

To answer this question, we introduce and formalize the \problemabb problem for LLMs,
aiming to identify the optimal combination of model and adaptation strategies that balance the performance and cost. 
Figure~\ref{fig:prob-def-choose-model-strategy} illustrates this problem, where practitioners can select from a pool of foundation models and adaptation strategies, each with their own configuration space, to solve tasks across various domains. 
The challenge lies in navigating this vast search space to find the best model--strategy combination without exhaustively evaluating all possibilities.

Intuitively, one way to tackle this problem is to \emph{predict} how much each adaptation buys us; this must be done in a cost-efficient manner. Based on this idea, we propose \methodfull (\methodabb), a unified framework for efficiently predicting the effectiveness of adaptation strategies. It obviates the need to run resource-intensive experiments. 
We demonstrate \methodabb's versatility by instantiating it for two prominent adaptation strategies: 1) QLoRA fine-tuning, where we predict adaptation gains with an embedding-augmented lightweight proxy model, and 2) retrieval-augmented in-context learning, predicting performance using observed scaling laws. 

Through extensive experiments across eight representative benchmarks spanning both general and specialized tasks, we demonstrate that \methodabb achieves excellent prediction accuracy (mean absolute error of 1.09\%) while reducing computational costs by an average of 92.72\% (up to 98.71\%) compared to exhaustive experimentation. 
This means that \textbf{efficient prediction} of adaptation outcomes is not only feasible, but can empower practitioners to navigate the space of model--strategy combinations and make informed decisions that optimize both performance and cost. 

Our main contributions are summarized as follows:
\begin{itemize}
    \item We introduce and formalize the optimal strategy selection problem for LLMs that jointly considers the model selection and adaptation strategies with performance--cost tradeoffs.
    \item To address this problem, we propose \methodabb, a novel framework that enables accurate and data-efficient estimation of both performance and cost across training-time and test-time adaptation strategies.
    \item Through extensive evaluation on eight diverse benchmarks, we demonstrate that \methodabb achieves superior prediction accuracy (MAE of 1.09\%) while reducing computational costs by an average of 92.72\% (up to 98.71\%) compared to baselines.
\end{itemize}
\section{Related Works}
\noindent \textbf{Adaptation strategies for LLMs.}
Strategies for adapting pre-trained LLMs for downstream tasks broadly fall into two categories: training-time and test-time adaptation. 
Popular training-time strategies include supervised fine-tuning~\citep{raffel2020exploring,wei2021finetuned} and parameter-efficient variants such as
LoRA~\citep{hu2021lora} and QLoRA~\citep{dettmers2024qlora}. 
Test-time methods such as in-context learning~\citep{brown2020language}, advanced search and prompting ~\citep{wei2022chain, yao2023tree}, and decoding algorithms~\citep{wang2024chain,gad2024} obtained significant attention in recent years~\citep{welleck2024decoding}. 
While these individual strategies are effective, efficiently selecting and configuring them jointly remains underexplored. 

\noindent \textbf{Model routing.} 
Routing (\ie, sending easier queries to cheaper models and reserving powerful models for harder queries) has 
been explored via a variety of approaches~\citep{chen2023frugalgpt,yue2023large,shnitzer2023large,vsakota2024fly,chen2024data}. 
While routers can balance performance and cost in model selection, they usually operate with a fixed model pool and do not consider a broad space of adaptation strategies. 
In contrast, our work \emph{goes beyond pure model selection and routing}.

\noindent \textbf{Training and test-time scaling laws.}
Training-time scaling laws~\citep{kaplan2020scaling,hoffmann2022empirical} explore the relationship between model size, training compute, and performance. Recent work on test-time scaling~\cite{snell2024scaling} has systematically analyzed how performance gains from inference strategies such as best-of-N sampling~\citep{cobbe2021training,lightman2023let} and beam-search~\citep{yao2023tree,feng2023alphazero} scale with task difficulty. 
Meanwhile, \citet{ruan2024observational} has revealed predictable relationships between LLM performance and low-level skills.
However, existing approaches are limited to coarse-grained, task-agnostic predictions that frequently fail to consider the interplay between training-time and test-time strategy adaptations. 
Moreover, they rarely account for the full spectrum of costs—including prediction, adaptation, and evaluation costs.
In contrast, our work provides accurate predictions for both \emph{performance and overall cost across multiple dimensions}. 

\noindent \textbf{AutoML, hyperparameter optimization, and NAS.}
Model selection, training hyperparameter optimization (HPO), and the development of new architectures optimized for each task are the traditional domains of AutoML and Neural Architecture Search (NAS). %
Recent work focuses on lifting these approaches to modern LLM settings \citep{saad2024archon, Roberts24Manticore}. 
Most of the techniques within these areas are either narrow (\eg, HPO \citep{yu2020hyper} focuses on one specific type of adaptation) or extremely expensive (\eg, \citet{Roberts21XDOps} and \citet{Shen23Orca}).
\textit{The unified approach we propose is simultaneously general and cost-efficient}. Our framework and techniques are compatible with tools from AutoML and NAS. 
\section{The Strategy Selection Problem}\label{sec3:problem-statement}
In this section, we formalize the \problemabb problem (Section~\ref{sec:general-prob-formulation}) and introduce \methodabb, a unified framework for predicting performance and total cost across prediction, adaptation, and evaluation processes (Section~\ref{sec:general-method}). 
We further analyze suitable adaptation strategies (Section~\ref{sec3.1:adaptation-strategy}), and develop a cost analysis methodology to evaluate the computational efficiency of various adaptation strategies and predictors within COSMOS (Section~\ref{sec3.2:cost-calculation}).

\subsection{Problem Formulation}\label{sec:general-prob-formulation}
Our goal is to systematically identify good choices of model and adaptation strategy in a cost-effective way. 
To formalize this problem, we define a \textit{\indicatorname} ($M_D$) whose purpose is to determine the optimal combination of model, adaptation strategy, and configuration for a downstream task \( D \in \mathcal{D}\) that balances performance and cost-efficiency.
Formally, we are given the following:
\begin{itemize}
    \item A \textit{model pool} $\mathcal{F} = \{ f_1, f_2, \dots, f_K \}$, where each model $f_k \in \mathcal{F}: \mathcal{X} \rightarrow \mathcal{Y}$ maps an input query $\textbf{x} \in \mathcal{X}$ to a model answer $\hat{\textbf{y}} \in \mathcal{Y}$,
    \item An \textit{adaptation strategy pool} $\mathcal{T} = \{ T_1, T_2, \dots, T_{J} \}$,
    \item A configuration space $\Omega$ where each configuration $\omega \in \Omega$ specifies parameters for applying a strategy,
    \item A performance metric $\pi$ and cost function $c$.
\end{itemize}

Let $T_j^\omega(f_k)$ represent a model after applying strategy $T_j$ with configuration $\omega$, resulting in a performance-cost pair $(\pi(T_j^\omega(f_k)), c(T_j^\omega(f_k)))$. The \indicatorabb selects the optimal combination by solving:
\begin{equation}
    M_D(\mathcal{F}, \mathcal{T}, \Omega) = \argmax_{f_k \in \mathcal{F}, T_j \in \mathcal{T}, \omega \in \Omega} s(\pi(T_j^\omega(f_k)), c(T_j^\omega(f_k))),
\end{equation}
where $s: \mathbb{R} \times \mathbb{R}_+ \rightarrow \mathbb{R}$ is a score function that captures the trade-off between performance and cost based on the practitioner's preferences.

\vspace{0.5em}
\begin{example}\label{ex:cosmos}
Consider a sentiment analysis task where we have one 7B parameter model $f_1$ and two adaptation strategies: $T_{\text{probe}}$ (linear probing) and $T_{\text{tune}}$ (full fine-tuning). Each strategy's configuration space includes learning rate, number of epochs, and training data size, each with 10 possible values. 
This results in 2000 total combinations (1 model × 2 strategy x 1000 configurations per strategy). 
\end{example}
\noindent \textbf{Computational cost of \problemabb problem.}
One approach is to exhaustively try all combinations. However, this quickly becomes prohibitively expensive.
This naturally raises the question: \textit{can we solve the \problemabb problem cost-efficiently?} We answer this affirmatively by introducing \methodabb, a unified framework to solve the \problemabb problem via predicting adaptation gains, resulting in a cheaper way to explore the search space (Section~\ref{sec4:method}).

\subsection{Predictive Adaptation Framework}\label{sec:general-method}
Given the computational challenges of the \problemabb problem, one intuitive approach is to employ a cheap predictor with a small amount of data rather than conducting expensive full-scale experiments.
We formalize this predictive approach through COSMOS (\methodfull), a unified framework that systematically predicts adaptation gains across diverse strategies while significantly reducing computational overhead.

For an adaptation strategy $T_j \in \mathcal{T}$ applied to a model $f_k \in \mathcal{F}$ with configuration $\omega \in \Omega$, our framework aims to: 
1) predict performance:
$    \hat{\pi}(T_j^\omega(f_k)) = P_{j,k}(\omega) \approx \pi(T_j^\omega(f_k))$,
and 2) predict associated costs: 
$\hat{c}(T_j^\omega(f_k)) = C_{j,k}(\omega) \approx c(T_j^\omega(f_k)) 
$. 
Here, $P_{j,k}$ and $C_{j,k}$ are strategy-specific predictors that map configurations to expected performance and cost respectively.
These predictors can take various forms depending on the nature of the adaptation strategy $T_j$.
Examples of predictors include a lightweight proxy model and calibration on a small validation set to estimate adaptation outcomes, \eg, linear probing on frozen embeddings, which is suitable for parameter-updating strategies (\eg, fine-tuning);
leveraging observed scaling law to extrapolate performance with a small amount of early sparse measurements.

Ideal performance and cost predictors should satisfy two properties:
1) \textit{Cost-efficiency}: The prediction cost must be significantly lower than the actual adaptation
$c_{\text{predict}}(P_{j,k}, C_{j,k}) \ll c_{\text{adapt}}(T_j^\omega, f_k)$ 
and 2) \textit{Strategy-specificity}:
While not strictly required, predictors tailored to strategy-specific characteristics typically yield superior prediction accuracy and cost efficiency in practice
$P_{j,k} \in \mathcal{P}_j, C_{j,k} \in \mathcal{C}_j$, 
where $\mathcal{P}_j$ and $\mathcal{C}_j$ are the sets of valid performance and cost predictors respectively for the strategy $T_j$.

\noindent \textbf{Cost analysis.} The total cost includes:
1) Prediction cost: 
$c_{\text{predict}}(P_{j,k}, C_{j,k})$
, which includes strategy-specific prediction overhead and calibration cost using validation data if necessary;
and 2) Selected strategy cost: 
$c(T_{\hat{j}}^{\hat{\omega}}, f_{\hat{k}})$
, which includes the cost of applying the chosen strategy and final evaluation cost, detailed in Section~\ref{sec3.2:cost-calculation}.
The framework is efficient when:
\begin{equation}
\sum_{j,k} 
c_{\text{predict}}(P_{j,k}, C_{j,k})
+ c(T_{\hat{j}}^{\hat{\omega}},f_{\hat{k}}) \ll \sum_{j,k,\omega} c(T_j^\omega, f_k).
\end{equation}

\paragraph{Framework instantiation.}
To apply this framework to a specific adaptation strategy $T_j$, one needs to: 1) Choose an appropriate predictor type based on strategy characteristics; 
2) Design the predictor architecture or model;
3) Define the prediction cost calculation. 
\vspace{0.5em}
\begin{example}
\label{ex:framework}
Continuing with the sentiment analysis task from Ex. ~\ref{ex:cosmos}, For $T_{\text{tune}}$ (full fine-tuning), the predictor $P_{\text{tune},f_1}$ could be a lightweight linear model trained on frozen embeddings and calibrate the performance from a small validation set. 
Assume the total cost  of prediction including the training cost of the proxy model and validation cost arising from performance calibration, for total 1000 configs prediction, the prediction costs \$5, however, the actual total cost of adaption can be \$500. 
This demonstrates the efficiency property as $c_{\text{predict}}$ (\$5) $\ll$ $c_{\text{adapt}}$ (\$500).
\end{example}

In Section~\ref{sec4:method}, we instantiate our framework for two diverse adaptation strategies: 1) \qlora fine-tuning: predict performance via an embedding-augmented linear proxy model; and 2) retrieval-augmented \icl: predict performance via observed-scaling law. We consider the full spectrum of cost based on both computing-based and token-based methods.

\subsection{Adaptation Strategies}\label{sec3.1:adaptation-strategy}
Adaptation strategies include, for example: 

\noindent \textbf{Training-time adaptation strategies.}
A Training-time adaptation strategy modifies the parameters of a language model$\AModelParam$, where ${\theta \in \Theta}$. Examples include full fine-tuning to parameter-efficient methods like LoRA and QLoRA. Two techniques in Ex.~\ref{ex:cosmos} are both training-time methods.
Formally, a training time adaptation function \( T^{\text{tr}}: \AModelParam \rightarrow \AModelParamT \), transforms a base model into a task-specialized model.

\noindent \textbf{Test-time adaptation strategies.}
Test-time (or inference time) adaptation strategies complement training-time approaches by modifying the input and/or output processing rather than the model parameters such as prompt tuning~\citep{lester-etal-2021-power}, CoT, and ICL.
A test-time adaptation function $T^{\text{inf}}: 
\mathcal{X} \times \mathcal{Y} \rightarrow \mathcal{X'} \times \mathcal{Y'}
$
transforms the input-output space to enhance model performance without parameter updates.

\noindent \textbf{Hybrid adaptation strategies.}
Recent research demonstrates growing interest in hybrid adaptation strategies that fall into the intersection of both training-time and test-time adaptations~\citep{soylu2024fine}.
Formally, hybrid approaches can be represented as composite adaptation functions where parameter transformations and input-output space modifications work in concert: $T^{\text{hybrid}} = T^{\text{tr}} \circ T^{\text{inf}}$.

\noindent \textbf{Model routing as a special case.}
Model routing (\ie, directing different queries to different models) can be viewed as a constrained instance of the \problemabb problem where: 1) The adaptation strategy pool contains a single strategy with fixed configuration; 
2) The router operates at the query level rather than the task level.

\subsection{Cost Analysis Framework}\label{sec3.2:cost-calculation}
The effectiveness of adaptation strategies must be balanced against their computational costs. To enable practical decision-making, we model the cost of adaptation, evaluation, and prediction phases. 

\noindent \textbf{Total cost.}
Given a specific task $D$, the total cost for any adaptation strategy $\AStrategy$ configured by $\omega$, applied to model $\AModel$ comprises two components:
\begin{equation}
    c(\AStrategy, \AModel) = c_{\text{adapt}}(\AStrategy, \AModel) + c_{\text{eval}}(\AStrategy(\AModel), D)
\end{equation}
where $c_{\text{adapt}}$ represents the adaptation cost (\ie, cost of applying the adaptation strategy) and $c_{\text{eval}}$ (\ie, cost of evaluating adapted model performance).

\noindent \textbf{Strategy-specific adaptation cost.}
Different strategies incur adaptation costs $c_{\text{adapt}}$ through different mechanisms: In test-time strategies, the cost includes only inference expenses, determined by both input and output token costs for a given model and scales with the number of inference passes. In training-time strategies, the cost can be calculated through either:
1) computing-based method: (GPU/TPU hourly rate) × usage duration, or
2) token-based method: per-token price × training data size × epochs.

\noindent \textbf{Prediction cost.} 
Prediction cost typically arises from training and calibrating performance predictor $P_{j,k}$ and cost predictor $C_{j,k}$. This includes:
1) lightweight proxy model training for performance prediction, 
2) calibrating performance on validation data if necessary, and
3) strategy-specific costs (\eg, early points generation to extrapolate according to scaling law).
Cost prediction typically leverages the same validation runs or dataset information used for performance prediction, thus incurring minimal additional overhead.
\begin{equation}
    c_{\text{predict}}(P_{j,k}, C_{j,k}) = c_{\text{proxy}} + c_{\text{overhead}}(T_j^\omega, f_k) + c_{\text{val}}(D_{\text{val}}).
\end{equation}
This unified cost framework enables direct comparison between diverse adaptation approaches while accounting for their distinct resource requirements. It provides a foundation for the cost-aware strategy selection developed in our prediction framework (Section~\ref{sec:general-method}).
\section{\methodabb: Solving The  Strategy Selection Problem}\label{sec4:method}
While there may exist some predictive approaches that can be applied universally across adaptation strategies, tailored prediction methods often yield superior results in both accuracy and efficiency. 
In this section, we demonstrate how \methodabb can be instantiated with strategy-specific predictors by presenting two complementary examples.

\subsection{Instantiation Setup}
We study two popular complementary adaptation strategies, each paired with a tailored prediction approach.
The first is \qlora fine-tuning $T^{\text{tr}}_{\text{\qlora}}$; we predict its performance using an embedding-augmented linear model. 
The second is retrieval-based in-context learning (\icl) $T^{\text{inf}}_{\text{\icl}}$; we predict its performance using scaling laws. 

Each strategy operates in a distinct configuration space that affects its resource requirements and potential gains:
For QLoRA, $\Omega_{\text{QLoRA}} = [0,1] \times \mathbb{N}^+$, representing the continuous spectrum of data proportion and discrete training iterations. The adaptation function then maps a model to its fine-tuned version:
$T^{\text{tr}}_{\text{QLoRA}}: f_{\eta, \phi} \times ([0,1] \times \mathbb{N}^+) \rightarrow f_{\eta', \phi}$.   
For ICL, we control the number of shots $n$ and sequence length
$
\Omega_{\text{ICL}} = \{n \in \mathbb{N}^+ : C(n) \leq L_{\text{max}}\}
$
where $
C(n) = L_{\text{query}} + \sum_{i=1}^n L_{\text{demo}_i} \leq L_{\text{max}}
$ represents the total sequence length. 
The \icl adaptation function modifies the input space:
$T^{\text{inf}}_{\text{ICL}}: \mathcal{X} \times \Omega_{\text{ICL}} \rightarrow \mathcal{X'}$. 

\subsection{Predicting Fine-tuning Gain}\label{sec:finetuning-gain-predictor}
For \qlora fine-tuning, we develop an embedding-based prediction method that works as follows. 
We use a language model \( f_{\theta} \) in the model pool that has two key components: (1) 
a function \( g_{\eta}: \mathbb{R}^{L \times d} \rightarrow \mathbb{R}^{L \times e} \), parameterized by \( \eta \) that maps a sequence \( \mathbf{x} = (x_1, \dots, x_L) \) to a representation, where \(d\) is the embedding dimension, and \(e\) is the hidden dimension. It also has (2) a projection head \( h_{\phi}: \mathbb{R}^{e} \rightarrow \mathbb{R}^{|\Sigma|} \), parameterized by \( \phi \).

Inspired by LLM2Vec~\citep{behnamghader2024llm2vec}, we first transform the traditional causal language model into a bidirectional embedding model. For input  \(\mathbf{x} = (x_1, \dots, x_L) \),
we compute:
$z_t^{\text{bi}} = g_{\eta}^{\text{bi}}(x_1, x_2, \dots, x_L)$
where \(g_{\eta}^{\text{bi}}(\mathbf{x}) \in \mathbb{R}^{T \times e} \) produces contextualized representations.
To obtain a fixed-dimensional representation for the entire sequence, we use mean pooling.
This sequence embedding \(e_{\eta}(\mathbf{x})\) will serve as the input to the projector for fine-tuning performance estimator. 
Given the fine-tuning training data \( D^\text{FT}_{\text{train}}=\{(\mathbf{x}_i, \mathbf{y}_i)\}_{i=1}^N \),  we learn a lightweight \textit{task-specific projector} \(l_{\phi''}: \mathbb{R}^e \rightarrow \mathcal{Y}\) that maps sequence embeddings to the target space:
$
\mathbf{\hat{y}} = l_{\phi''}(e_{\eta}(\mathbf{x}))
$
where $l_{\phi''}$ is a linear layer. 
This minimal architecture ensures computational efficiency while leveraging the rich representations from the frozen model embeddings.

Finally, to bridge the gap between projector predictions and actual fine-tuning performance, we use a calibration mechanism:
$
    \hat{\pi}(T^{\text{tr}}_{\text{\qlora}}(f_{\theta, \phi})) = a\pi_{\phi''} + b
    $
where $\pi_{\phi''}$ is the projector performance and \(a, b \in \mathbb{R}\) are parameters learned from a small validation set (\eg, 10\% of full training data).
This step ensures our predictions align with actual performance while maintaining computational efficiency.

Next, we describe costs for fine-tuning (and our predictor). Our primary method calculates costs based on computing time, which enables direct comparison with prediction costs. 
The fine-tuning cost \(c^{\text{FT}}\) depends on the number of tokens in the training set \( N^{\text{FT}}_{\text{train}}\), the number of epochs \( E\), batch size \( B\), gradient accumulation step $G$, and the type of computational resources used. 
This approach factors in the total number of training steps, processing time per gradient update step $t_{\text{step}}$, and the hourly cost of compute resources $\gamma_{\text{compute}}$. 

We use token packing to optimize token usage and consider memory utilization $\psi_{\text{peak}}$ (the ratio of peak training memory occupation to total available memory), enabling fair comparison between prediction costs and full adaptation experiments.
The total cost of fine-tuning is modeled as:
\begin{equation} 
 c^{\text{FT}} = E \times \frac{\text{pack}(N^{\text{FT}}_{\text{train}}, L_{\max})}{B \times G} \times t_{\text{step}}
     \nonumber \ \times \gamma_{\text{compute}} \times N_{\text{compute}} \times \psi_{\text{peak}}
    + c_{\text{eval}}
\end{equation}
where $\text{pack}(\cdot,\cdot)$ computes the number of effective sequences after optimal packing of training tokens $N^{\text{FT}}_{\text{train}}$ sequences subject to max sequence length $L_{\max}$ constraint. In terms of prediction, we derive the peak memory usage from the small validation set during performance calibration.

\subsection{Predicting Retrieval-Augmented \icl Gain}\label{sec:prompting-gain-predictor}
For retrieval-based \icl, our key insight is that retrieval-based ICL performance typically follows an exponential saturation curve that early measurements can characterize.

\noindent \textbf{Performance prediction.}
Given sparse performance measurements $\{(d_i, \pi_i)\}_{i=1}^m$, we fit an exponential saturation model:
\begin{equation}
 \hat{\pi}(T^{\text{inf}}_{\text{\icl}}(f_{\eta, \phi})) = \alpha(1 - e^{-\beta d}) + \pi_0
   \label{eq:icl-perf}
\end{equation}
where $d$ is the shot count, and $(\alpha, \beta, \pi_0)$ capture the saturation behavior. 
This allows us to predict performance at any demonstration count while requiring only a few initial measurements---as few as two points.

\noindent \textbf{Cost prediction.}
For \icl, given each query $\mathbf{x}$, we can estimate the cost:
$c^{\text{ICL}}(d, \mathbf{x}) = c_{\text{token}}(\mathbb{E}[L_{\text{in}}] + \mathbb{E}[L_{\text{out}}]) \times d + c_{\text{token}}( \mathbf{x} + \mathbb{E}[L_{\text{out}}]) + c_{\text{eval}}$
where $\mathbb{E}[L_{\text{in}}]$ and $\mathbb{E}[L_{\text{out}}]$ are expected input/output lengths. 
This can be estimated from the demonstration knowledge base, providing cost estimation for arbitrary shot counts.

The above methods equipped \methodabb practical approach to predicting adaptation gains, enabling efficient exploration of the adaptation space without exhaustive computation. Our empirical results (Section~\ref{sec:experiments}) demonstrate that these predictions closely align with actual performance while reducing computational overhead by orders of magnitude.
\section{Experiments}\label{sec:experiments}
We conduct extensive experiments to validate \methodabb. 
\begin{tcolorbox}[colback=calloutColor,
top=3pt,
bottom=3pt,
left=3pt,
right=3pt,
title=Key Takeaway at a Glance,
colbacktitle=black,
coltitle=white
]
Optimizing training-time and inference-time strategies \emph{jointly} can be more cost-effective than scaling them separately.
Our COSMOS framework can help guide strategy selection by making accurate predictions on both performance and cost efficiently, enabling practitioners to make choices \emph{flexibly} based on their specific performance-cost preferences.
\end{tcolorbox}

Our evaluation aims to answer the following key questions:
\begin{itemize}
    \item \textbf{Prediction Accuracy with Cost Efficiency} (\Secref{sec5.1-overall-perf}): Can \methodabb effectively predict the optimal adaptation strategy? We show that our method achieves 92.72\% cost reduction while maintaining high prediction fidelity (1.09\% MAE) across 8 tasks, 55 strategy combinations, and spanning multiple cost regimes.
    \item \textbf{Robust Strategy-Specific Prediction Capabilities} (\Secref{sec5.2:detailed-pred-analysis}): How well does \methodabb predict the performance and cost of each combination? We demonstrate strong prediction capabilities for both performance gains and computational costs across multiple adaptation strategies and for general and specific tasks.
    \item \textbf{Cost-effective Training-time and Test-time Scaling Synergies and Tradeoffs} (\Secref{sec5.3:scaling-behavior}): What are the optimal performance-cost tradeoffs under different computational budgets when comparing training-time vs. test-time adaptation? We provide critical insights into the efficiency of these two scaling approaches.
    \item \textbf{Strategy Space Expansion Benefits} (\Secref{sec5.4:aug-routing}): How does broadening the adaptation strategy pool beyond simple model selection enhance the performance-cost tradeoffs in routing? 
    We show augmenting model routing with our approach  advances the Pareto frontier.
    \item \textbf{Potential Implications}~(\Secref{sec5.5:implications}): How will \methodabb benefit real industrial deployment?
\end{itemize}

\noindent \textbf{Language models.}
We use instruction-tuned versions of \gemma~\citep{team2024gemma} as a weaker model and \llama~\citep{dubey2024llama} as the stronger model.

\noindent \textbf{Tasks.}  
We evaluate \methodabb on a comprehensive suite of tasks spanning multiple domains. 1) \textit{General Domain:} We evaluate on established benchmarks including Winogrande \citep{sakaguchi2021Winogrande}, ARC-Challenge \citep{clark2018think}, HellaSwag \citep{zellers2019hellaswag} for commonsense reasoning, and \mmlu \citep{hendrycks2020measuring} for knowledge-based language understanding.
2) \textit{Financial Domain:} We include \fpb and \fiqasa for sentiment analysis, and \headline and \multifin~\citep{xie2023pixiu} for classification, representing domain-specific challenges.
Detailed dataset information can be found in the Appendix~\ref{app:dataset}.

\noindent \textbf{Evaluation Metrics.}
We assess \methodabb via:
\begin{enumerate}
[topsep=0.4pt,itemsep=0.5pt]
    \item \textit{Prediction Accuracy:} We measure the fidelity of performance and cost predictions using Mean Absolute Error (MAE):
$\text{MAE} = \frac{1}{n}\sum_{i=1}^{n}|y_i - \hat{y}_i|$\label{eq:mae}. Lower MAE indicates better prediction accuracy.
\item \textit{Cost Efficiency:} We quantify computational savings using Cost Reduction Ratio (CRR):
\begin{equation}
\text{CRR} = \frac{C_{\text{full}} - C_{\text{ours}}}{C_{\text{full}}} \times 100\%\label{eq:crr} 
\end{equation}
where $C_{\text{full}}$ represents total cost of evaluating all adaptation configurations, and $C_{\text{ours}}$ is the total cost of \methodabb to predict all those possibilities.
\end{enumerate}

\noindent \textbf{Setup.}
We evaluate \methodabb with two representative adaptation paradigms:  \qlora fine-tuning in training-time strategies and retrieval-augmented \icl in test-time strategies. 
For \qlora, the configuration space we explore includes a grid of hyperparameters including training iterations $\in$ \{4, 5, 6, 7, 8\} and data portions $\in$ \{0.1, ..., 1.0\} at 0.1 increments. 
For \icl, we vary the number of retrieved demonstrations $\in$ \{1, 2, 4, 8, 16\}, constrained by the model's maximum sequence length (8,196 tokens for both \llama and \gemma). 
We employ retrieval-augmented \icl using a BM25 ~\cite{robertson2009probabilistic} retriever to identify demonstrations from the training set.
This results in 55 distinct transformation combinations. 
All experiments are conducted with three random seeds and results are averaged.
We present additional details in Appendices~\ref{app:exp-details} and~\ref{app:perf-cost-pred}.

We partition the strategy space into three cost bands (low, medium, high) by uniformly dividing the range between minimum and maximum observed costs for each task, then categorize strategies into these bands based on their computational costs. 
Within each band, we evaluate \methodabb's ability to identify strategies that optimize the accuracy-cost tradeoff by maximizing predicted accuracy while minimizing the total monetary cost. Our score function over performance and cost is thus: $s(\pi, c) = \pi - \epsilon\frac{c}{c_{\text{max}}}$ where $c_{\text{max}}$ is the maximum cost in that cost band, and $\epsilon$ is a small positive constant (\eg, $\epsilon = 10^{-6}$).

\begin{table*}[t]
\centering
\tiny
\setlength{\tabcolsep}{2.6pt}
\begin{tabular}{l|ccc|ccc|ccc|ccc|ccc|ccc|ccc|ccc|c}
\toprule
\multicolumn{1}{l|}{Tasks} 
& \multicolumn{3}{c|}{\textbf{MMLU}} & \multicolumn{3}{c|}{\textbf{Winogrande}} & \multicolumn{3}{c|}{\textbf{ARC-Challenge}} & \multicolumn{3}{c|}{\textbf{HellaSwag}} 
& \multicolumn{3}{c|}{\textbf{FPB}} & \multicolumn{3}{c|}{\textbf{FiQA-SA}} & \multicolumn{3}{c|}{\textbf{Headline}} & \multicolumn{3}{c|}{\textbf{Multifin EN}} & 
\multicolumn{1}{c}{\multirow{2}{*}{\textbf{Avg.}}} 
\\
\cmidrule(lr){2-4} \cmidrule(lr){5-7} \cmidrule(lr){8-10} \cmidrule(lr){11-13} \cmidrule(lr){14-16} \cmidrule(lr){17-19} \cmidrule(lr){20-22} \cmidrule(lr){23-25} 
Cost Level & L & M & H & L & M & H & L & M & H & L & M & H & L & M & H & L & M & H & L & M & H & L & M & H &  \\
\midrule
Pred. Acc (\%) & 61.42 & 62.10 & 61.97 & 58.30 & 63.92 & 65.75 & 78.12 & 76.76 & 76.64 & 94.38 & 93.68 & 93.15 & 83.26 & 85.29 & 84.78 & 82.41 & 83.97 & 83.40 & 95.44 & 96.62 & 96.80 & 80.91 & 83.94 & 85.76 & - \\
Act. Acc (\%) & 61.58 & 62.33 & 61.97 & 63.27 & 66.54 & 67.19 & 79.48 & 79.37 & 77.89 & 94.38 & 94.11 & 93.31 & 84.98 & 85.98 & 86.01 & 84.54 & 85.96 & 85.67 & 96.06 & 96.73 & 96.90 & 80.91 & 83.94 & 85.76 & - \\
\rowcolor[HTML]{EFEFEF}
MAE $\downarrow$ & 0.16 & 0.23 & 0.00 & 4.97 & 2.62 & 1.44 & 1.36 & 2.61 & 1.25 & 0.00 & 0.43 & 0.16 & 1.72 & 0.69 & 1.23 & 2.13 & 1.99 & 2.27 & 0.62 & 0.11 & 0.10 & 0.00 & 0.00 & 0.00 & \textbf{1.09} \\
\midrule
Act. Cost (\$) & 10.08 & 17.50 & 10.96 & 0.35 & 0.62 & 0.44 & 0.69 & 1.12 & 0.92 & 13.30 & 17.28 & 10.39 & 1.44 & 2.51 & 1.78 & 0.26 & 0.43 & 0.36 & 8.58 & 15.10 & 10.71 & 0.29 & 0.50 & 0.36 & - \\
Ours Cost (\$) & 0.33 & 0.32 & 0.14 & 0.07 & 0.04 & 0.03 & 0.09 & 0.05 & 0.04 & 0.67 & 0.52 & 0.22 & 0.10 & 0.07 & 0.04 & 0.06 & 0.03 & 0.02 & 0.41 & 0.33 & 0.17 & 0.08 & 0.05 & 0.03 & - \\
\rowcolor[HTML]{EFEFEF}
CRR $\uparrow$ (\%) & 96.68 & 98.17 & 98.71 & 80.91 & 93.99 & 94.31 & 87.35 & 95.35 & 96.12 & 94.99 & 96.99 & 97.90 & 92.88 & 97.33 & 97.81 & 76.48 & 92.77 & 93.38 & 95.19 & 97.81 & 98.44 & 70.92 & 89.84 & 90.95 & \textbf{92.72} \\
\bottomrule
\end{tabular}
\caption{
Results summary of predicted vs. actual optimal strategies across tasks and low (L), medium (M), high (H) cost regimes (over \emph{55} strategy combinations of \qlora and \icl).
Our method achieves substantial cost reduction across all levels (92.72\% average savings) while maintaining prediction fidelity (1.09\% mean absolute error). Cost efficiency scales favorably with task size, with larger tasks demonstrating greater absolute cost savings. Results demonstrate consistent performance across low, medium, and high-cost bands.
}
\label{tab:strategy-prediction-performance}
\end{table*}

\subsection{How well does \methodabb address the \problemabb problem?}\label{sec5.1-overall-perf}
Table~\ref{tab:strategy-prediction-performance} presents a comprehensive evaluation of \methodabb, comparing predicted vs. actual optimal strategies across 8 diverse tasks and multiple cost regimes on \llama. 
For clarity, we report the accuracy of the predicted strategy, the actual best achievable accuracy, and MAE (Eq.~\ref{eq:mae}). For each cost level, we report the total cost of running all combinations (Act. Cost), the total cost of running \methodabb (Ours Cost), and CRR (Eq.~\ref{eq:crr}).
Our approach demonstrates remarkable efficiency-accuracy trade-offs, achieving an average cost reduction of 92.72\% while maintaining strong prediction fidelity with only 1.09\% MAE. 
Notably, \methodabb exhibits two key scaling properties:
(1) cost savings systematically increase as we move from low to high-cost ranges (improvement ranging from 2.03\% for \mmlu to 20.03\% for \multifin), suggesting better prediction capability in computationally intensive scenarios, and 
(2) cost efficiency improves with task scale, with larger tasks demonstrating greater absolute cost savings (\eg, MMLU: \$9.74-\$17.18, HellaSwag: \$12.64-\$16.76).
This dual scaling behavior suggests that \methodabb becomes increasingly advantageous as both computational demands and task complexity grow. 

We provide detailed performance metrics including cost ranges and direct comparisons with established search-based methods in Appendix~\ref{app:detailed_perf}. 
We also present evaluation results across an expanded model pool that further validate COSMOS's generalizability and efficiency in Appendix~\ref{app:expandmodel}.

\subsection{Strategy-Specific Analysis}\label{sec5.2:detailed-pred-analysis}

Having established \methodabb's overall effectiveness in Section~\ref{sec5.1-overall-perf}, we now present a detailed strategy-specific analysis of its prediction capabilities on all combinations.

\begin{figure*}[h]
    \includegraphics[width=\textwidth]{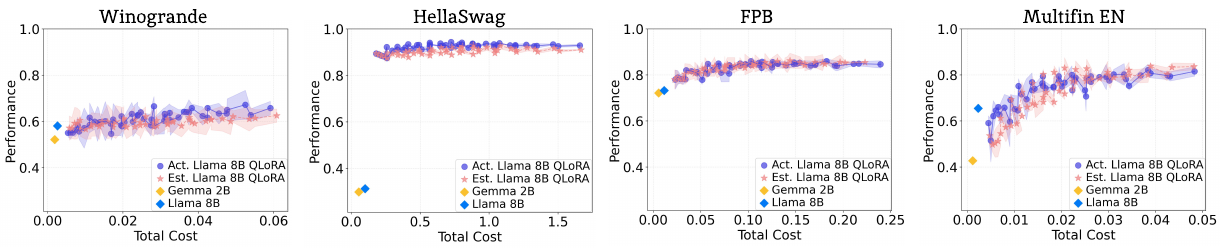}
    \vspace{-1em}
    \caption{Predicted vs. actual performance-cost analysis for \qlora fine-tuning. Each plot compares actual (\textcolor{actualQloraColor}{$\bullet$}) vs. predicted (\textcolor{predictedColor}{$\star$}) performance-cost trajectories for \llama \qlora fine-tuning.
    Base models \gemma~($\gemmadiamond$), and \llama~($\llamadiamond$) serve as reference points.
    The closer predicted performance-cost trajectories (red) to the actual (purple) trajectories indicates better performance-cost prediction.
    The results show a consistent alignment between predicted and actual curves across both general and domain-specific tasks. This demonstrates \methodabb's robust prediction capabilities.}
    \label{fig:detailed_qlora_perf_cost}
\end{figure*}

\noindent \textbf{Fine-tuning.}
Figure~\ref{fig:detailed_qlora_perf_cost} demonstrates the prediction accuracy for \qlora fine-tuning across our task suite.
Each point represents a distinct fine-tuning configuration's performance-cost outcome (\eg, training on 50\% data for 5 iterations). 
The results reveal strong prediction capabilities for all tasks.
For instance, on \fpb, \methodabb achieves notably high accuracy with MAE of 0.007 for both performance and cost predictions. 
We observe improved prediction accuracy at higher computational budgets, where fine-tuning performance stabilizes. 
Even in situations with limited training data and high-performance variance (low-cost scenario), \methodabb is able to capture the vibrate pattern and indicates if fine-tuning is worthwhile.
These patterns persist across task domains, indicating \methodabb's robust generalization.

\begin{figure*}[t]
    \includegraphics[width=\textwidth]{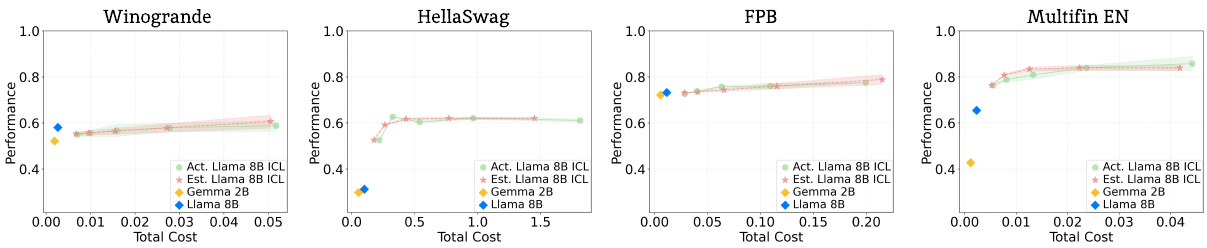}
    \vspace{-1em}
    \caption{Predicted vs. actual performance-cost analysis for retrieval-based \icl. Each plot compares actual (\textcolor{actualIclColor}{$\bullet$}) vs. predicted (\textcolor{predictedColor}{$\star$}) performance-cost trajectories for \llama \icl.
    Base models Gemma 2B ($\gemmadiamond$), and Llama 8B ($\llamadiamond$) serve as reference points.
    The consistent alignment between predicted and actual curves across all tasks demonstrates \methodabb's robust prediction capabilities.}
    \label{fig:detailed_icl_perf_cost}
\end{figure*}

\noindent \textbf{In-context learning.}
Figure~\ref{fig:detailed_icl_perf_cost} demonstrates the prediction accuracy for retrieval augmented \icl across our task suite.
Each point represents a distinct \icl configuration's performance-cost outcome (\eg, providing 8 demonstrations of input-output pairs in the query). 
The results reveal strong prediction capabilities across general-domain benchmarks
and specialized financial tasks. 
For instance, on \fiqasa, \methodabb achieves notably high accuracy with MAE of 0.003 and 0.001 for performance and cost predictions, respectively. 
Full results and detailed analyses examining the accuracy and cost predictions separately for all tasks and strategies are provided in Appendix~\ref{app:details-pred-cap-zoom-in}.
\vspace{-0.5em}
\subsection{
Combining Training- and Test-time Strategies
}\label{sec5.3:scaling-behavior}

\begin{figure*}[h]
\includegraphics[width=\textwidth]{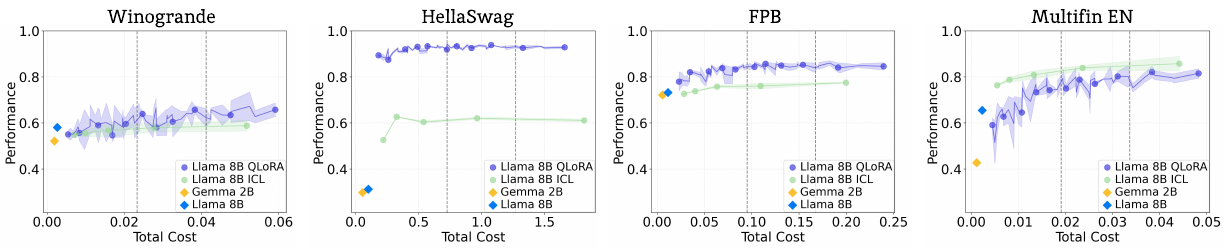}
\caption{Actual \qlora vs. \icl performance-cost trajectories across diverse tasks. Each plot presents the performance-cost curves for \qlora (\textcolor{actualQloraColor}{$\bullet$}) and \icl (\textcolor{actualIclColor}{$\bullet$}) on \llama, with \gemma ($\gemmadiamond$), and \llama ($\llamadiamond$) serving as baselines.
Vertical dashed lines demarcate low, medium, and high-cost thresholds, determined by the minimum and maximum costs of both adaptation strategies. 
The shaded regions represent the standard deviation across 3 seeds for each configuration.}
\label{fig:actual-qlora-icl-performance-cost}
\end{figure*}

We also conduct an analysis of the fundamental trade-offs between training-time and inference-time adaptation strategies by comparing \qlora fine-tuning and retrieval-augmented \icl as illustrated in Figure~\ref{fig:actual-qlora-icl-performance-cost}. Full results can be found in Appendix~\ref{app:combine-train-test}. We find: 

(1) \textbf{Non-linear Scaling Behaviors.} 
Both adaptation strategies exhibit diminishing returns with increased computational investment, despite consistently outperforming the base \llama model.
Maximizing computational resources (shots for \icl or iterations/data for \qlora) does not guarantee optimal performance. 

(2) \textbf{Stability-Performance Trade-offs.}
\qlora and \icl demonstrate distinct stability characteristics across different cost regimes. While \qlora shows higher performance variance, particularly evident in \multifin where performance fluctuates significantly in low-cost settings ($\leq$\$0.019) before stabilizing at higher thresholds ($\geq$\$0.034), retrieval-augmented \icl maintains more consistent performance profiles, especially in resource-constrained scenarios.

(3) \textbf{Resource-dependent Strategy Selection.}
The optimal choice between fine-tuning and prompting depends on available resources. Fine-tuning typically achieves superior performance in medium to high-cost scenarios. \icl is a more reliable option in low-resource settings. 

(4) \textbf{Hybrid Strategy Benefits.} 
Strategically combining the approaches can achieve superior performance at lower costs. \methodabb  can help produce this selection efficiently.

\begin{wrapfigure}[23]{r}{0.51\textwidth}
\begin{center}
\includegraphics[width=.9\columnwidth]{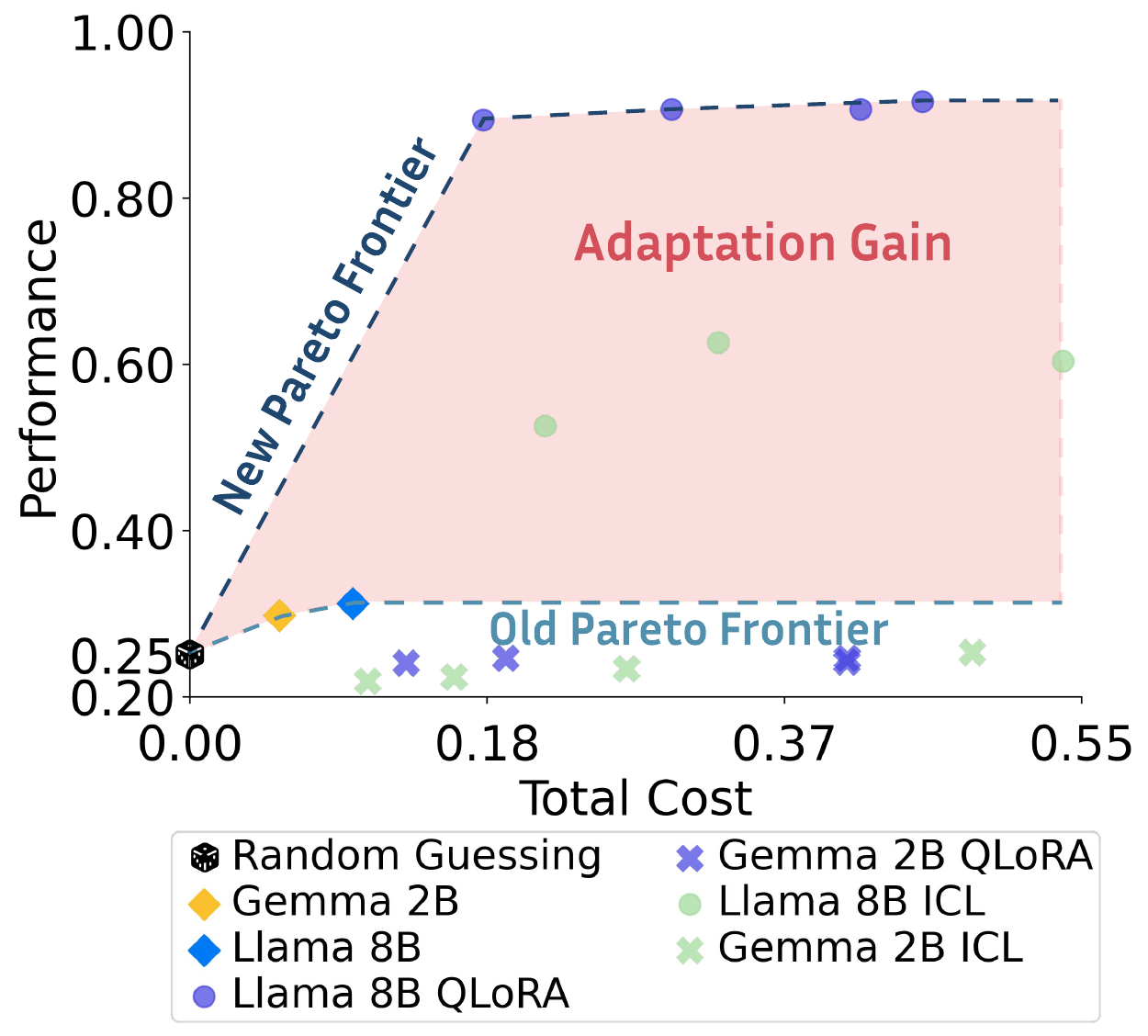}
\end{center}
\vspace{-1em}
\caption{Benefit of adaptation-augmented routing. 
The \textit{Old Pareto Frontier} (\textcolor{oldPareto}{$--$}) connects performance-cost points achievable through traditional routing between base models; 
\textit{New Pareto Frontier} (\textcolor{newPareto}{$--$}) incorporates adaptation strategies (\qlora, \icl). The red-shaded region represents the \textit{adaptation gains}---quantifying how much better performance can be achieved at the same cost, or how much cost can be reduced for the same performance, by expanding the strategy space beyond pure model selection.
} 
\label{fig:pareto-frontier}
\end{wrapfigure}

\subsection{Augmenting Routing}\label{sec5.4:aug-routing}

While traditional model routing focuses on selecting from a pool of base models for each query, we demonstrate that expanding the routing space to include adaptation strategies can significantly enhance the performance-cost frontier. 
Instead of simply choosing between base models like \gemma and \llama, our approach considers a richer set of options that includes both model selection and adaptation strategies such as \qlora fine-tuning and retrieval-augmented \icl.

The benefits of this expanded strategy space are illustrated in Figure~\ref{fig:pareto-frontier}. 
In conventional routing between \gemma and \llama (old Pareto frontier), the performance-cost tradeoff is limited to interpolating between these base models. However, by incorporating adaptation strategies, we establish a new Pareto frontier that substantially dominates the original one. This expansion leads to significant adaptation gains, visualized by the shaded red region between the old and new frontiers, quantifying the superior performance that can be achieved at the same cost or the cost reduction for equivalent performance.

Importantly, while this expanded strategy space offers practitioners more flexibility, exhaustively evaluating all possible combinations becomes intractable as the space grows exponentially.
We mitigate this challenge through \methodabb, which efficiently predicts the performance and cost of adaptation strategies without requiring extensive computation, as validated in Sections~\ref{sec5.1-overall-perf} and~\ref{sec5.2:detailed-pred-analysis}. 
This enables informed strategy selection while avoiding the computational overhead of evaluating every configuration, making adaptation-augmented routing practical and scalable.

\subsection{Potential Implications}\label{sec5.5:implications}
Fine-tuning large language models (LLMs) at scale presents significant financial challenges. In Appendix~\ref{app:implication}, we analyze these costs through a practical case study of fine-tuning GPT-4o, using OpenAI's current pricing structure. We obtain a rough approximation of cost savings of ~\$939,830, bringing down the cost by a factor of 24.7x.
\section{Conclusion}
We formalized and studied the strategy selection problem--determining optimal combinations of models, adaptation approaches, and configurations while balancing performance and cost constraints. We introduced  \methodabb, a unified framework to predict both task performance and costs.
We explored our framework's effectiveness through two representative strategies: (1) a lightweight embedding-augmented linear proxy model for predicting fine-tuning performance, and (2) a low-sample scaling laws predictor for retrieval-augmented in-context learning. We validated capabilities of both components via exhaustive experiments. \methodabb enables practitioners to make informed decisions by providing a flexible and cost-effective approach to strategy selection based on their specific performance-cost trade-off preferences. 

\bibliography{ref}
\bibliographystyle{icml2025}

\newpage
\appendix
\onecolumn
\begin{center}
	\textbf{\LARGE Appendix }
\end{center}
The appendix is organized as follows:
We first detail the datasets
 (Appendix~\ref{app:dataset}) and experimental setup (Appendix~\ref{app:exp-details}).
Next, we provide comprehensive information about our performance and cost prediction frameworks in Appendix~\ref{app:perf-cost-pred}). 
We then present detailed evaluations of COSMOS (Appendices~\ref{app:detailed_perf}-\ref{app:details-pred-cap-zoom-in}), including extended results for Sections~\ref{sec5.1-overall-perf} and~\ref{sec5.2:detailed-pred-analysis}, comparisons against baselines, and in-depth strategy-specific analyses.
Additionally, we offer extended results for strategy combinations in Appendix~\ref{app:combine-train-test}, discusses broader implications in Appendix~\ref{app:implication}. 
We demonstrate COSMOS's generalizability across diverse model families in Appendix~\ref{app:expandmodel} and with limited data access in Appendix~\ref{app:limiteddata}. 
Finally, we present a concrete example illustrating how practitioners can conduct strategy selection using COSMOS's predicted metrics in Appendix~\ref{app:concrete-example}.

\section{Datasets Details}\label{app:dataset}
Our evaluation spans both general-domain and domain-specific tasks, with dataset sizes varying from 380 to 21,570 examples (Table~\ref{tab:dataset-detail}). 
This range allows us to systematically investigate model performance across both low-resource and data-rich scenarios.

For general-domain benchmarks (\mmlu, \wino, \hella, and \arc), we maintain consistency with RouterBench~\citep{hu2024routerbench} by adopting their prompting templates. 
This choice ensures comparability with model routing settings and eliminates potential performance variations due to prompt differences. 
For financial domain tasks~(\fpb, \fiqasa, \headline, \multifin), we preserve the original dataset format as these typically include well-structured instructions and question-answer pairs. 
\begin{table}[!ht]
\centering
\begin{tabular}{l>{\raggedleft\arraybackslash}r}
\toprule
\multicolumn{1}{l}{\textbf{Tasks}} & \multicolumn{1}{r}{\textbf{\# of Train}} \\
\midrule
\texttt{\mmlu} & 9,809 \\
\texttt{\wino} & 886 \\
\texttt{\hella} & 7,029 \\
\texttt{\arc} & 1,029 \\
\texttt{\fpb} & 3,100 \\
\texttt{\fiqasa} & 750 \\
\texttt{\headline} & 21,570 \\
\texttt{\multifin} & 380 \\
\bottomrule
\end{tabular}
\caption{Number of training examples per task.}
\label{tab:dataset-detail}
\end{table}

\section{Experimental Details}\label{app:exp-details}
\paragraph{Training and retrieval setup.}
For datasets without predefined splits (general tasks), we implement a standard partition ratio of 70/10/20 for training, validation, and testing, respectively.
For \qlora fine-tuning (both full dataset training and prediction-time validation), we select data from the training set and evaluate performance on the test set. To account for data sampling variability when training with different data portions, we report average performance across three random seeds.
For \icl, we employ BM25-based retrieval to dynamically select demonstrations from the training set for each test query. Our reported results represent averages across multiple demonstration orderings: the original retrieval order and two random permutations.

\paragraph{Fine-tuning hyperparameters.}\label{app:hyperparams}
For fine-tuning, we use a learning rate of 2e-4~(following \citet{dettmers2024qlora}), batch size of 1, maximum sequence length of 512, gradient accumulation steps of 2, warmup ratio of 0.03, maximum gradient norm of 0.3, LoRA alpha of 32, LoRA dropout of 0.05, LoRA rank of 64.
To standardize measurements across varying cluster loads, we assume an average step time of 1.09 seconds, which corresponds to the mean processing time on uncontested GPU resources.

\paragraph{Hardware and software.}
We conduct experiments using 15 NVIDIA A100-PCIE-40GB and 1 NVIDIA A100-SXM4-40GB GPUs, with Python 3.10, PyTorch 2.5.1, and Transformers 4.45.2.

\paragraph{Cost assumptions.}
We use an hourly rate of \$1/h for A100-40GB based on Vast.ai pricing listed on a GPU comparison website\footnote{\url{https://cloud-gpus.com/}} for computing-based cost estimation. 
For token-based fine-tuning costs, we extrapolate from Together.ai\footnote{\url{https://www.together.ai/pricing}}'s online cost calculator and fit \llama's pricing using a power law to calculate the total cost per epoch $c_{\text{epoch}} = a \times (N_{\text{token}})^b$, excluding their \$5 minimum cost requirement, where $a \approx 8.69e-7$, $b \approx 0.956$.
For inference cost, we utilize Together.ai's pricing of \$0.2 for \llama per million tokens and \$0.1 for \gemma per million tokens.

\section{Performance and Cost Prediction}\label{app:perf-cost-pred}
\subsection{Fine-tuning}
Our prediction framework employs task-specific approaches based on complexity. For complex tasks (\mmlu, \hella, \wino, \arc), we implement a contrastive learning approach using a lightweight linear projector. This model is trained using correct answers as positive examples and incorrect options as negative samples, with the following configuration: batch size of 8, maximum sequence length of 512, learning rate of 1e-6, and temperature of 0.07. The architecture consists of layer normalization followed by a linear projection layer, with model selection based on peak test accuracy over 300 iterations.

For financial domain tasks, we adopt a more streamlined approach, utilizing a single linear layer trained with cross-entropy loss on model-generated embeddings. This simplified architecture achieves rapid training convergence on CPU, resulting in negligible proxy model training costs.

To ensure robust performance predictions across different data regimes, we employ a calibration process using a minimal subset of training data—either 200 examples or 10\% of the training set, whichever is larger. This one-time calibration yields scaling factors applicable across all data portion configurations (0.1 through 1.0) per epoch.

\paragraph{Cost prediction.}
The training cost is calculated by multiplying GPU usage duration and peak memory utilization by the compute price.
To standardize measurements across varying cluster loads, we assume a fixed processing time of 0.0009 seconds per data point per epoch on an idle NVIDIA A100-PCIE-40GB, which corresponds to the mean processing time on uncontested GPU resources. 

The primary prediction cost stems from validation set training. This explains the varying cost reduction rates across tasks–achieving up to 98.71\% reduction for \mmlu in high-budget scenarios, while showing lower reductions for smaller datasets like \multifin (380 total points). Since validation costs are amortized across all strategies using the same scaling factor (currently 10 in our experiments), increasing the number of strategies or configurations would further improve cost efficiency.

\subsection{Retrieval-augmented \icl}
We discover that retrieval-augmented ICL performance can be effectively predicted using minimal data (as few as 2 samples). Building on this finding, we fit an exponential saturation function (Eq.~\ref{eq:icl-perf}) using performance measurements from 1-shot and 8-shot settings. For the baseline performance $\pi_0$, we select the lower value between zero-shot performance and 1-shot results.

Cost estimation for ICL leverages the average input and output lengths observed in the training set for efficiency.

\section{A Detailed Evaluation and Comparative Analysis of COSMOS}\label{app:detailed_perf}

\subsection{Detailed Performance Analysis of \methodabb}\label{app:detailed_perf_only}

\begin{table*}[h]
\centering
\tiny
\setlength{\tabcolsep}{2.2pt}
\begin{tabular}{l|ccc|ccc|ccc|ccc|ccc|ccc|ccc|ccc}
\toprule
\multicolumn{1}{l|}{Tasks} 
& \multicolumn{3}{c|}{\textbf{MMLU}} & \multicolumn{3}{c|}{\textbf{Winogrande}} & \multicolumn{3}{c|}{\textbf{ARC-Challenge}} & \multicolumn{3}{c|}{\textbf{HellaSwag}} 
& \multicolumn{3}{c|}{\textbf{FPB}} & \multicolumn{3}{c|}{\textbf{\fiqasa}} & \multicolumn{3}{c|}{\textbf{Headline}} & \multicolumn{3}{c}{\textbf{Multifin EN}} \\
\cmidrule(lr){2-4} \cmidrule(lr){5-7} \cmidrule(lr){8-10} \cmidrule(lr){11-13} \cmidrule(lr){14-16} \cmidrule(lr){17-19} \cmidrule(lr){20-22} \cmidrule(lr){23-25}
Cost Level & L & M & H & L & M & H & L & M & H & L & M & H & L & M & H & L & M & H & L & M & H & L & M & H \\
\midrule
Min Acc (\%) & 55.91 & 57.32 & 61.14 & 54.64 & 57.91 & 58.82 & 73.70 & 75.96 & 75.06 & 52.56 & 62.02 & 61.09 & 72.65 & 76.05 & 77.49 & 74.18 & 78.16 & 80.85 & 72.91 & 74.90 & 70.52 & 51.52 & 70.68 & 75.27 \\
Max Acc (\%) & 61.58 & 62.33 & 61.97 & 63.27 & 66.54 & 67.19 & 79.48 & 79.37 & 77.89 & 94.38 & 94.11 & 93.31 & 84.98 & 85.98 & 86.01 & 84.54 & 85.96 & 85.67 & 96.06 & 96.73 & 96.90 & 80.91 & 83.94 & 85.76 \\
Avg Acc (\%) & 59.96 & 61.14 & 61.60 & 58.22 & 62.16 & 63.73 & 77.44 & 77.62 & 76.70 & 88.35 & 91.36 & 88.34 & 80.70 & 84.29 & 83.89 & 80.14 & 82.17 & 83.65 & 91.51 & 95.07 & 93.72 & 70.05 & 77.86 & 80.27 \\
Act. Acc (\%) & 61.58 & 62.33 & 61.97 & 63.27 & 66.54 & 67.19 & 79.48 & 79.37 & 77.89 & 94.38 & 94.11 & 93.31 & 84.98 & 85.98 & 86.01 & 84.54 & 85.96 & 85.67 & 96.06 & 96.73 & 96.90 & 80.91 & 83.94 & 85.76 \\
Range Avg Acc (\%) & 58.75 & 59.83 & 61.55 & 58.95 & 62.22 & 63.01 & 76.59 & 77.62 & 76.47 & 73.47 & 78.07 & 77.20 & 78.81 & 81.01 & 81.75 & 79.36 & 82.06 & 83.65 & 84.49 & 85.81 & 83.71 & 66.21 & 77.31 & 80.52 \\
Pred Acc (\%) & 61.42 & 62.10 & 61.97 & 58.30 & 63.92 & 65.75 & 78.12 & 76.76 & 76.64 & 94.38 & 93.68 & 93.15 & 83.26 & 85.29 & 84.78 & 82.41 & 83.97 & 83.40 & 95.44 & 96.62 & 96.80 & 80.91 & 83.94 & 85.76 \\
\midrule
Min Cost (\$) & 0.163 & 0.677 & 1.189 & 0.005 & 0.024 & 0.042 & 0.011 & 0.047 & 0.079 & 0.181 & 0.758 & 1.270 & 0.023 & 0.097 & 0.171 & 0.004 & 0.018 & 0.031 & 0.135 & 0.584 & 1.034 & 0.004 & 0.019 & 0.034 \\
Max Cost (\$) & 0.639 & 1.152 & 1.663 & 0.023 & 0.041 & 0.059 & 0.044 & 0.072 & 0.110 & 0.725 & 1.192 & 1.815 & 0.092 & 0.166 & 0.239 & 0.017 & 0.028 & 0.044 & 0.550 & 0.999 & 1.448 & 0.018 & 0.033 & 0.048 \\
Avg Cost (\$) & 0.388 & 0.875 & 1.370 & 0.013 & 0.031 & 0.049 & 0.026 & 0.059 & 0.092 & 0.443 & 0.960 & 1.484 & 0.055 & 0.125 & 0.197 & 0.010 & 0.023 & 0.036 & 0.330 & 0.755 & 1.190 & 0.011 & 0.025 & 0.040 \\
Act. Total Cost (\$) & 10.076 & 17.501 & 10.963 & 0.351 & 0.621 & 0.443 & 0.688 & 1.116 & 0.920 & 13.305 & 17.282 & 10.385 & 1.440 & 2.506 & 1.775 & 0.261 & 0.430 & 0.357 & 8.580 & 15.097 & 10.713 & 0.287 & 0.504 & 0.360 \\
Range Avg Cost (\$) & 0.401 & 0.914 & 1.426 & 0.014 & 0.033 & 0.051 & 0.027 & 0.059 & 0.095 & 0.453 & 0.975 & 1.543 & 0.057 & 0.131 & 0.205 & 0.010 & 0.023 & 0.036 & 0.342 & 0.792 & 1.241 & 0.011 & 0.026 & 0.041 \\
Ours Total Cost (\$) & 0.335 & 0.320 & 0.142 & 0.067 & 0.037 & 0.025 & 0.087 & 0.052 & 0.036 & 0.666 & 0.520 & 0.218 & 0.103 & 0.067 & 0.039 & 0.061 & 0.031 & 0.024 & 0.413 & 0.331 & 0.167 & 0.083 & 0.051 & 0.033 \\
\bottomrule
\end{tabular}
\caption{
Comprehensive evaluation of our strategy prediction framework across 8 diverse tasks under different cost regimes. Results compare predicted vs. actual optimal strategies across low (L), medium (M), and high (H) cost settings, evaluating \emph{55} combinations of \qlora and \icl techniques. 
The analysis encompasses multiple accuracy metrics (predicted, actual, mean, extremal values, and range averages) and their corresponding cost measurements, demonstrating our method's effectiveness in identifying optimal strategies while maintaining performance across varying computational budgets.
}
\label{tab:strategy-prediction-performance-details}
\end{table*}

In Section~\ref{sec5.1-overall-perf}, we presented the comparison of predicted versus actual optimal strategies. Here, we provide a comprehensive analysis through Table~\ref{tab:strategy-prediction-performance-details}, which details multiple performance dimensions across different cost regimes. For each task and cost level (Low/Medium/High), we report both accuracy and cost metrics. The accuracy metrics include our strategy's predicted performance (Pred Acc), the actual optimal strategy's performance (Act Acc), and statistical measures across all strategies (minimum, maximum, and average accuracy). This allows us to evaluate our method's effectiveness from multiple angles. To quantify the cost-efficiency of our approach, we compare three key metrics: (1) the total cost of exhaustively evaluating all 55 strategy combinations (Act Total Cost), (2) our method's prediction cost (Ours Total Cost), and (3) baseline costs from random strategy selection within each cost band (Range Avg Cost).

To further illustrate our method's effectiveness, consider the \hella task under the high-cost regime: our predictor achieves 93.15\% accuracy at \$0.218, significantly outperforming random strategy selection which yields 77.2\% accuracy at \$1.543. This demonstrates that our approach not only maintains near-optimal performance but also reduces computational costs by over 7x compared to random selection within the target cost range.

We also present a concrete example on how to select the optimal strategy based on predicted metrics given by COSMOS in Appendix~\ref{app:concrete-example}.

\subsection{Comparative Evaluation Against Search-based Methods}\label{app:baselines-search}

\begin{table}[!ht]
\centering
\begin{tabular}{lccc}
\toprule
\textbf{Cost Level} & \textbf{Methods} & \textbf{Acc.} & \textbf{Prediction Cost$\downarrow$ (\$)} \\
\midrule
\multirow{4}{*}{Low} & Oracle & 0.944 & - \\
 & RS-CV & 0.933 & 1.474 \\
 & SH-CV & 0.923 & 1.391 \\
 & COSMOS (Ours) & \textbf{0.944} & \textbf{0.666} \\
\midrule
\multirow{4}{*}{Medium} & Oracle & 0.941 & - \\
 & RS-CV & 0.936 & 3.621 \\
 & SH-CV & 0.934 & 3.999 \\
 & COSMOS (Ours) & \textbf{0.937} & \textbf{0.520} \\
\midrule
\multirow{4}{*}{High} & Oracle & 0.933 & - \\
 & RS-CV & 0.927 & 5.990 \\
 & SH-CV & 0.931 & 5.914 \\
 & COSMOS (Ours) & \textbf{0.932} & \textbf{0.218} \\
\bottomrule
\end{tabular}
\caption{COSMOS achieves near-oracle accuracy (99.3\%-100\%) while reducing computational costs by up to 27.1x compare to RS-CV and SH-CV baselines across all budget constraints.}
\label{tab:method-comparison}
\end{table}

We further conduct a study comparing COSMOS against two established hyperparameter optimization approaches: Random Search with Cross-Validation (RS-CV) and Successive Halving with Cross-Validation (SH-CV) on the HellaSwag benchmark. 
Our evaluation focuses on two critical metrics: (1) prediction accuracy compared to the oracle (optimal performance), and (2) computational cost efficiency. 
As demonstrated in Table~\ref{tab:method-comparison}, COSMOS consistently outperforms both baselines across all cost constraints. In terms of prediction accuracy, our approach matches or closely approximates the oracle performance across all budget levels (achieving 100\%, 99.3\%, and 99.9\% of oracle accuracy respectively), while simultaneously delivering substantial cost reductions.

The computational efficiency advantages of COSMOS are also noteworthy. 
When compared to the best-performing baseline in each cost regime, COSMOS reduces computational expenditure by factors of 2.2×, 7.0×, and 27.1× for low, medium, and high budget scenarios respectively. For instance, in the high-cost regime, COSMOS achieves 99.9\% of oracle accuracy (0.932 vs. 0.933) while requiring only \$0.218 in prediction costs—a 27.1× reduction compared to SH-CV's \$5.914.

Beyond these quantitative improvements, COSMOS offers a fundamental paradigm shift from traditional approaches. 
Unlike search-based methods that require running complete experiments to determine actual performance and cost, COSMOS accurately forecasts these metrics with minimal computational overhead. 
This predictive capability eliminates the need to execute full experimental cycles or extensive validation procedures, enabling practitioners to make informed decisions without incurring the substantial computational costs associated with conventional methods. 
This prediction-based approach is particularly valuable in resource-constrained environments where practitioners need reliable performance estimates without committing to extensive computational expenditures.

\section{Detailed Analysis of Prediction Capabilities of \methodabb}\label{app:details-pred-cap-zoom-in}

\subsection{Full Results for Strategy-specific Analysis}
\begin{figure*}[h]
\begin{subfigure}[t]{0.24\textwidth}
\includegraphics[width=\textwidth]{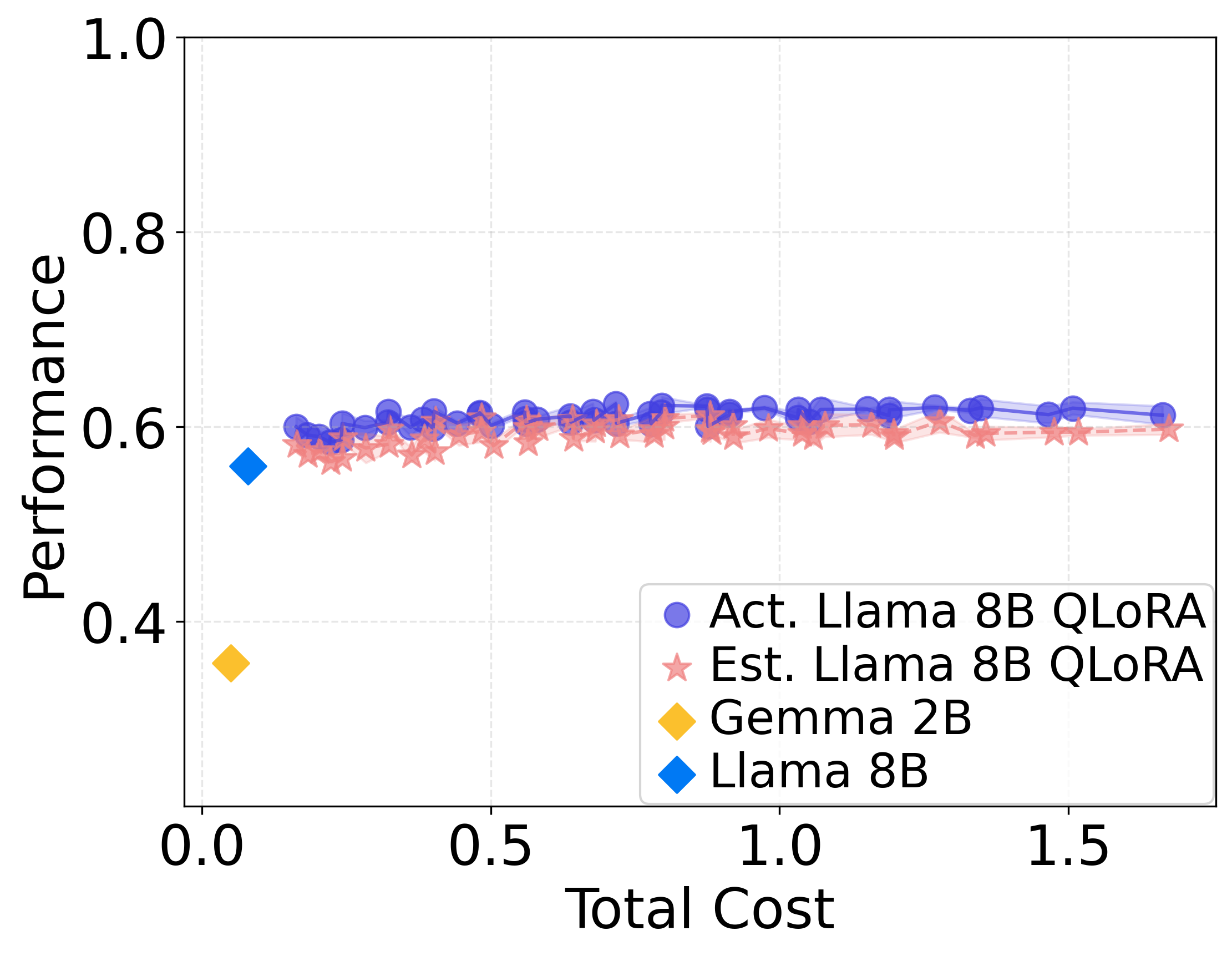}
\caption{\mmlu}
\end{subfigure}
\hfill
\begin{subfigure}[t]{0.24\textwidth}
\includegraphics[width=\textwidth]{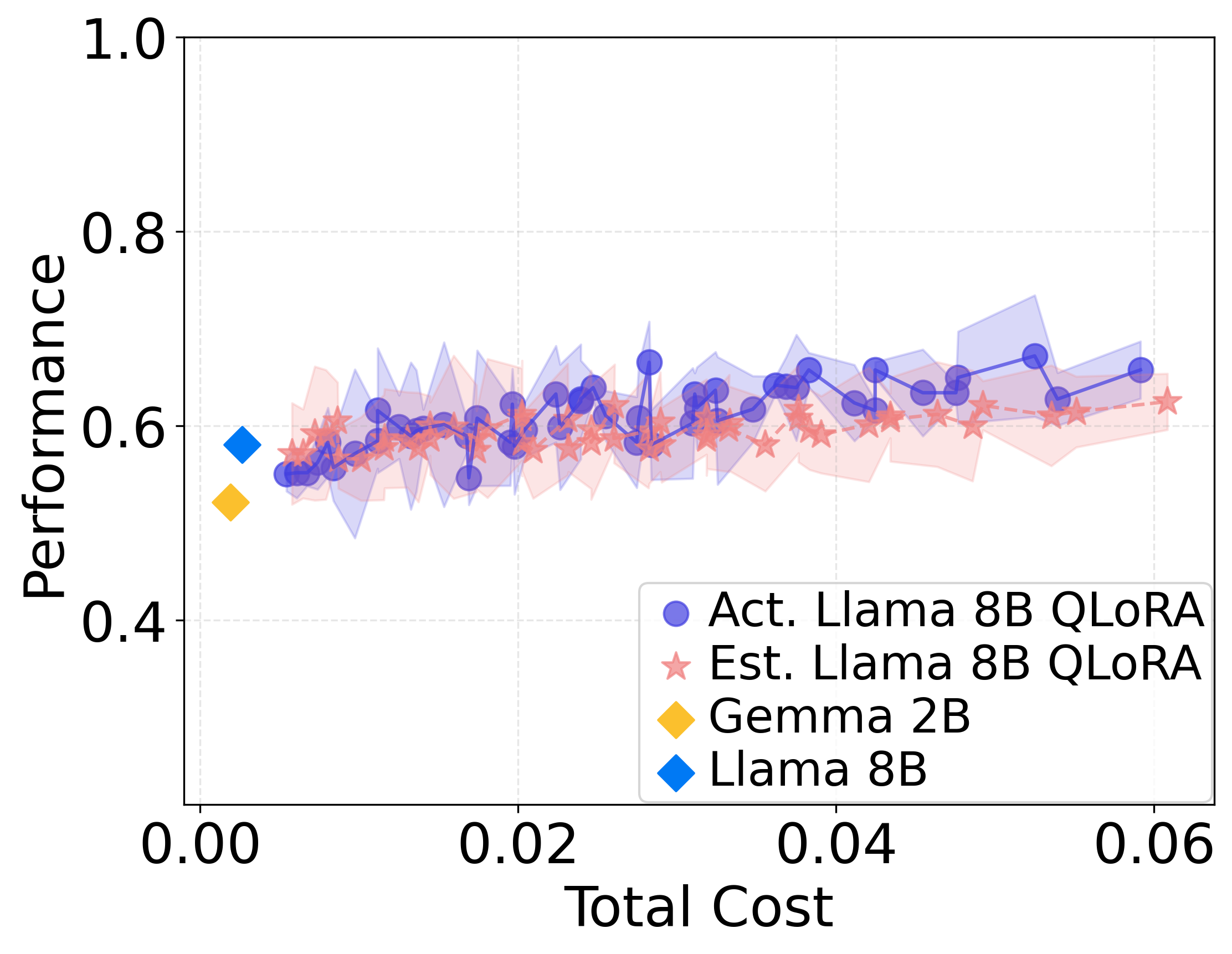}
\caption{\wino}
\end{subfigure}
\hfill
\begin{subfigure}[t]{0.24\textwidth}
\includegraphics[width=\textwidth]{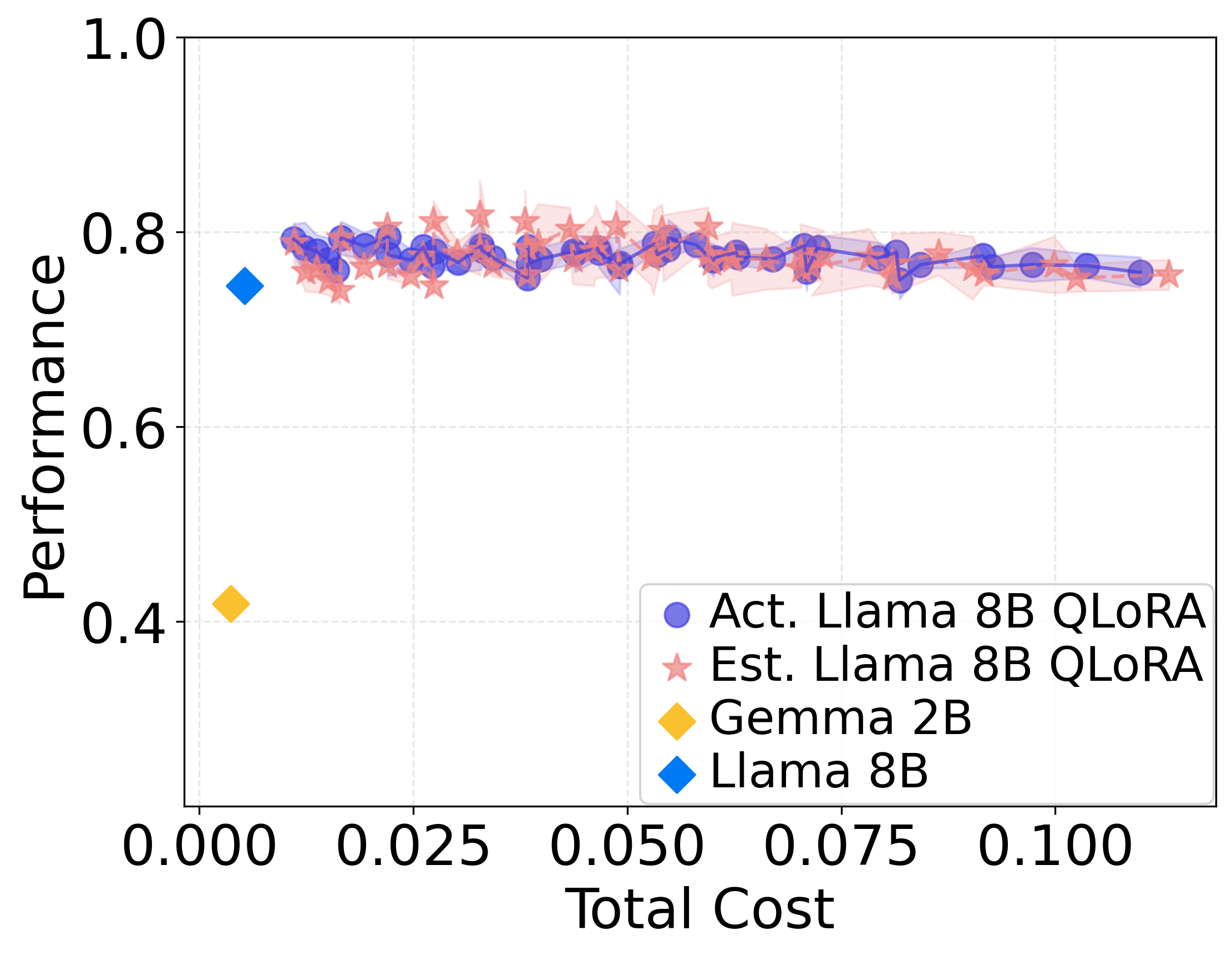}
\caption{\arc}
\end{subfigure}
\hfill
\begin{subfigure}[t]{0.24\textwidth}
\includegraphics[width=\textwidth]{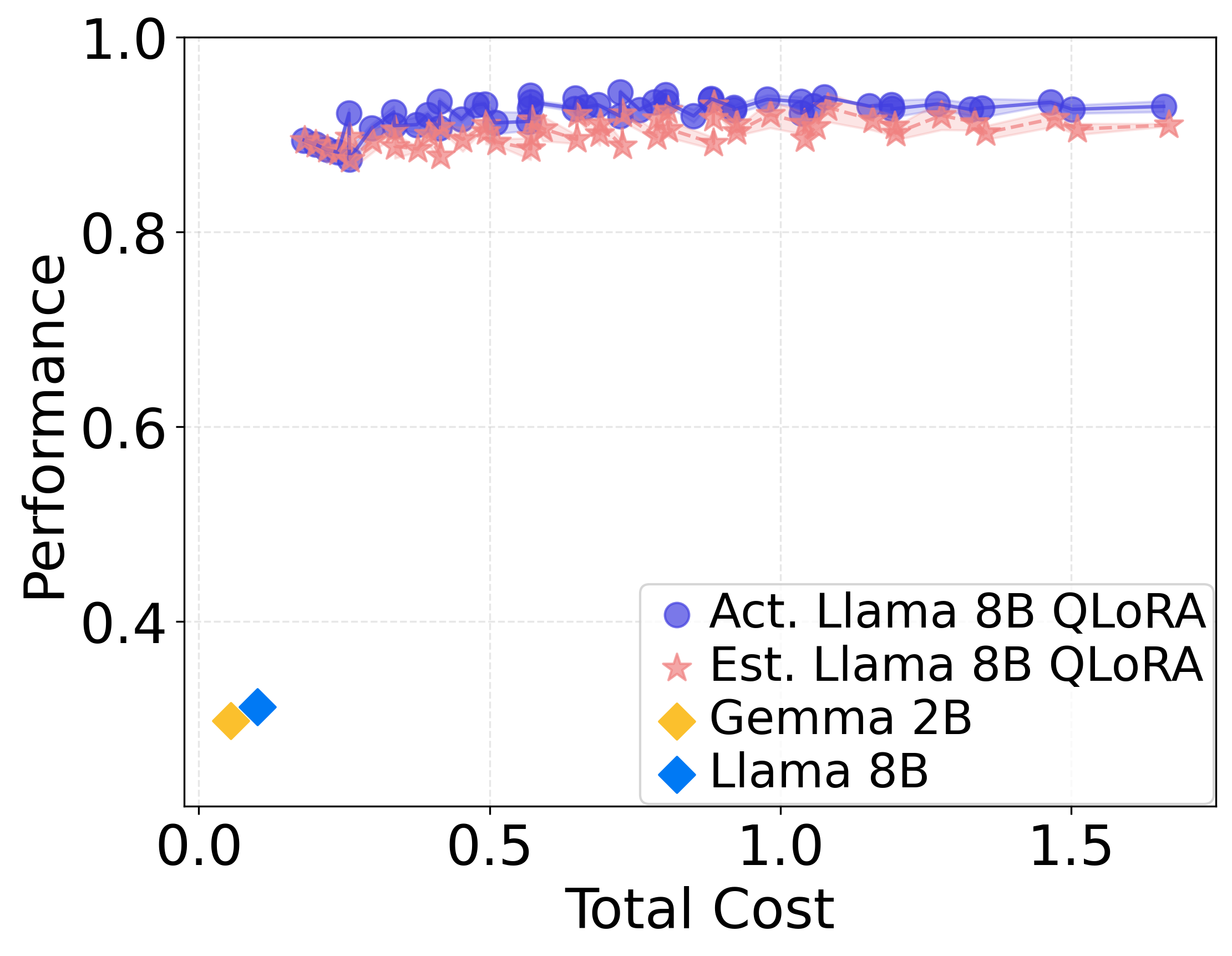}
\caption{\hella}
\end{subfigure}
\\[1em]
\begin{subfigure}[t]{0.24\textwidth}
\includegraphics[width=\textwidth]{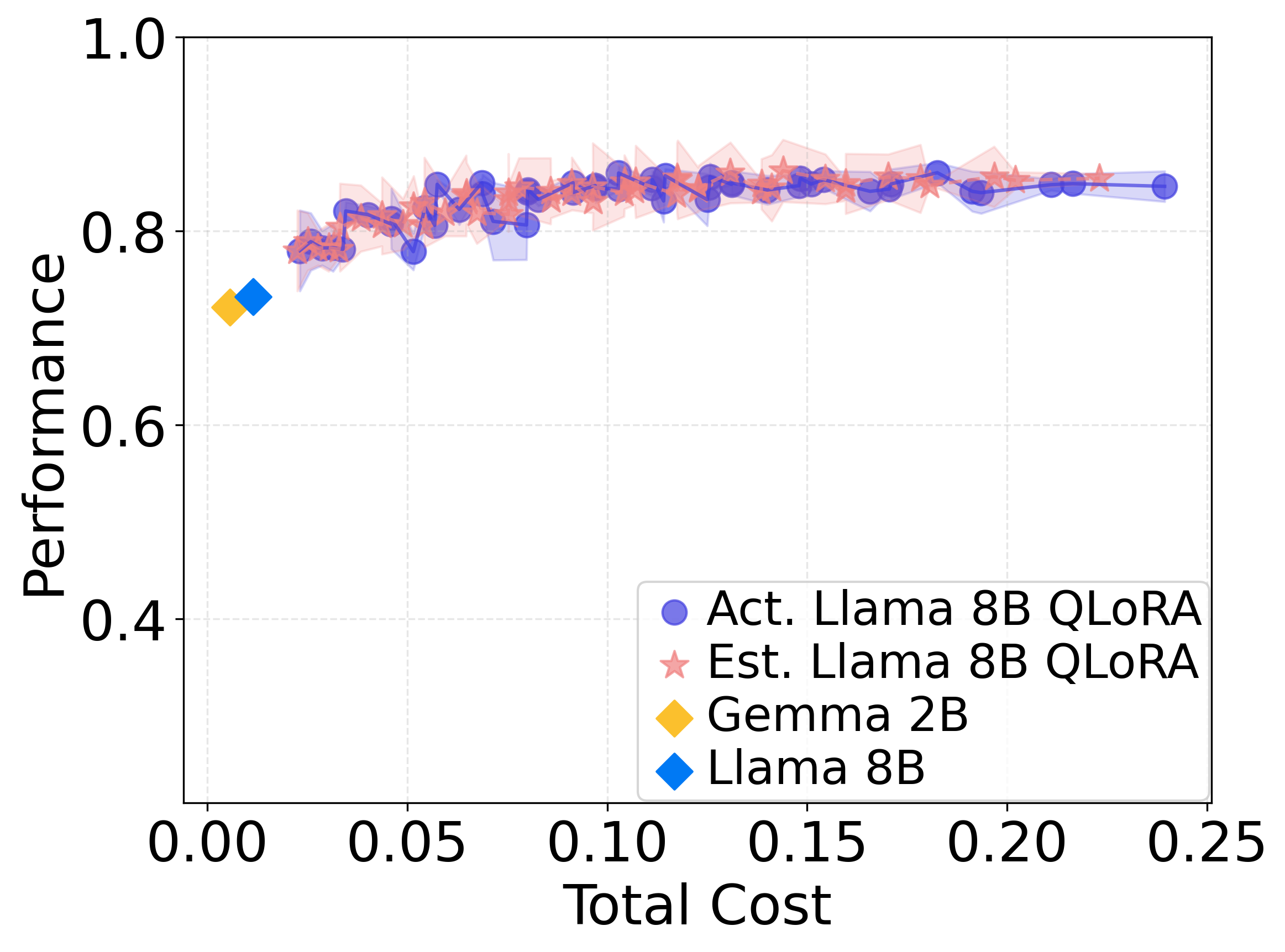}
\caption{\fpb}
\end{subfigure}
\hfill
\begin{subfigure}[t]{0.24\textwidth}
\includegraphics[width=\textwidth]{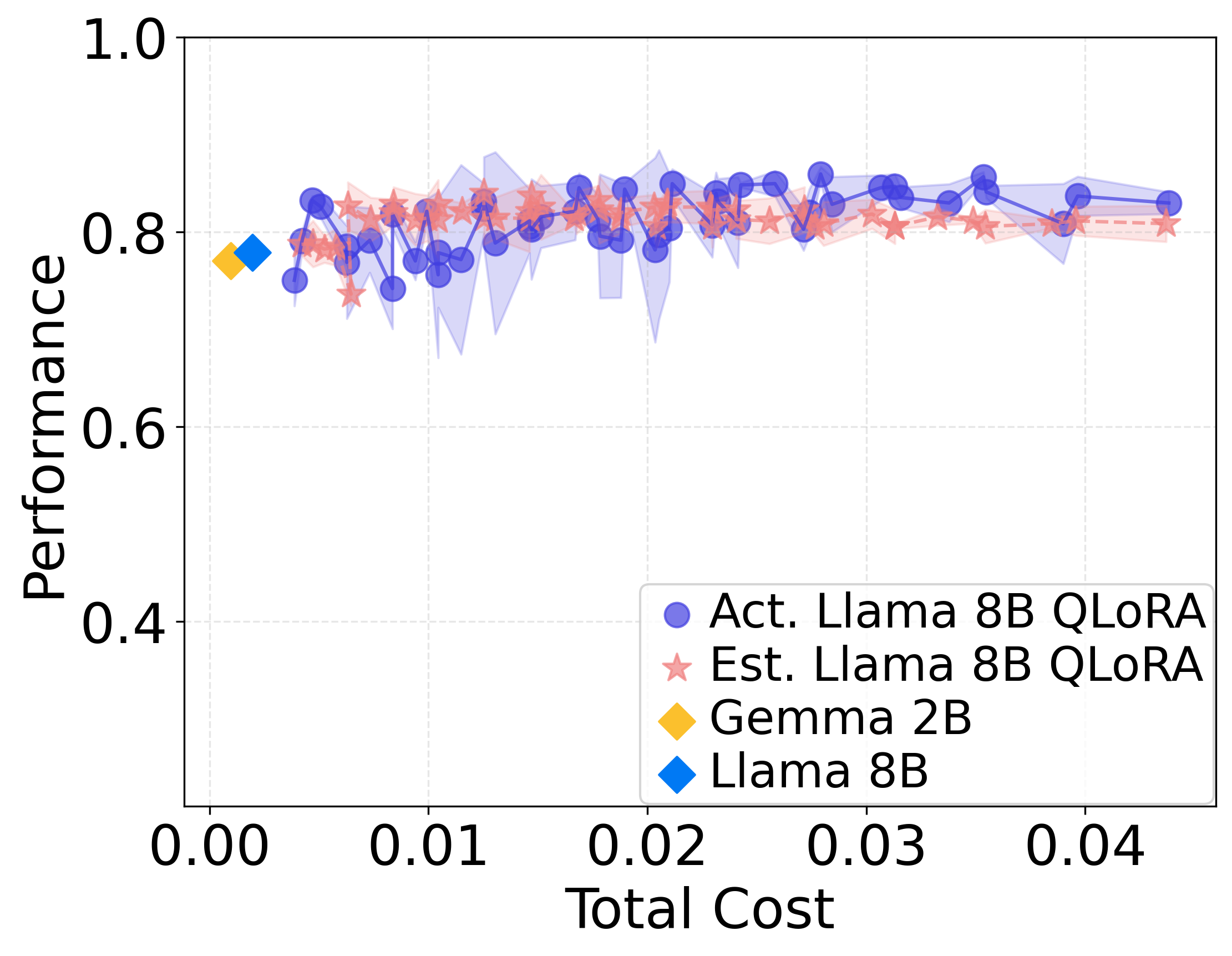}
\caption{\fiqasa}
\end{subfigure}
\hfill
\begin{subfigure}[t]{0.24\textwidth}
\includegraphics[width=\textwidth]{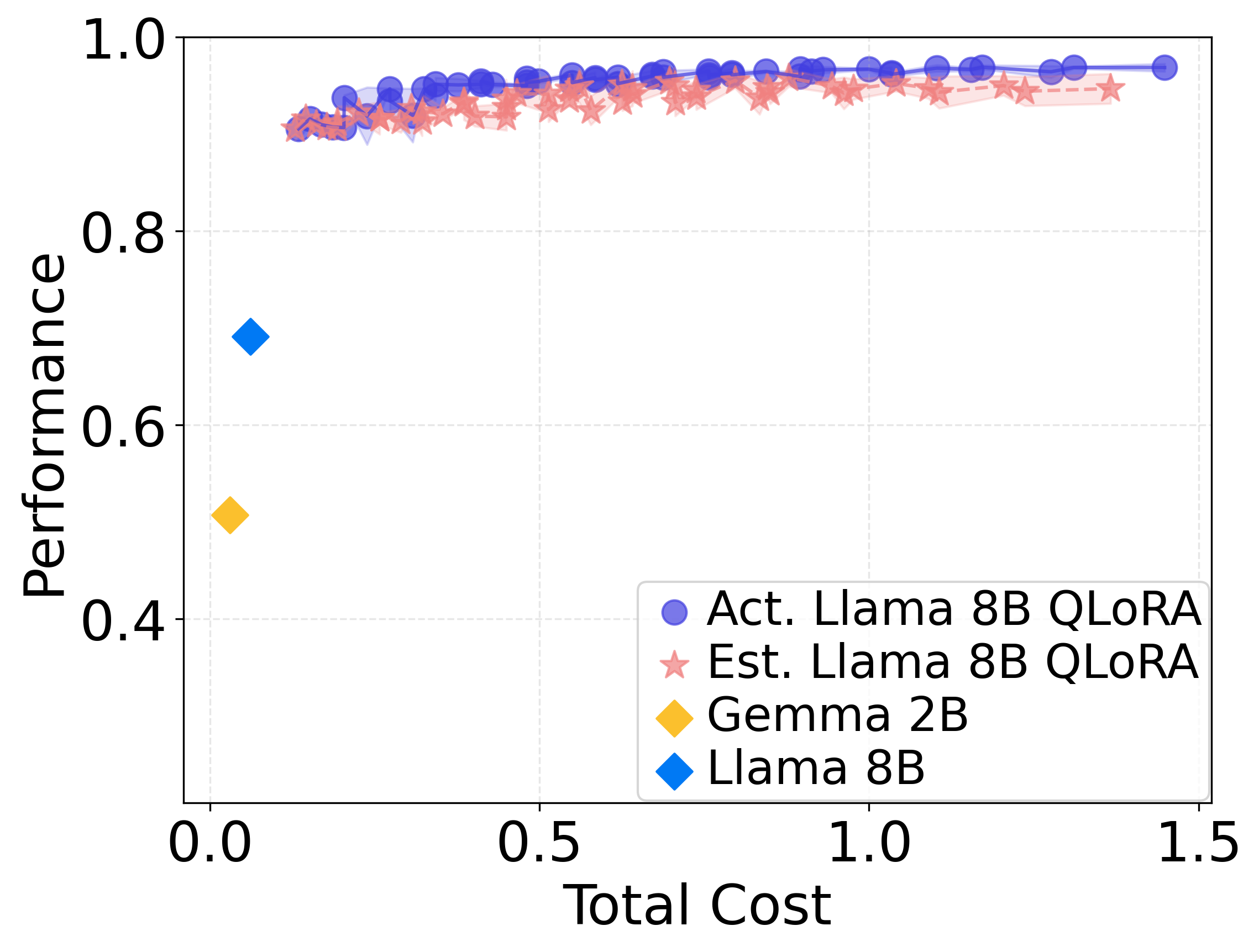}
\caption{\headline}
\end{subfigure}
\hfill
\begin{subfigure}[t]{0.24\textwidth}
\includegraphics[width=\textwidth]{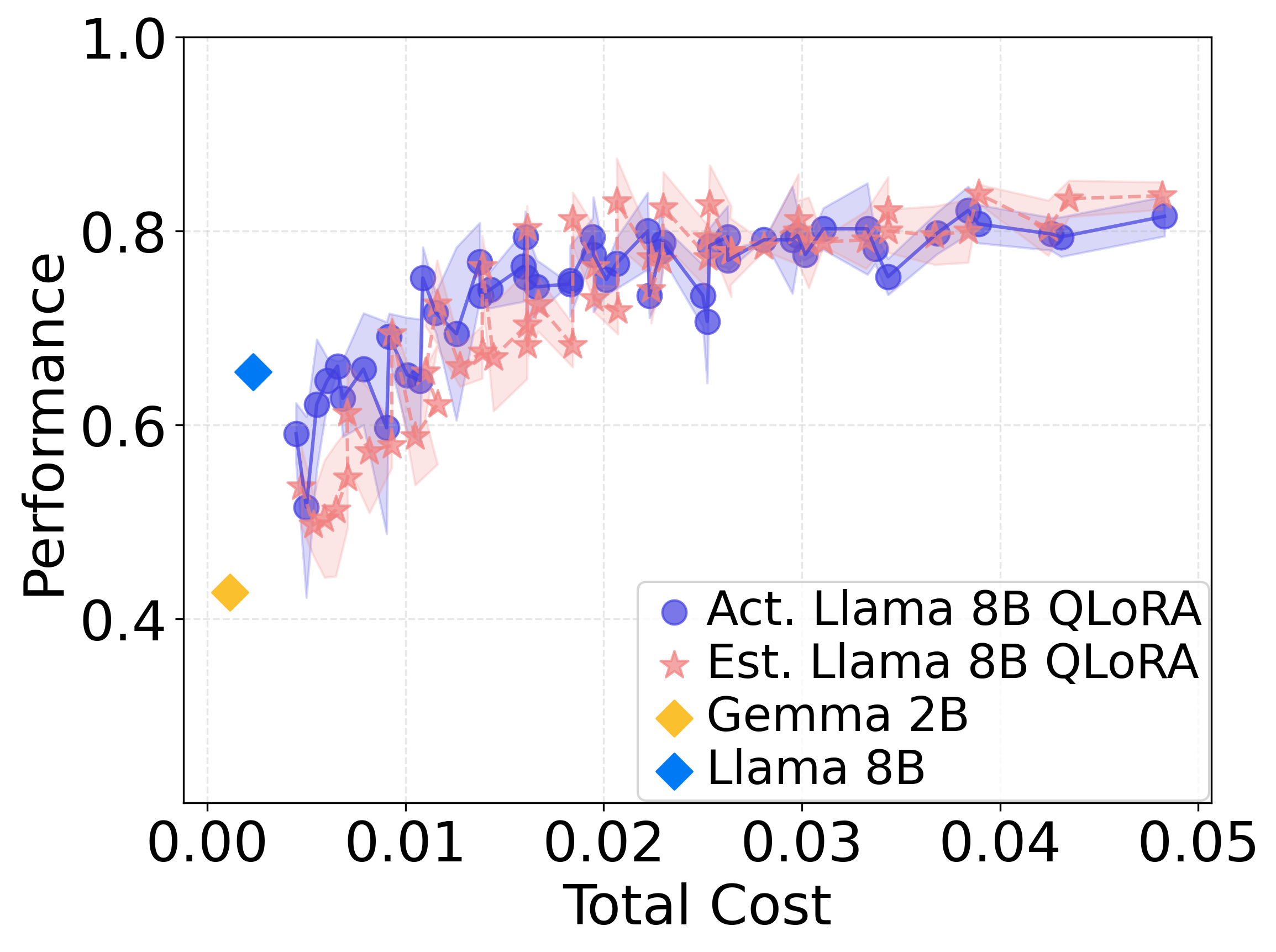}
\caption{\multifin}
\end{subfigure}
\caption{Predicted vs. actual performance-cost analysis for \qlora fine-tuning across eight diverse tasks. Each plot compares actual (\textcolor{actualQloraColor}{$\bullet$}) vs. predicted (\textcolor{predictedColor}{$\star$}) performance-cost trajectories for \llama \qlora fine-tuning.
Base models \gemma~($\gemmadiamond$), and \llama~($\llamadiamond$) serve as reference points.
The closer predicted performance-cost trajectories (red) to the actual (purple) trajectories indicates better performance-cost prediction.
The results show a consistent alignment between predicted and actual curves across both general and domain-specific tasks. This demonstrates \methodabb's robust prediction capabilities.}
\label{fig:detailed_qlora_perf_cost_full}
\end{figure*}

\begin{figure*}[t]
\begin{subfigure}[t]{0.24\textwidth}
\includegraphics[width=\textwidth]{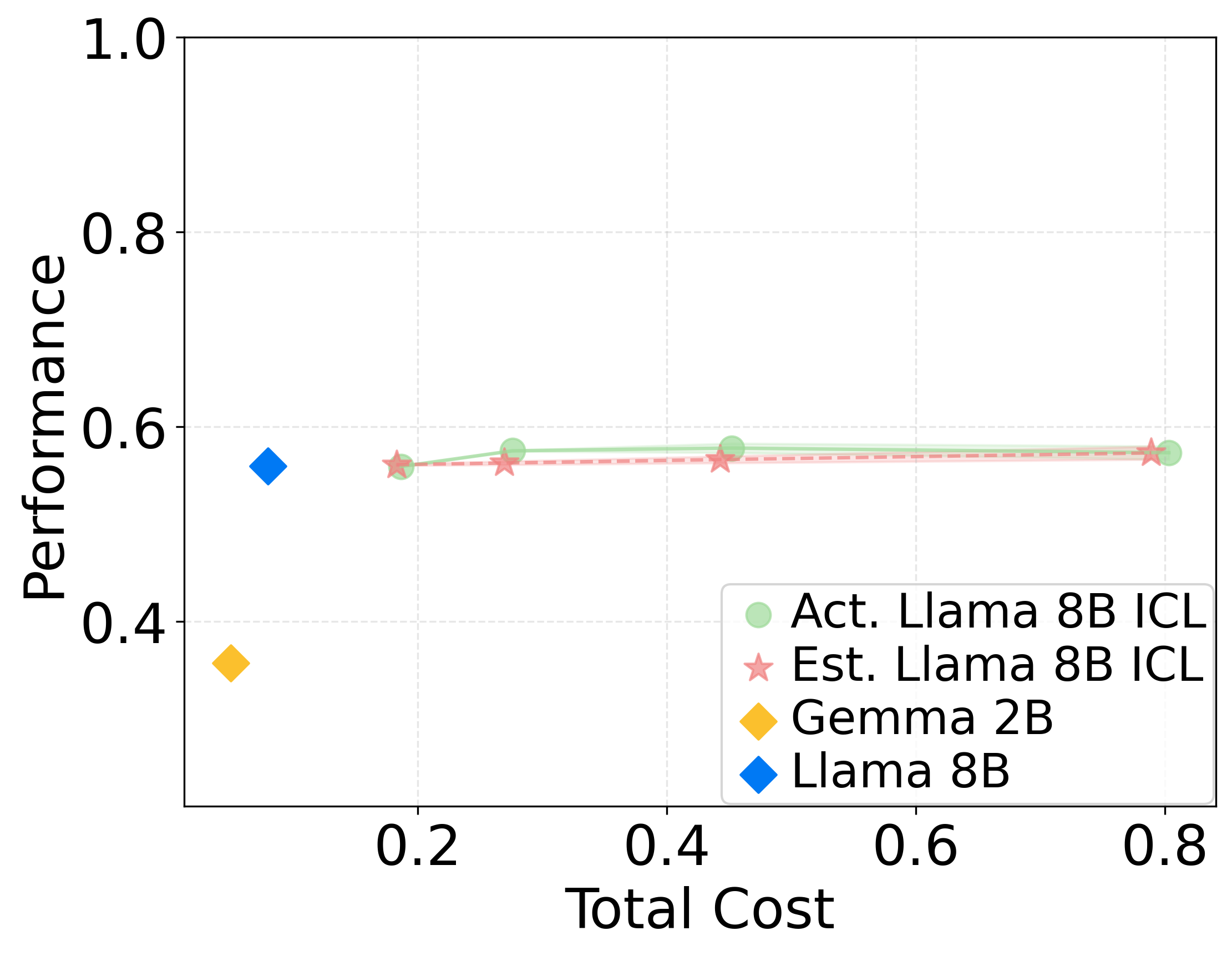}
\caption{\mmlu}
\end{subfigure}
\hfill
\begin{subfigure}[t]{0.24\textwidth}
\includegraphics[width=\textwidth]{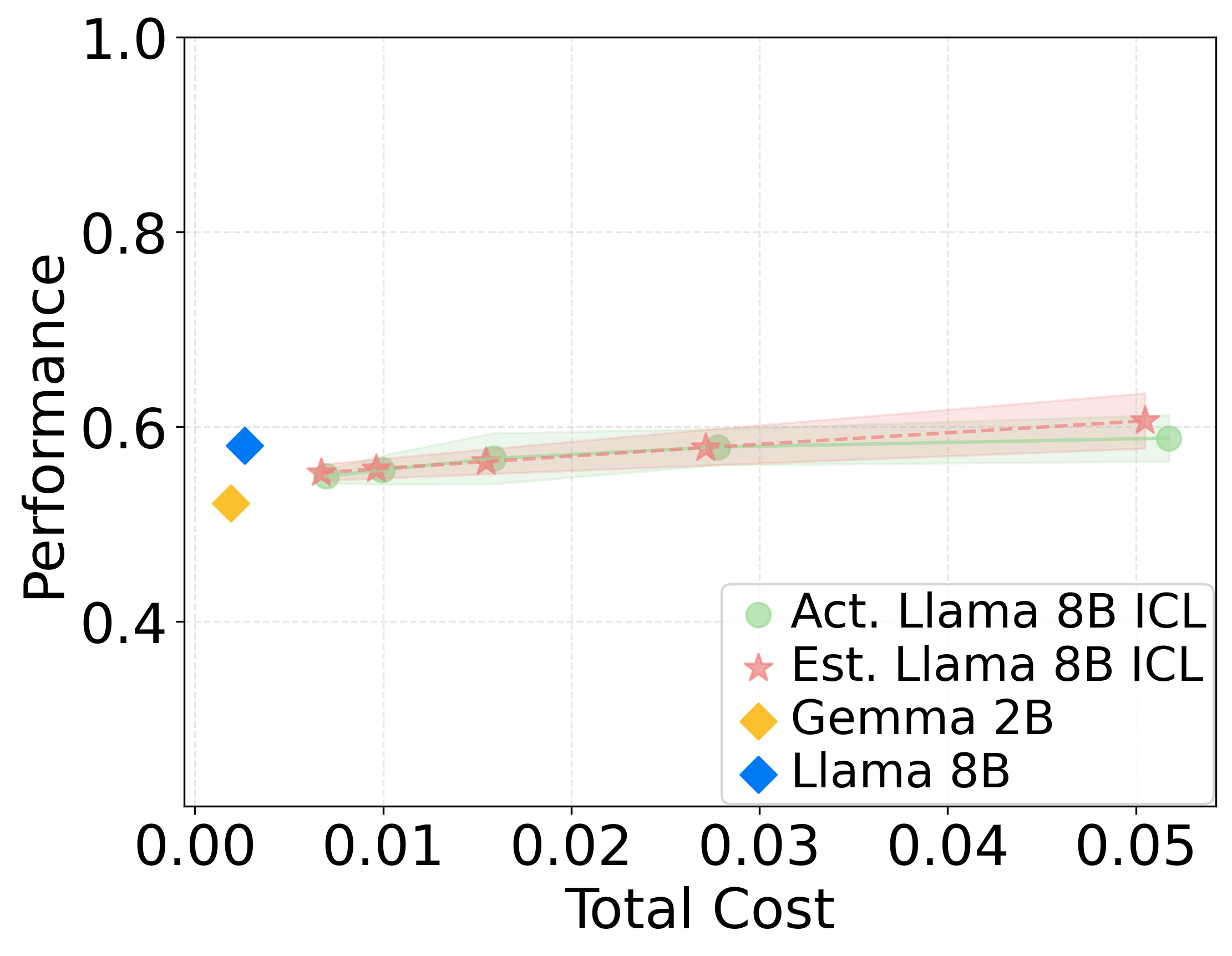}
\caption{\wino}
\end{subfigure}
\hfill
\begin{subfigure}[t]{0.24\textwidth}
\includegraphics[width=\textwidth]{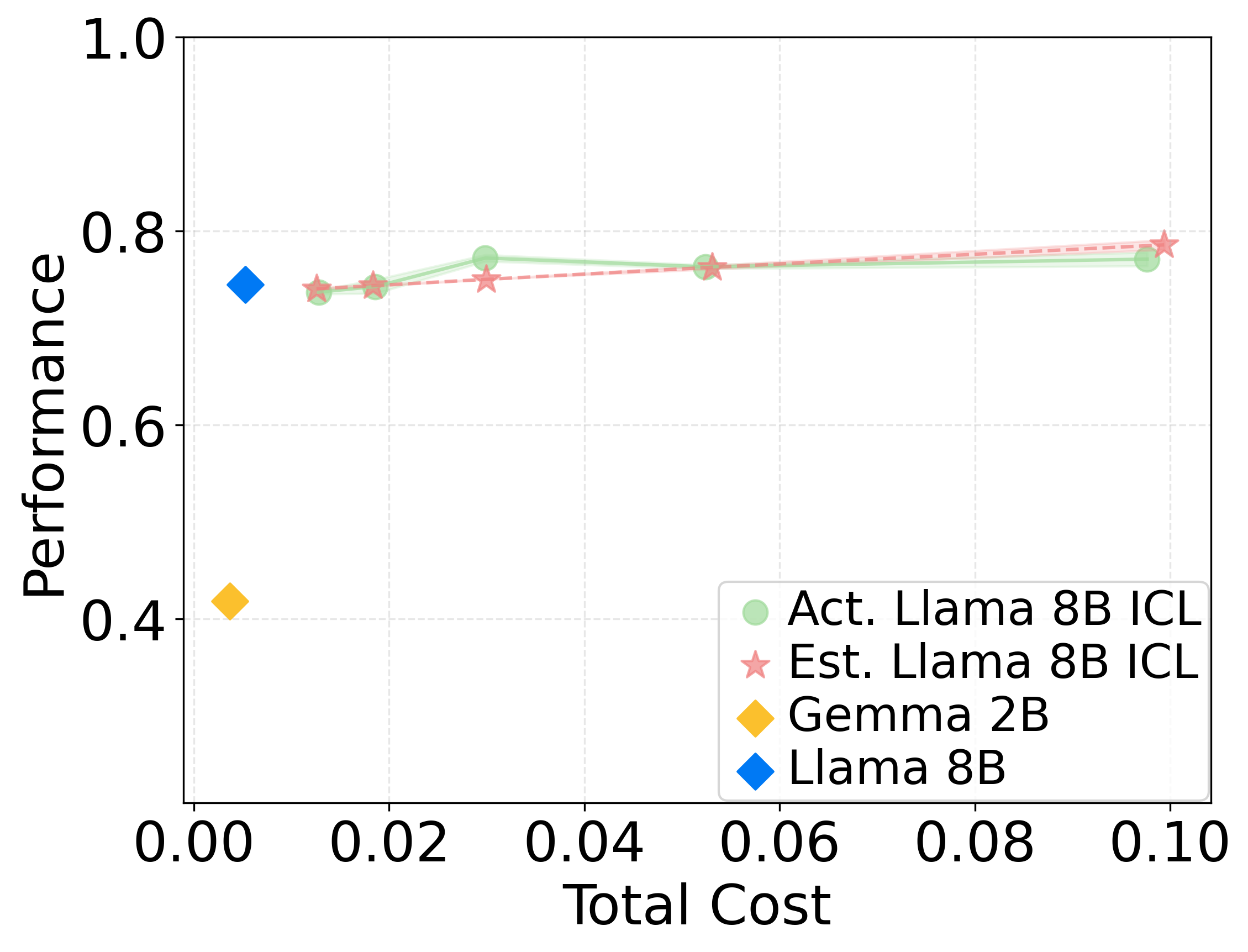}
\caption{\arc}
\end{subfigure}
\hfill
\begin{subfigure}[t]{0.24\textwidth}
\includegraphics[width=\textwidth]{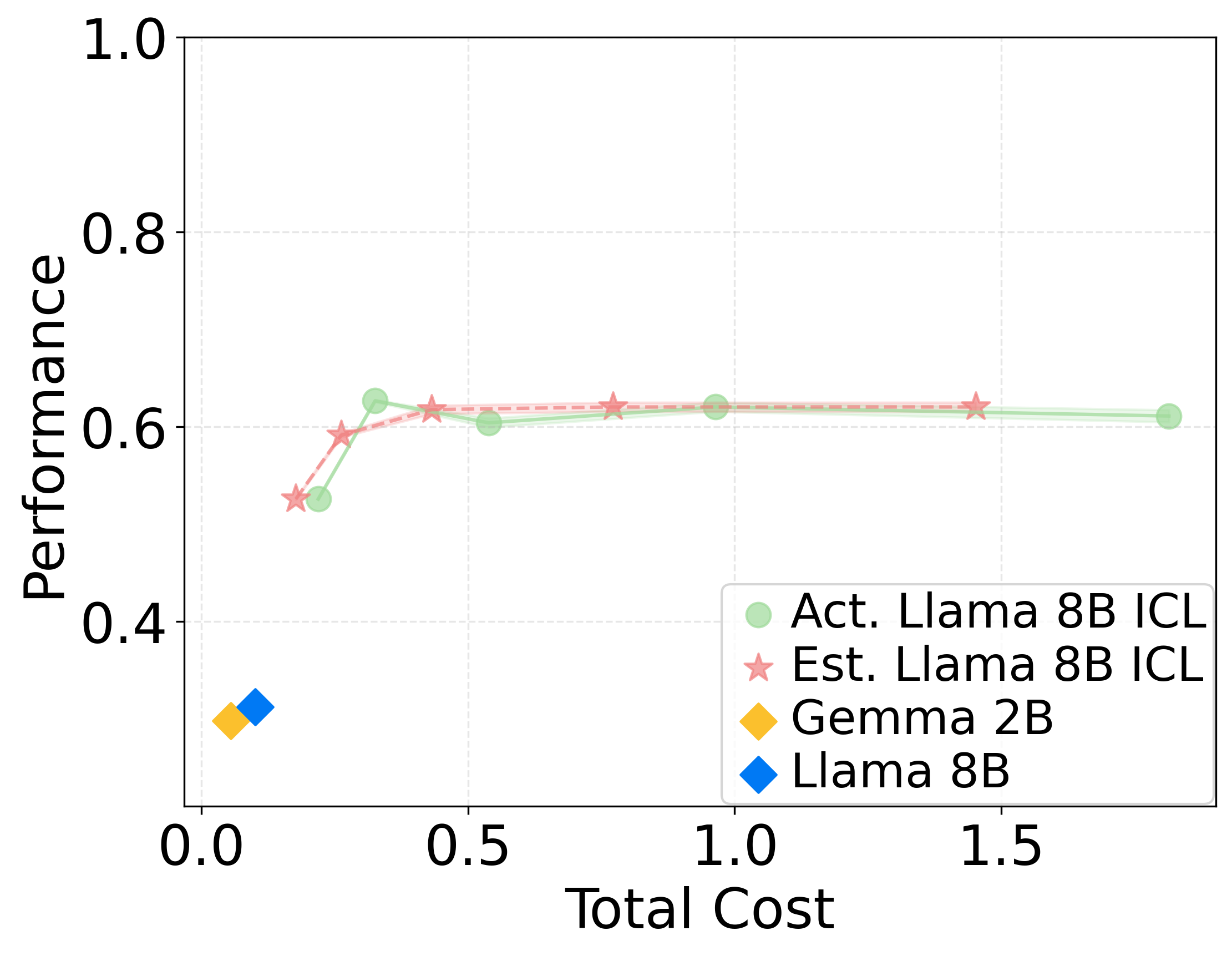}
\caption{\hella}
\end{subfigure}
\\[1em]
\begin{subfigure}[t]{0.24\textwidth}
\includegraphics[width=\textwidth]{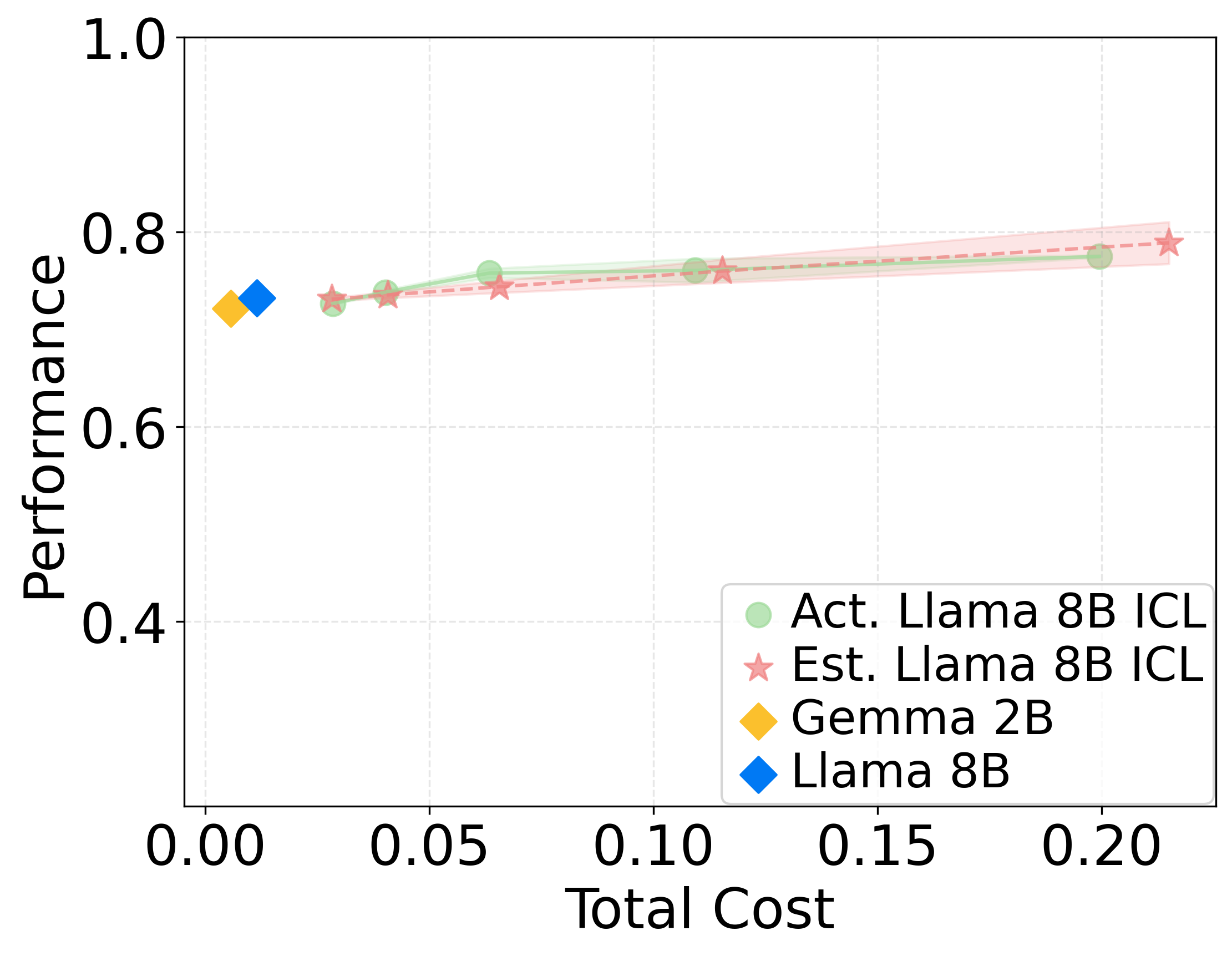}
\caption{\fpb}
\end{subfigure}
\hfill
\begin{subfigure}[t]{0.24\textwidth}
\includegraphics[width=\textwidth]{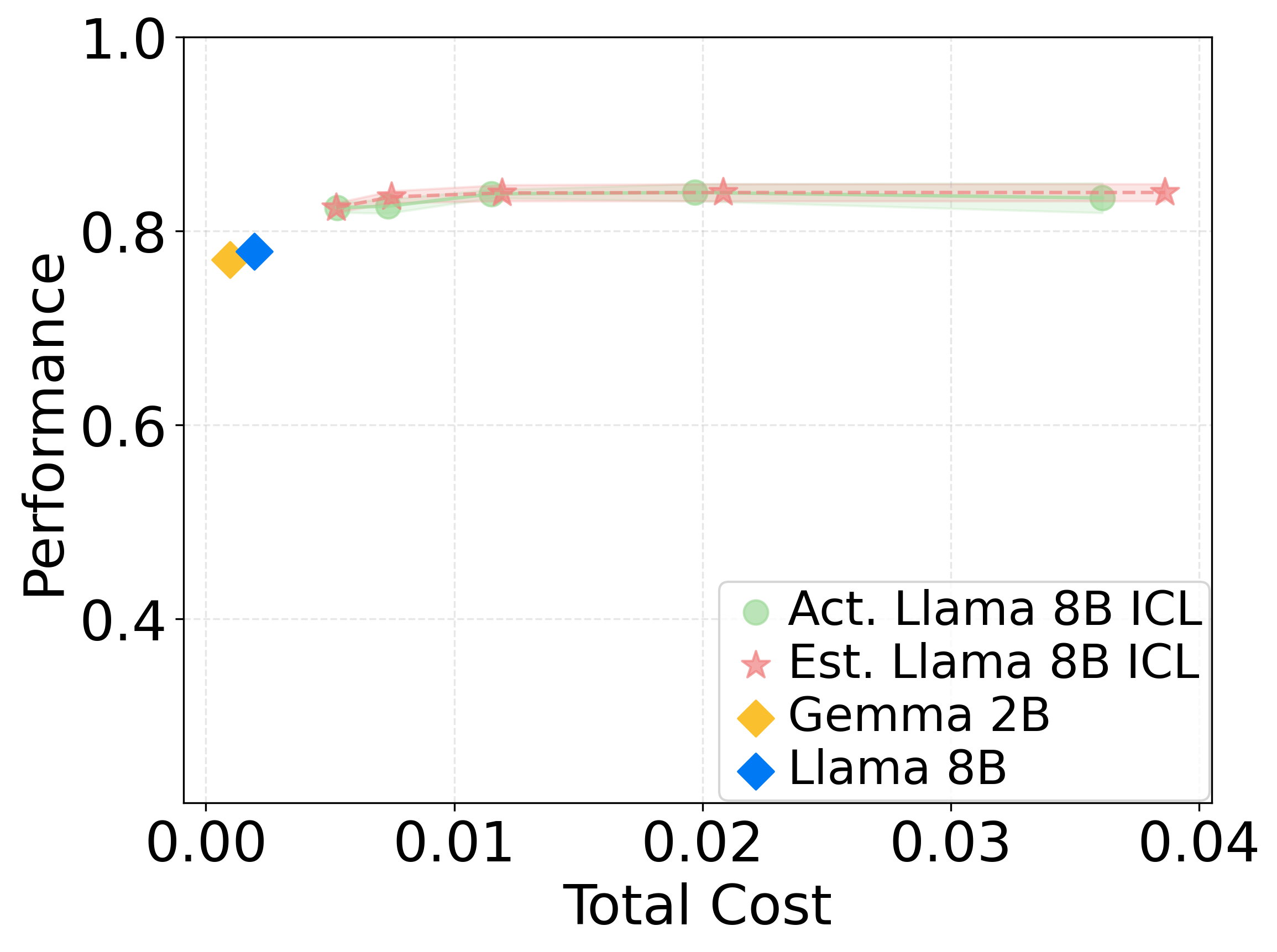}
\caption{\fiqasa}
\end{subfigure}
\hfill
\begin{subfigure}[t]{0.24\textwidth}
\includegraphics[width=\textwidth]{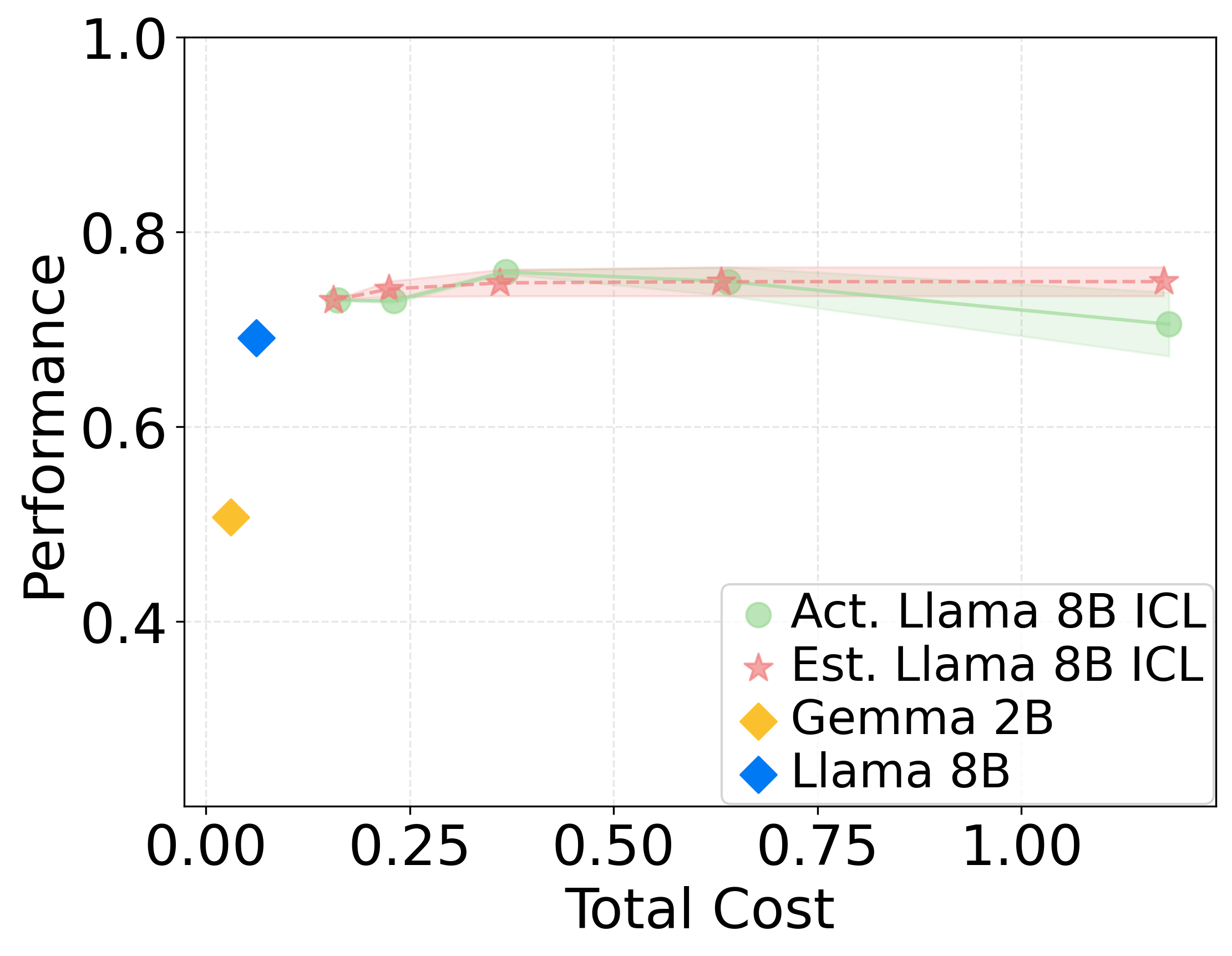}
\caption{\headline}
\end{subfigure}
\hfill
\begin{subfigure}[t]{0.24\textwidth}
\includegraphics[width=\textwidth]{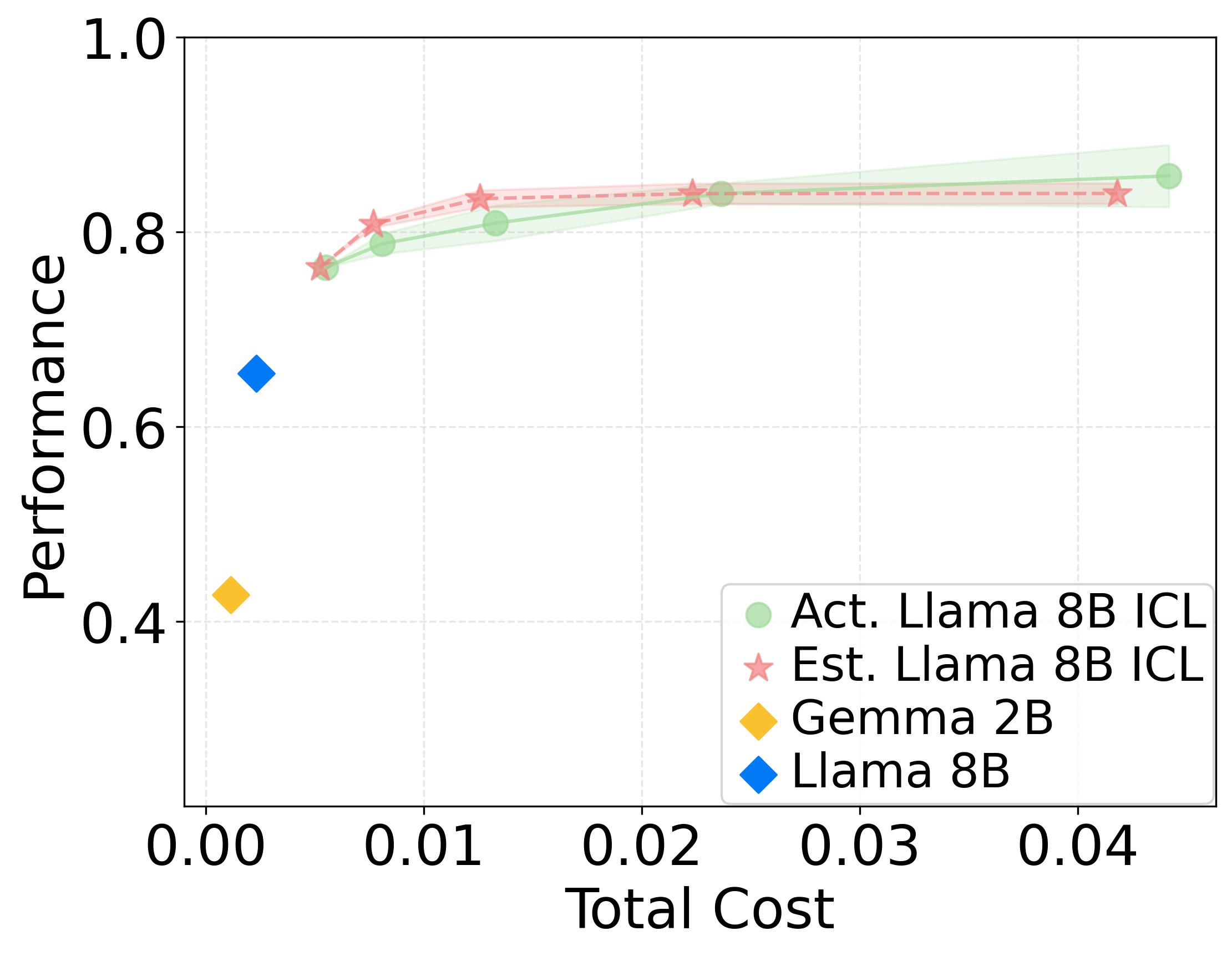}
\caption{\multifin}
\end{subfigure}
\caption{Predicted vs. actual performance-cost analysis for retrieval-based \icl across eight diverse tasks. Each plot compares actual (\textcolor{actualIclColor}{$\bullet$}) vs. predicted (\textcolor{predictedColor}{$\star$}) performance-cost trajectories for \llama \icl.
Base models Gemma 2B ($\gemmadiamond$), and Llama 8B ($\llamadiamond$) serve as reference points.
The consistent alignment between predicted and actual curves across both general and domain-specific tasks demonstrates \methodabb's robust prediction capabilities.}
\label{fig:detailed_icl_perf_cost_full}
\end{figure*}

Following our analysis in Section~\ref{sec5.2:detailed-pred-analysis}, we present comprehensive performance--cost trajectories for all eight tasks in Figure~\ref{fig:detailed_qlora_perf_cost_full} and~\ref{fig:detailed_icl_perf_cost_full}, examining \qlora fine-tuning and retrieval-augmented \icl, respectively.
The strong alignment between predicted and actual performance trajectories across all tasks and strategies validates our method's robustness. Our framework demonstrates particular strength in capturing complex performance dynamics--not only predicting standard improvement curves, but also accurately forecasting non-monotonic patterns, such as the performance degradation observed in the \arc task as the computational budget increases.
This ability to capture both positive and negative performance trends further substantiates the generalizability of our prediction framework.

\subsection{A Closer Look at \methodabb's Prediction Ability for Fine-tuning}\label{app:cap-zoom-in-qlora}

\begin{figure*}[h]
\begin{subfigure}[t]{0.24\textwidth}
\includegraphics[width=\textwidth]{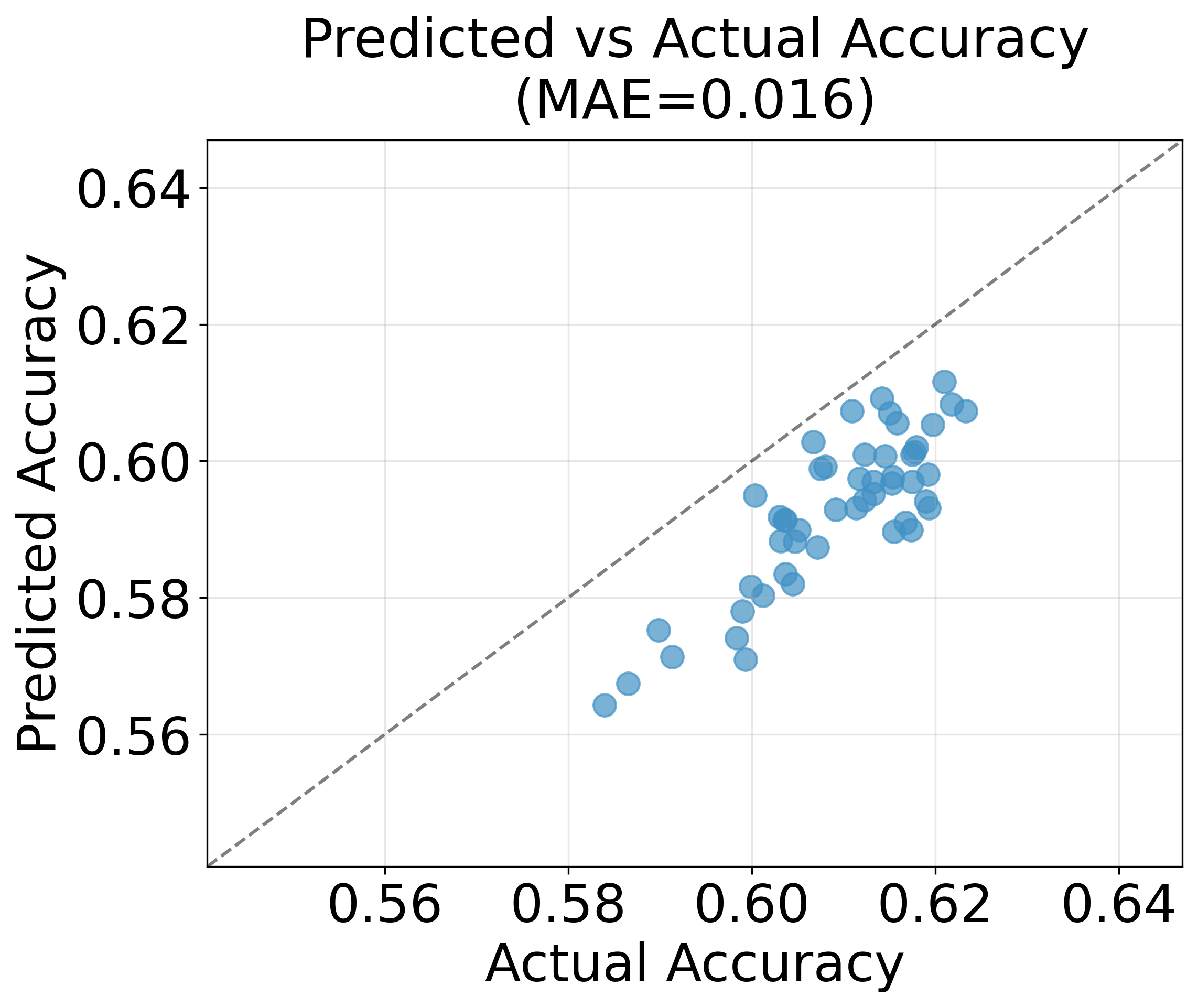}
\caption{\mmlu}
\end{subfigure}
\hfill
\begin{subfigure}[t]{0.24\textwidth}
\includegraphics[width=\textwidth]{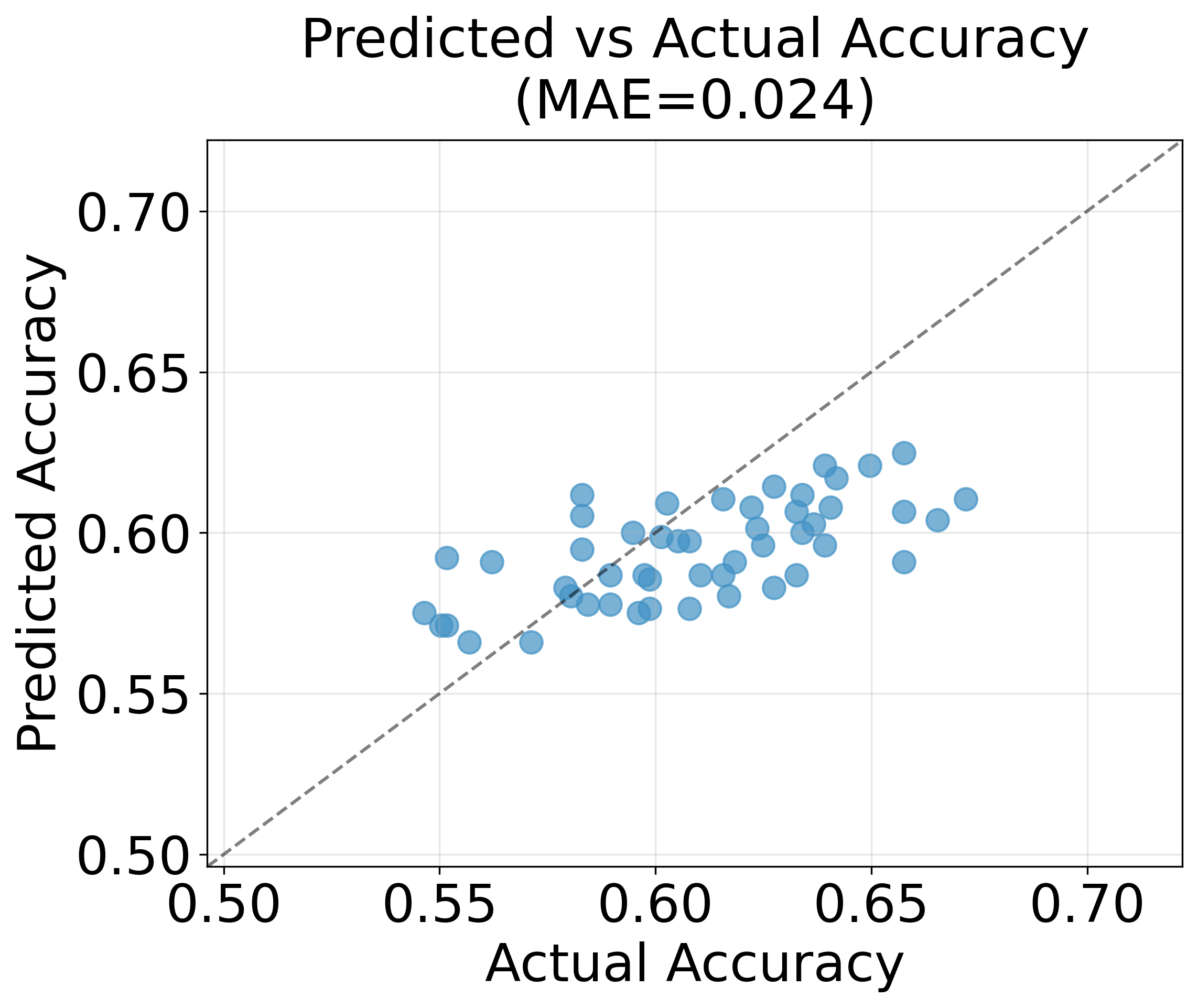}
\caption{\wino}
\end{subfigure}
\hfill
\begin{subfigure}[t]{0.24\textwidth}
\includegraphics[width=\textwidth]{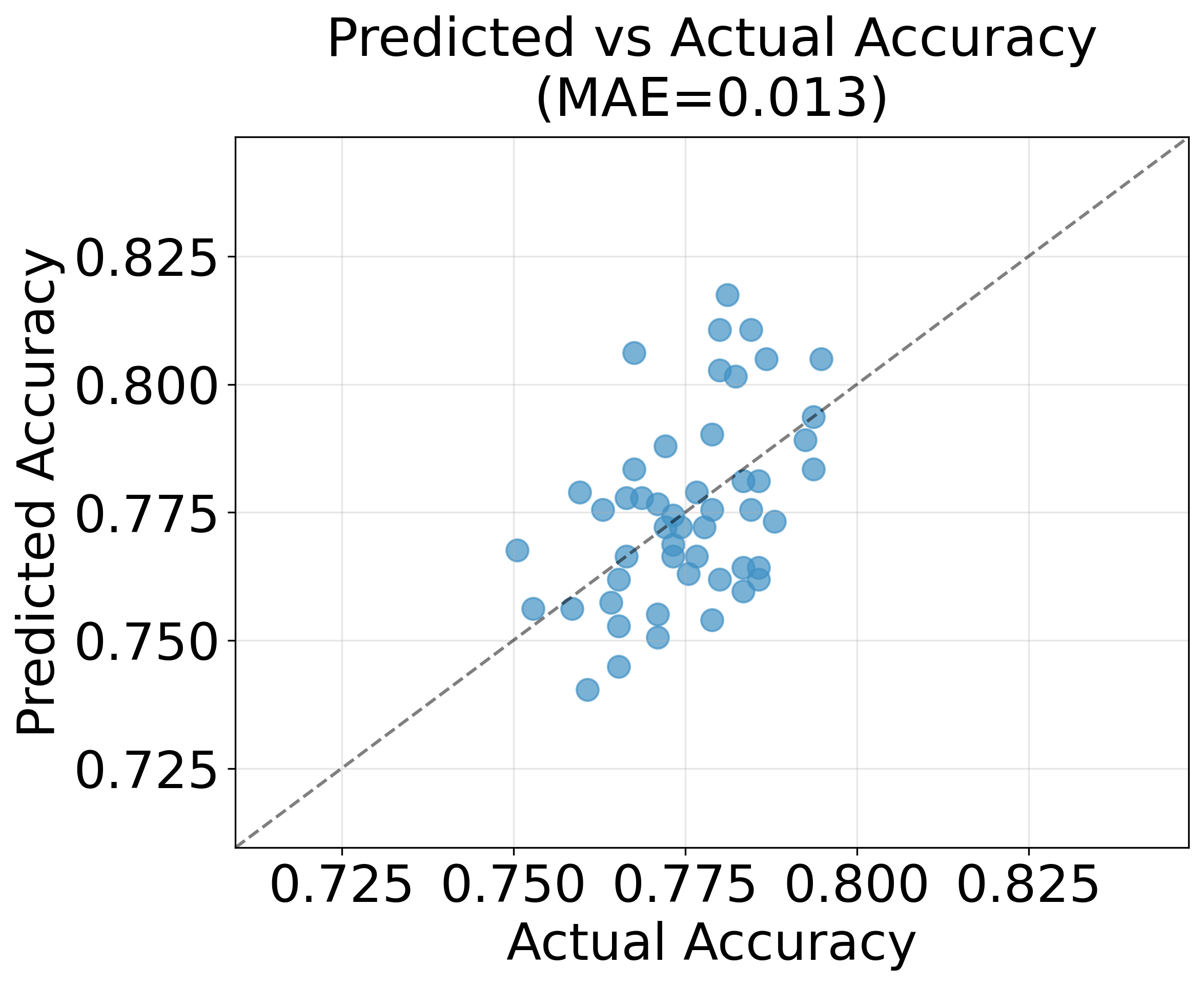}
\caption{\arc}
\end{subfigure}
\hfill
\begin{subfigure}[t]{0.24\textwidth}
\includegraphics[width=\textwidth]{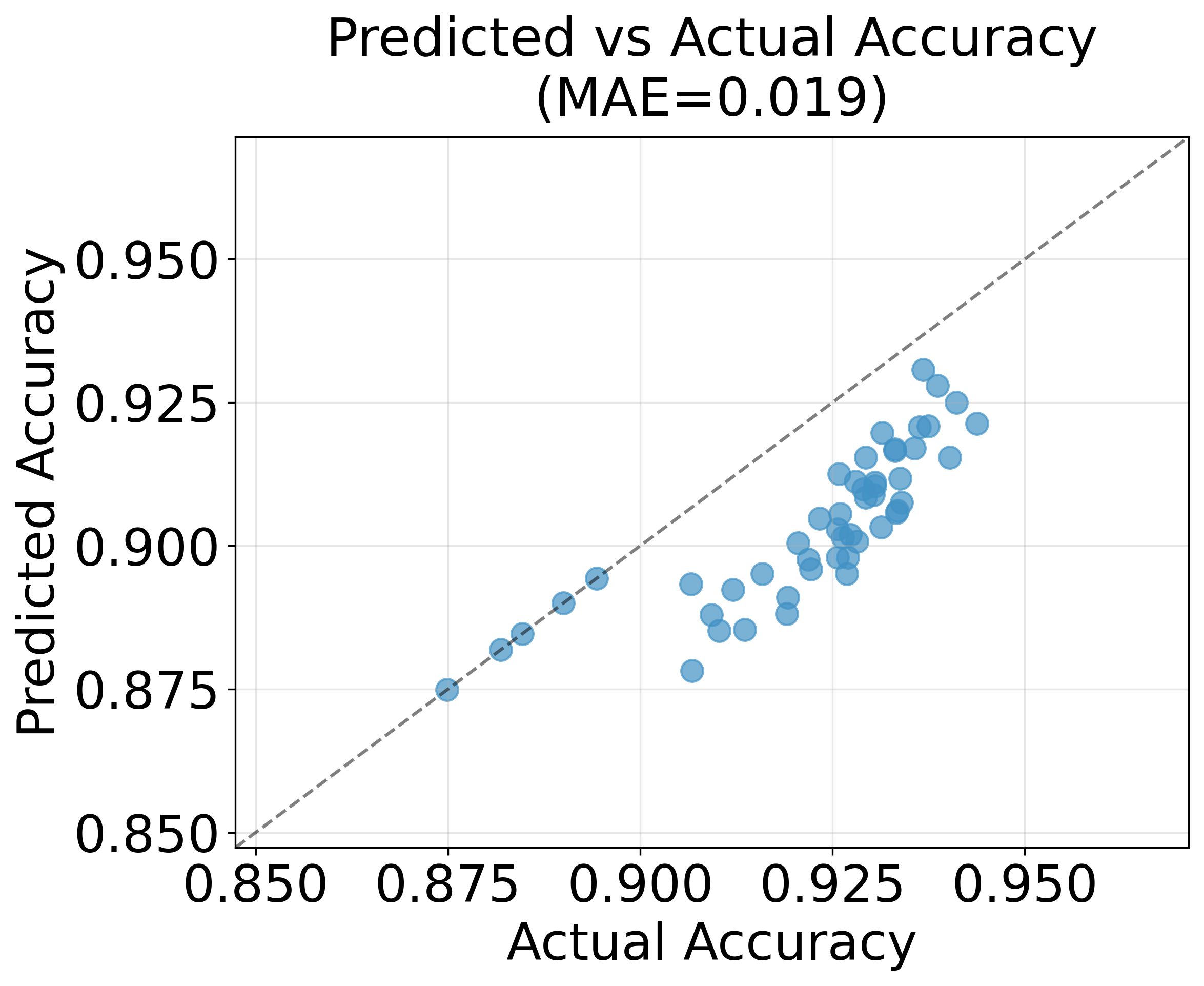}
\caption{\hella}
\end{subfigure}
\\[1em]
\begin{subfigure}[t]{0.24\textwidth}
\includegraphics[width=\textwidth]{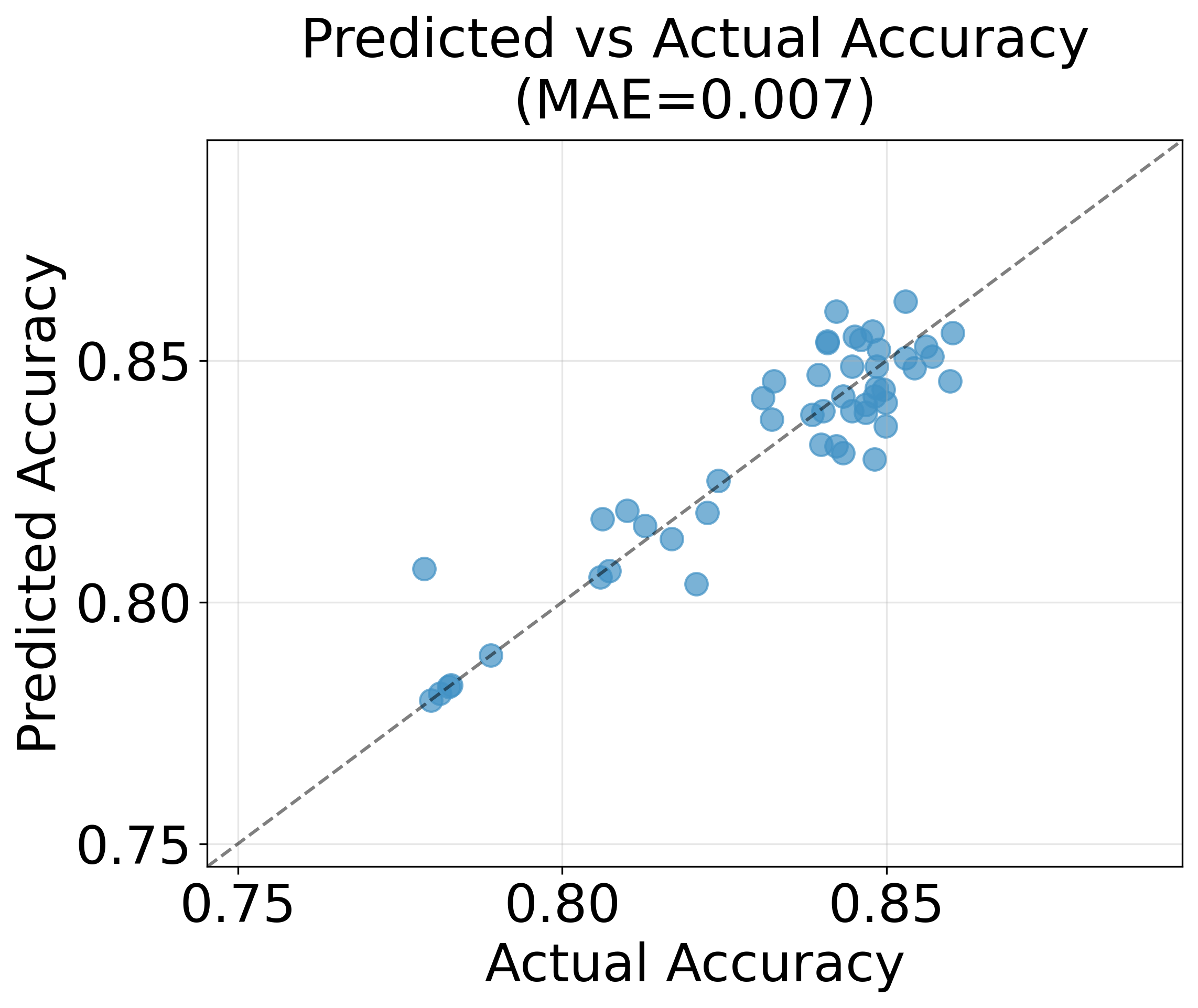}
\caption{\fpb}
\end{subfigure}
\hfill
\begin{subfigure}[t]{0.24\textwidth}
\includegraphics[width=\textwidth]{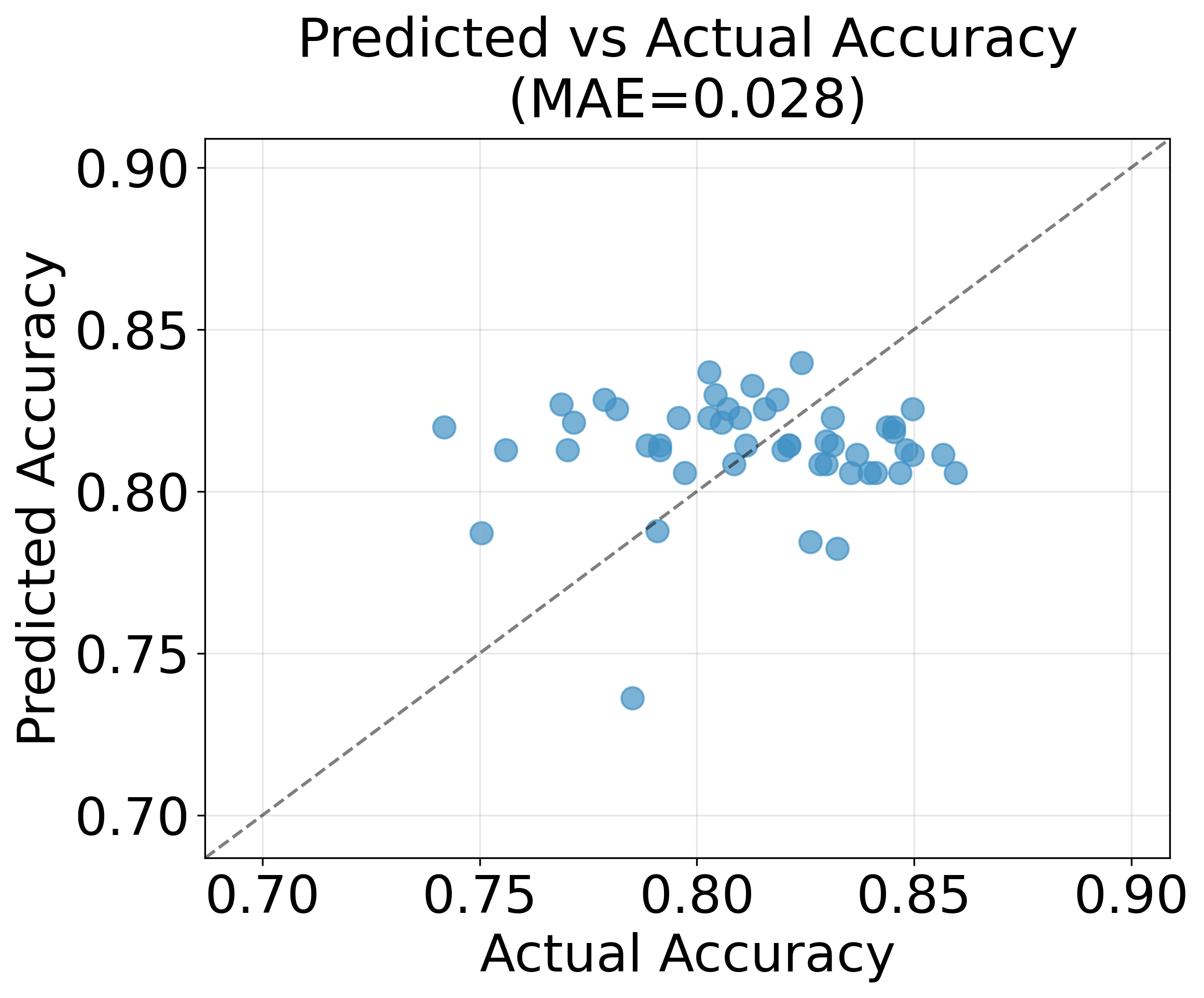}
\caption{\fiqasa}
\end{subfigure}
\hfill
\begin{subfigure}[t]{0.24\textwidth}
\includegraphics[width=\textwidth]{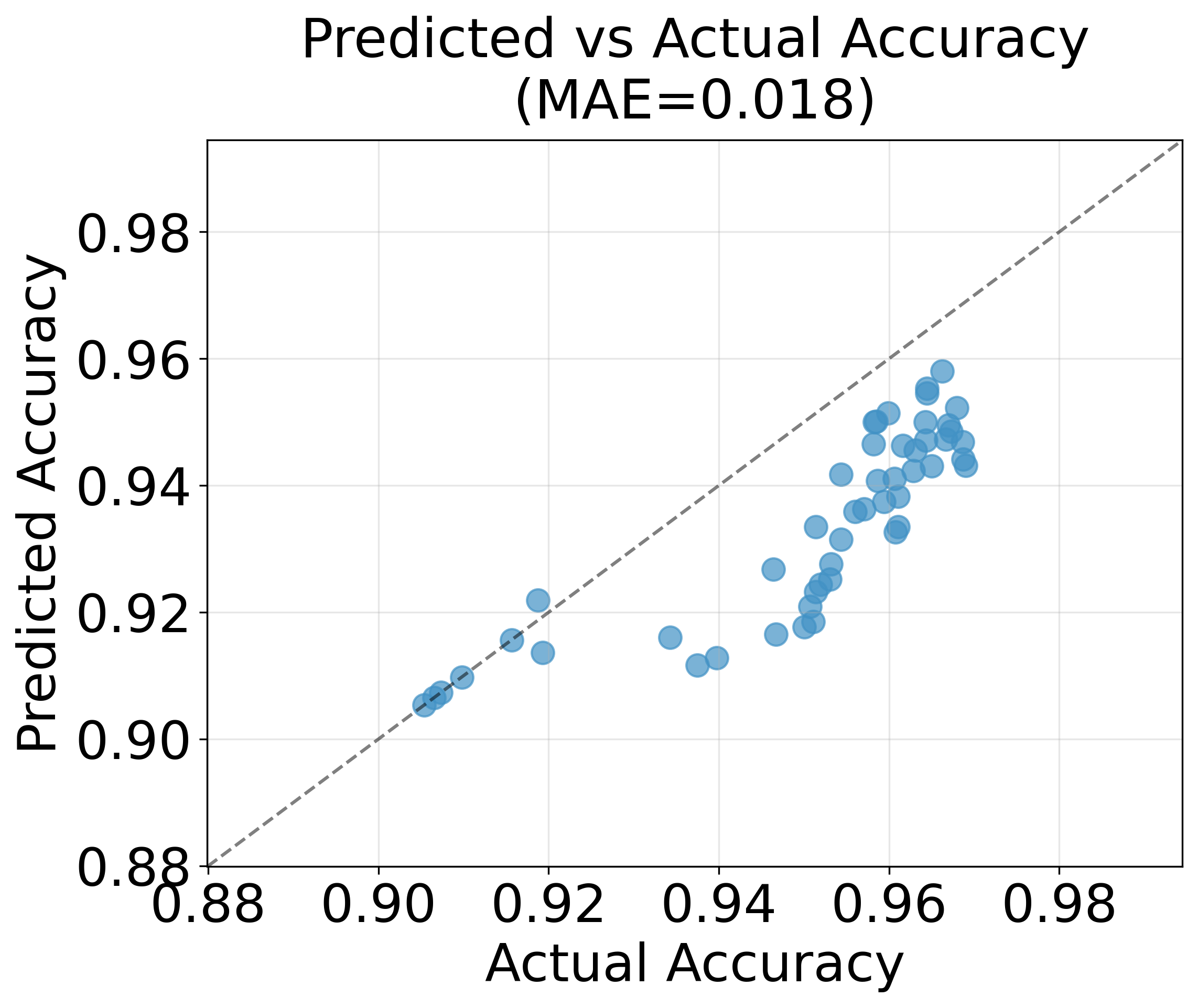}
\caption{\headline}
\end{subfigure}
\hfill
\begin{subfigure}[t]{0.24\textwidth}
\includegraphics[width=\textwidth]{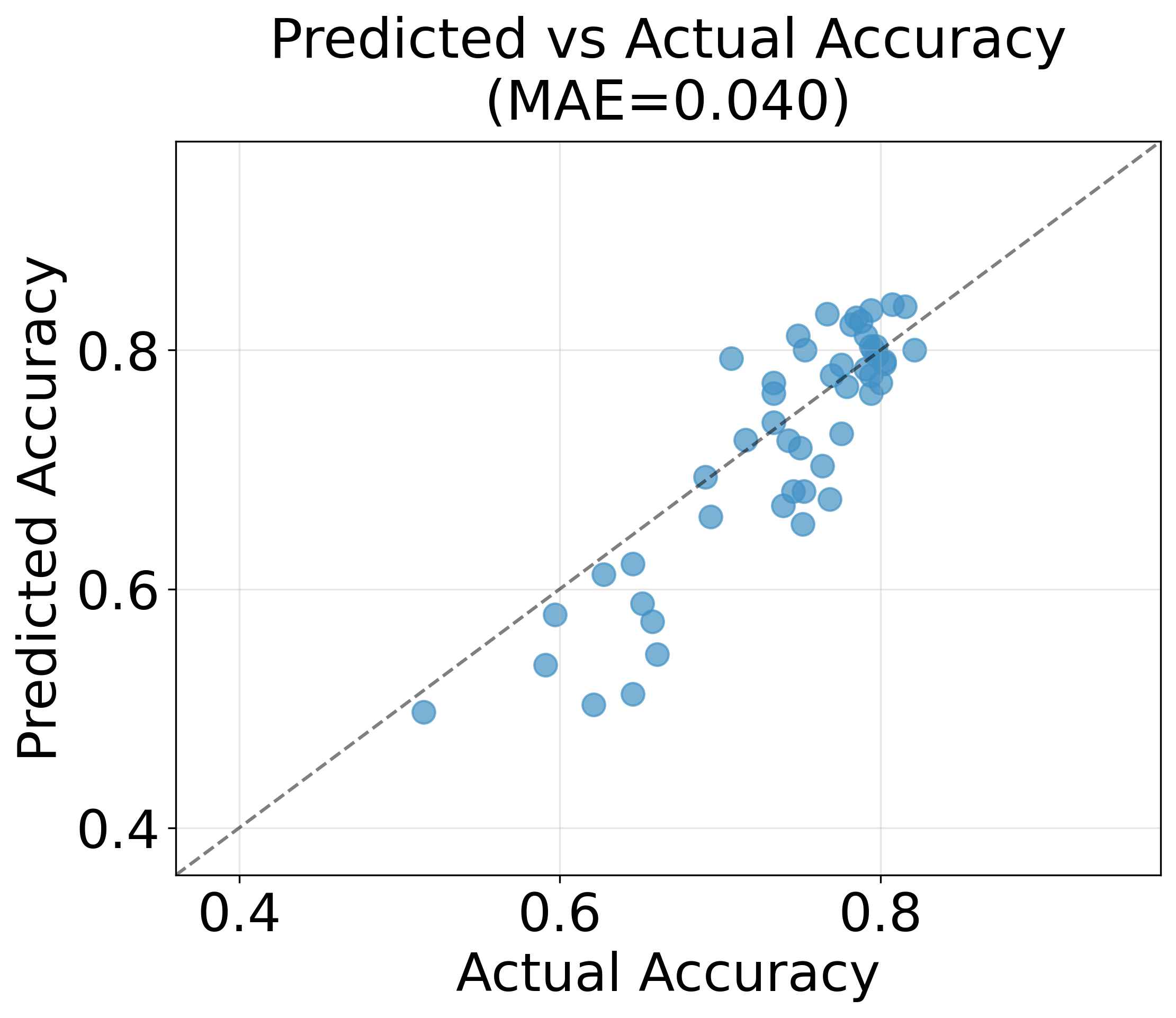}
\caption{\multifin}
\end{subfigure}
\caption{Scatter plots comparing predicted vs. actual performance (accuracy) for \qlora fine-tuning across eight diverse tasks Axes are strategically zoomed to reveal fine-grained prediction details, with points closer to the diagonal indicating higher prediction accuracy. For example, in \headline, a \textcolor{scatterColor}{$\bullet$} at (0.92, 0.915) shows that for a specific configuration (training data size and iteration count), \methodabb predicts 0.915 accuracy while the actual performance achieves 0.92. 
The tight clustering around the diagonal across both general domain (a-d) and financial domain (e-h) tasks demonstrates \methodabb's robust prediction capabilities.}
\label{fig:detailed_qlora_perf}
\end{figure*}

\paragraph{Performance prediction.}
Figure~\ref{fig:detailed_qlora_perf} provides a detailed analysis of \methodabb's prediction accuracy on fine-tuning across eight tasks. For each task, we plot predicted versus actual performance with axes deliberately zoomed to highlight prediction granularity. Each scatter point represents a specific \qlora fine-tuning configuration, with proximity to the diagonal indicating prediction accuracy. The Mean Absolute Error (MAE) ranges from 0.007 (FPB) to 0.040 (Multifin EN), with most tasks showing MAE below 0.02, demonstrating remarkable precision. 
The framework exhibits consistent performance across both general-domain benchmarks (MMLU: 0.016, Winogrande: 0.024, ARC-Challenge: 0.013, HellaSwag: 0.019) and financial tasks (FPB: 0.007, \fiqasa: 0.028, Headline: 0.018, Multifin EN: 0.040). 
The tight clustering around the diagonal, particularly evident in the zoomed visualization, underscores our method's robust predictive capabilities regardless of task domain or performance level. 

\begin{figure*}[h]
\begin{subfigure}[t]{0.24\textwidth}
\includegraphics[width=\textwidth]{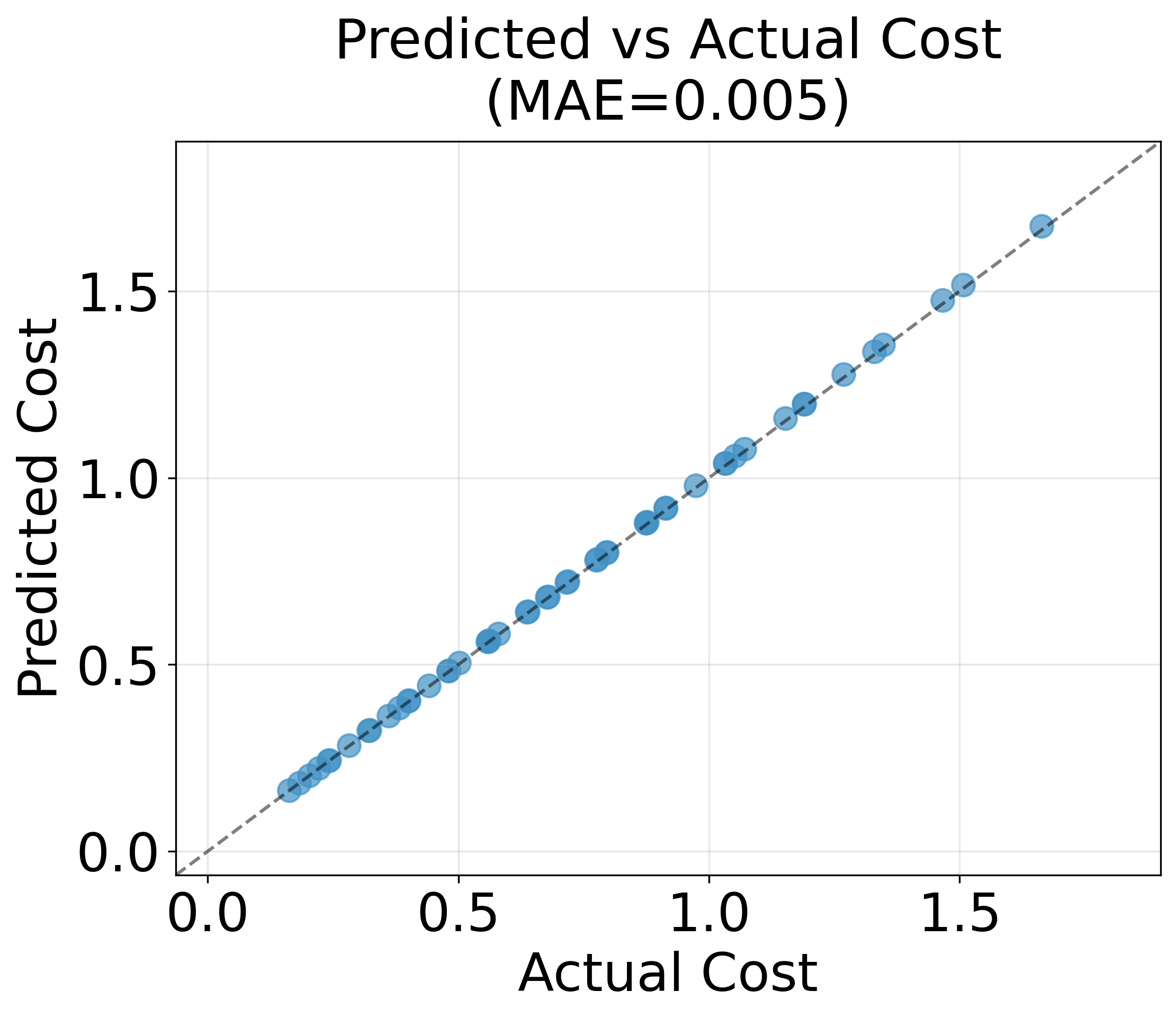}
\caption{\mmlu}
\end{subfigure}
\hfill
\begin{subfigure}[t]{0.24\textwidth}
\includegraphics[width=\textwidth]{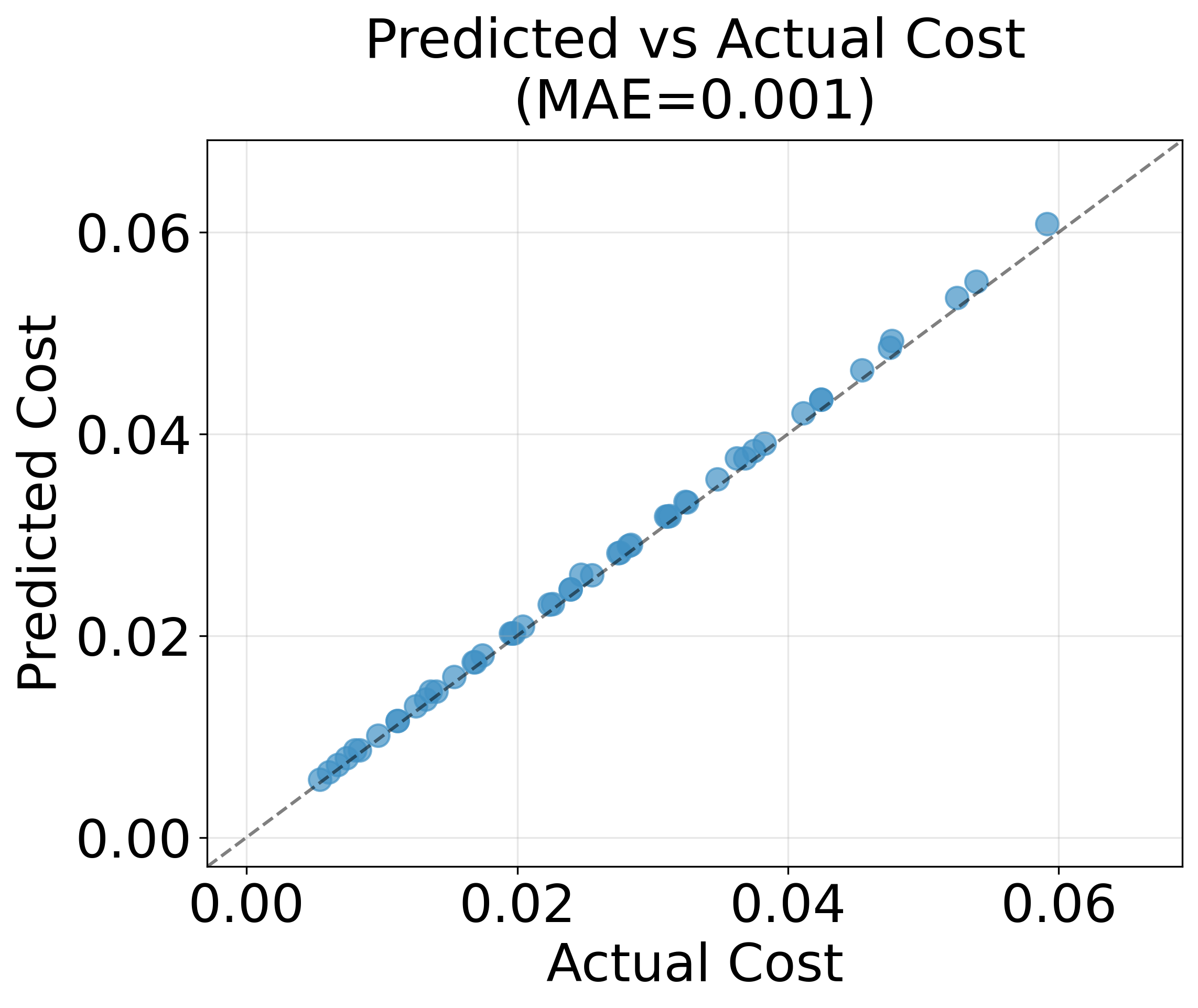}
\caption{\wino}
\end{subfigure}
\hfill
\begin{subfigure}[t]{0.24\textwidth}
\includegraphics[width=\textwidth]{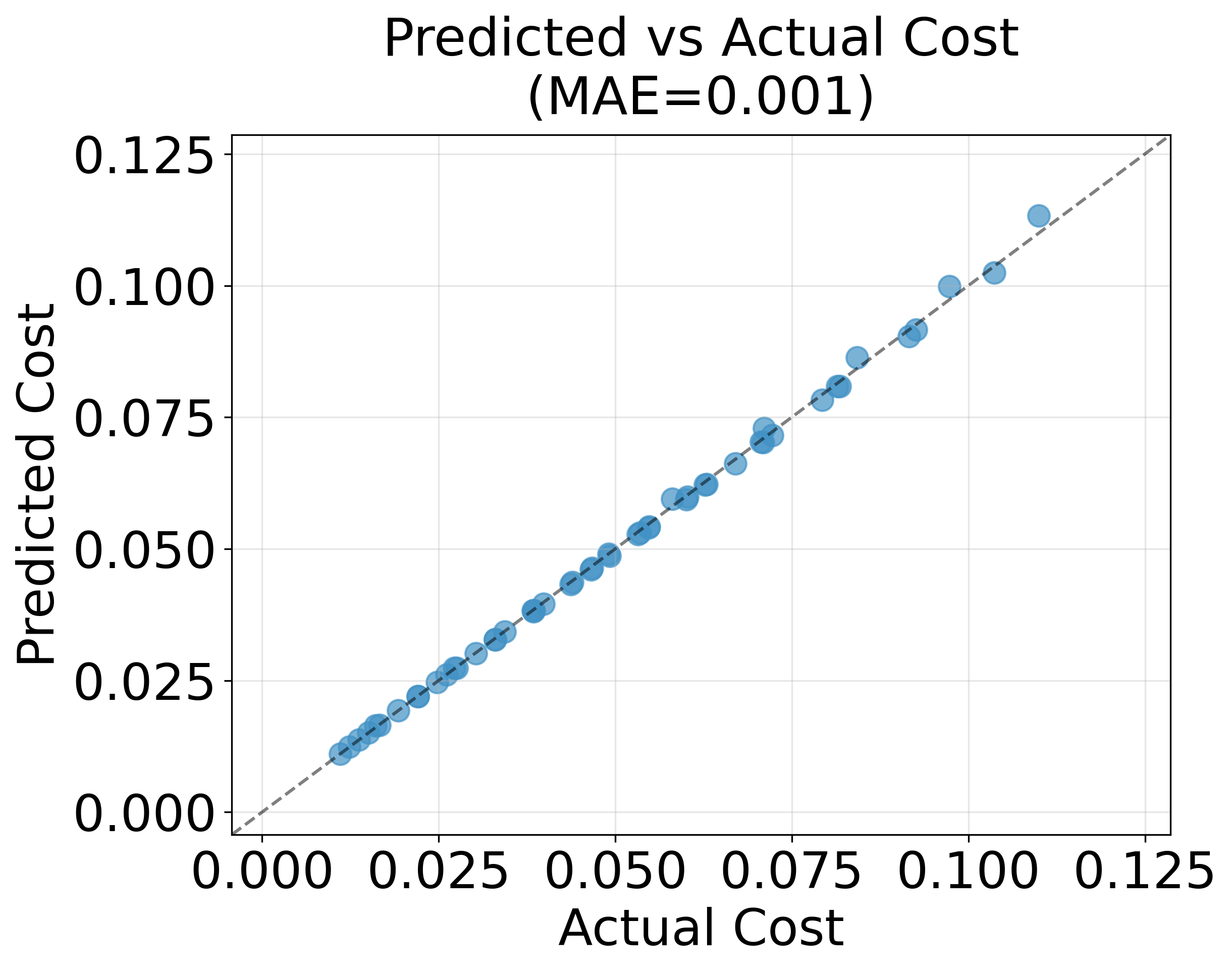}
\caption{\arc}
\end{subfigure}
\hfill
\begin{subfigure}[t]{0.24\textwidth}
\includegraphics[width=\textwidth]{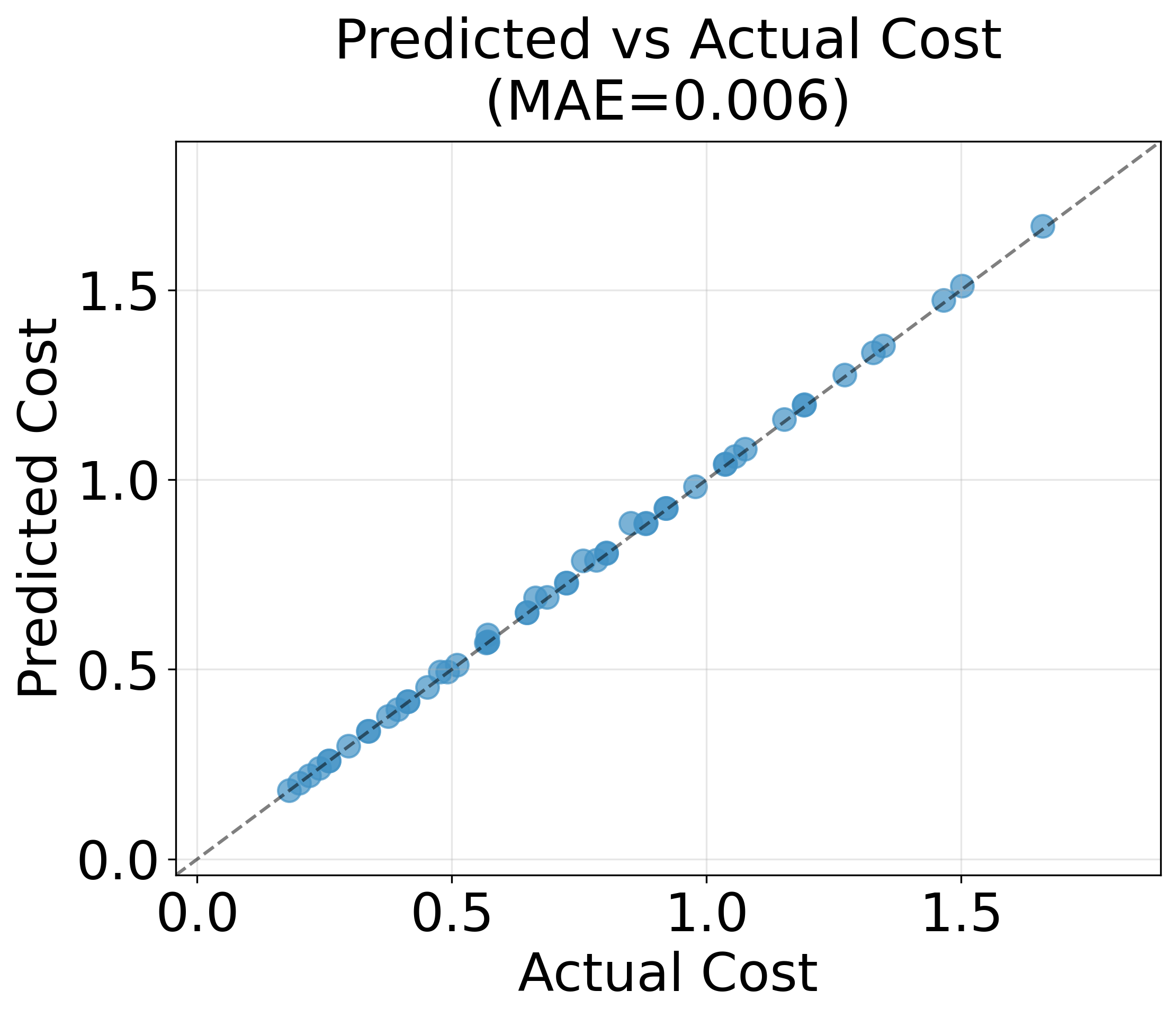}
\caption{\hella}
\end{subfigure}
\\[1em]
\begin{subfigure}[t]{0.24\textwidth}
\includegraphics[width=\textwidth]{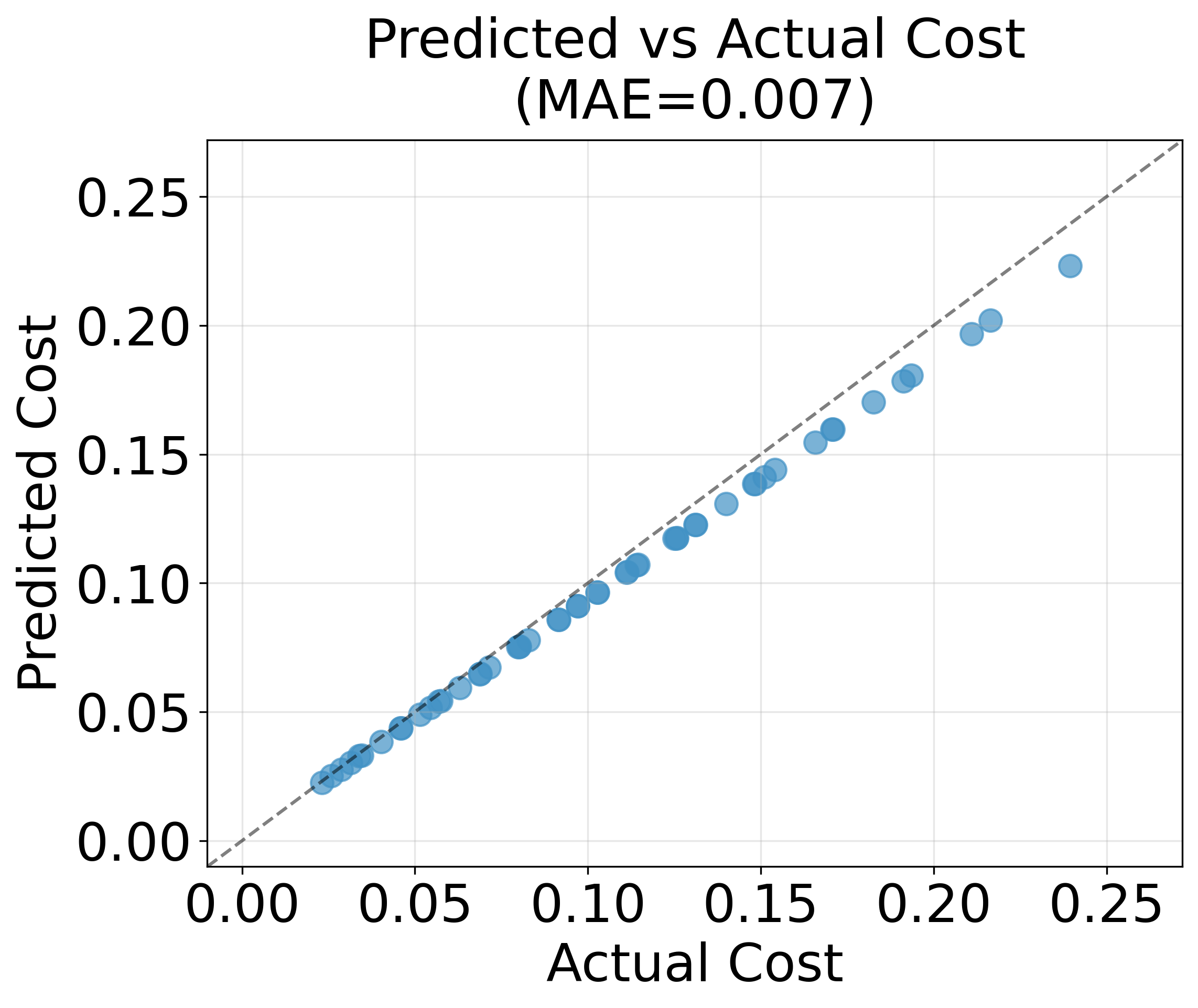}
\caption{\fpb}
\end{subfigure}
\hfill
\begin{subfigure}[t]{0.24\textwidth}
\includegraphics[width=\textwidth]{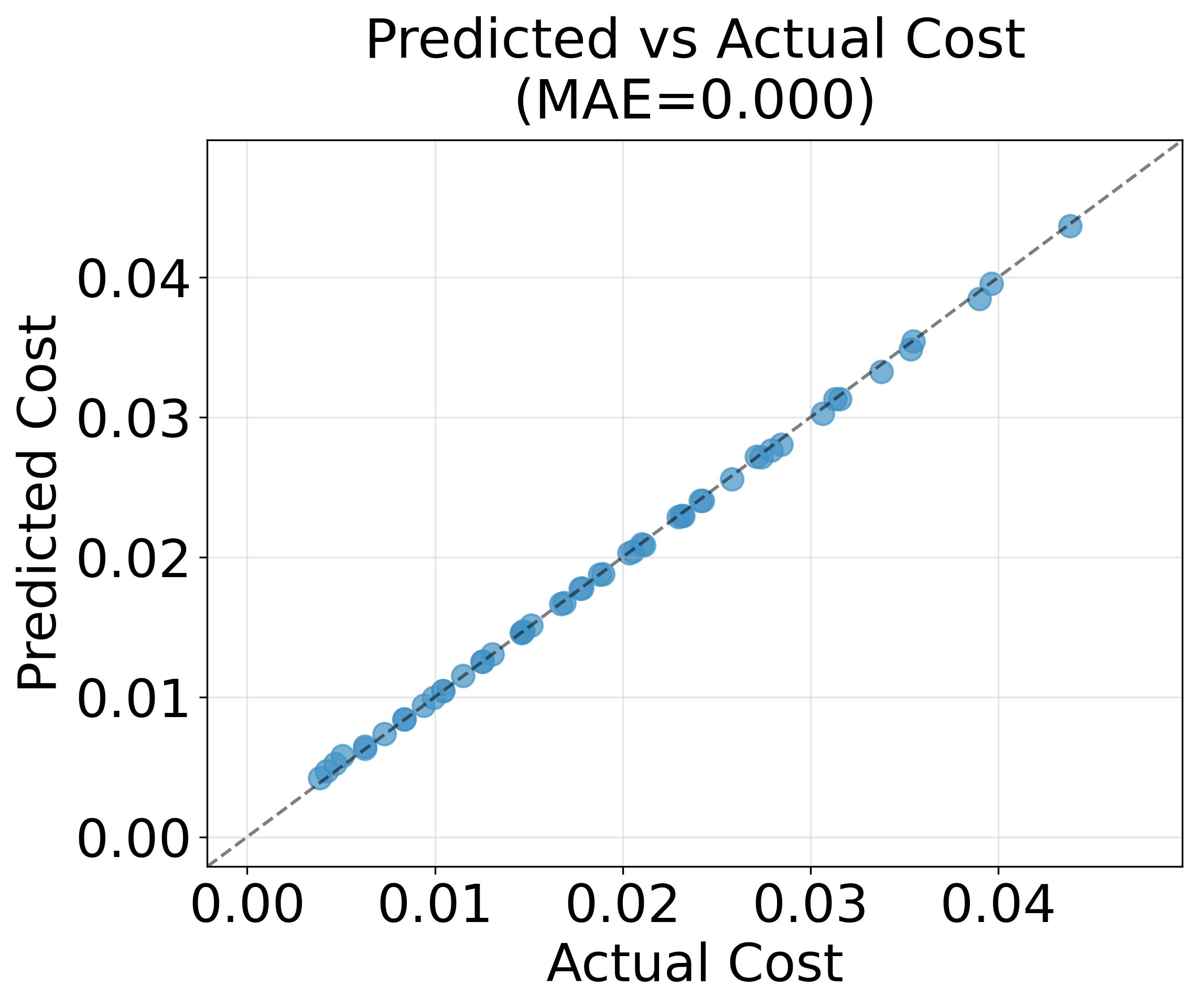}
\caption{\fiqasa}
\end{subfigure}
\hfill
\begin{subfigure}[t]{0.24\textwidth}
\includegraphics[width=\textwidth]{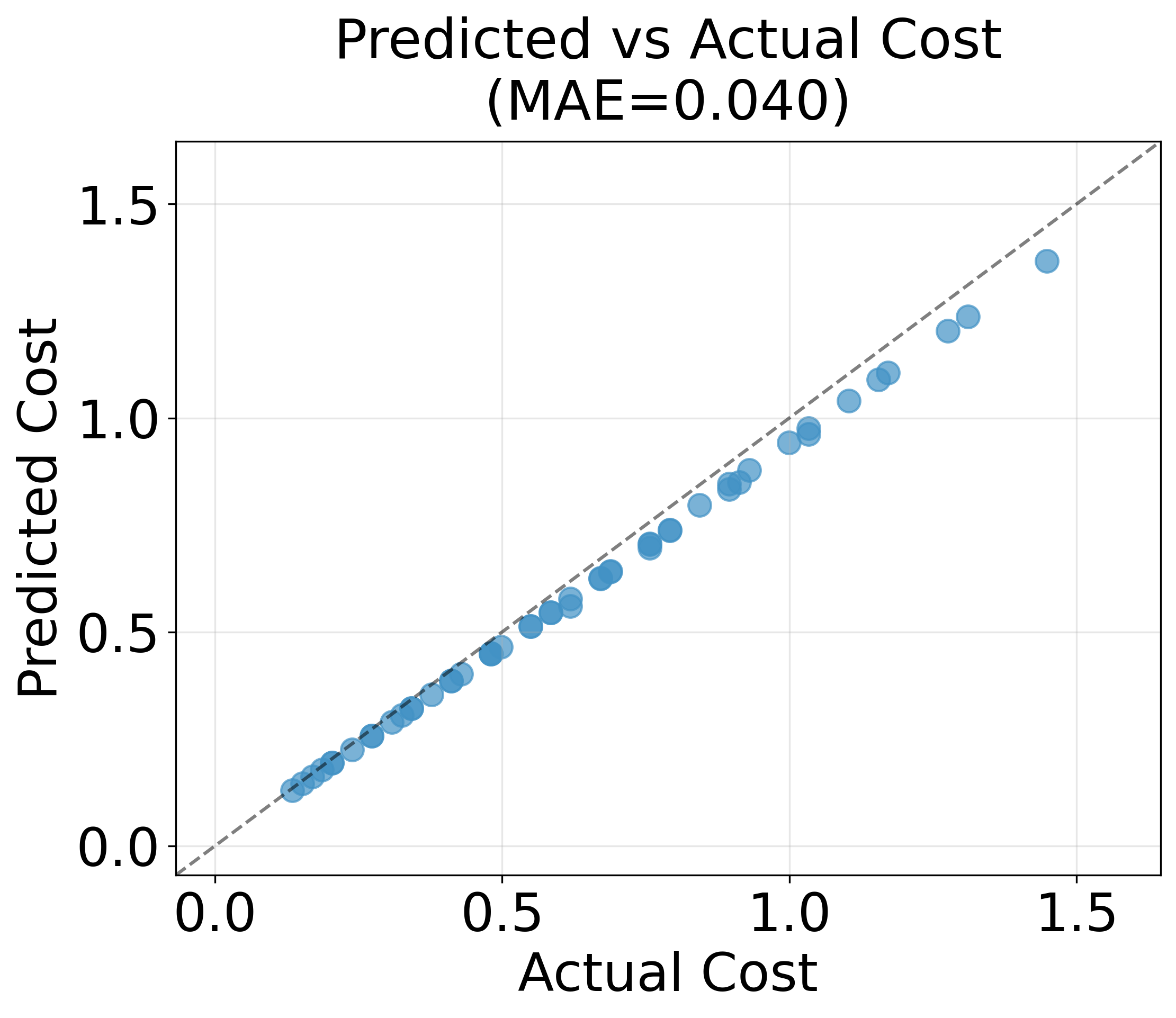}
\caption{\headline}
\end{subfigure}
\hfill
\begin{subfigure}[t]{0.24\textwidth}
\includegraphics[width=\textwidth]{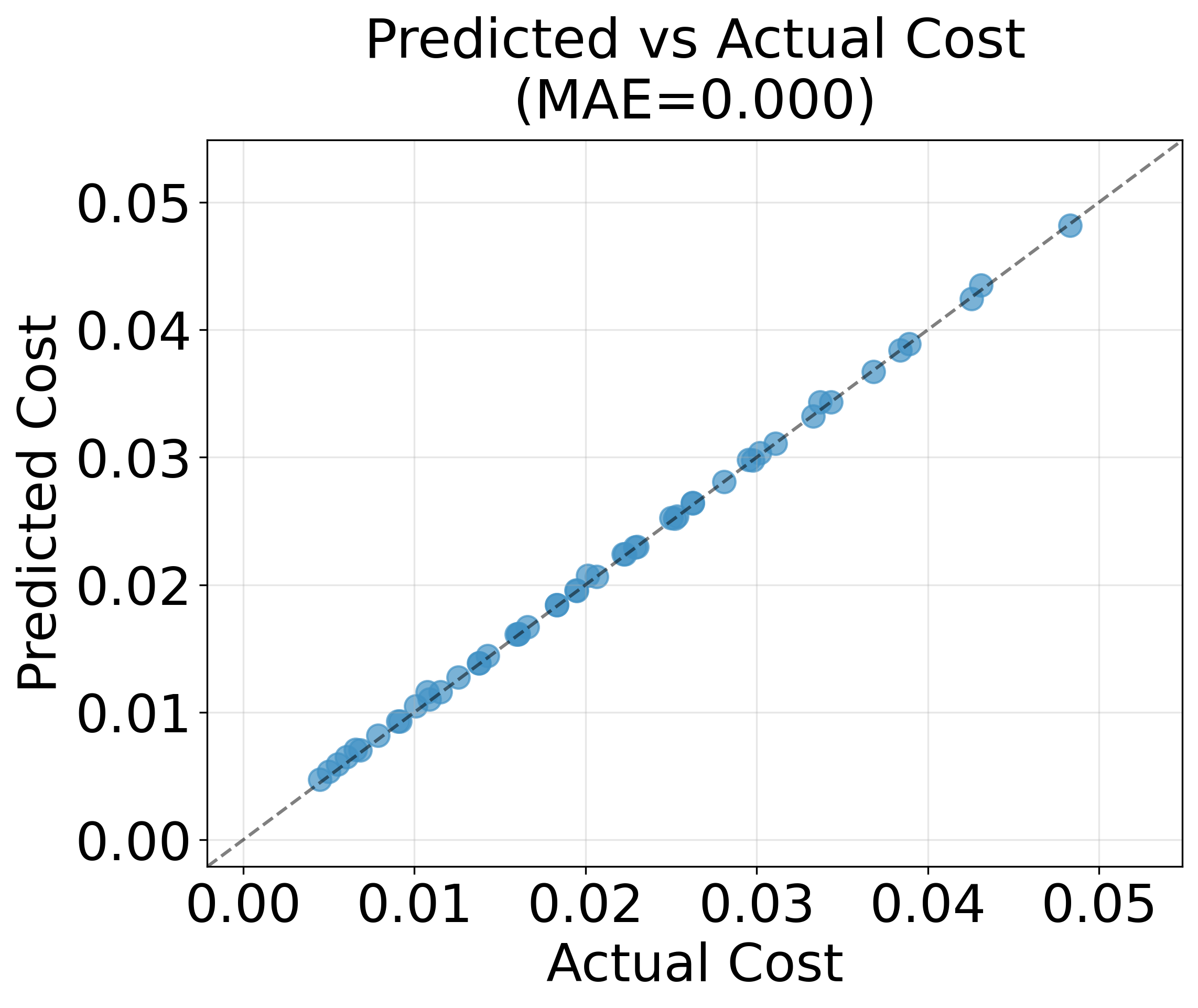}
\caption{\multifin}
\end{subfigure}
\caption{Scatter plots of predicted vs. actual cost for \qlora fine-tuning across eight diverse tasks. The near-perfect diagonal alignment and low MAE values (0.000-0.007) demonstrate precise cost prediction capabilities across both resource-intensive (a-d) and lightweight tasks (e-h).}
\label{fig:detailed_qlora_cost}
\end{figure*}

\paragraph{Cost prediction.}
Figure~\ref{fig:detailed_qlora_cost} demonstrates our framework's cost prediction capabilities for \qlora fine-tuning across eight tasks. The scatter plots reveal near-perfect diagonal alignment with remarkably low MAE (0.000-0.007) across all tasks, from resource-intensive benchmarks like MMLU to lightweight tasks like \fiqasa. This consistency across varying cost scales validates our framework's robust cost estimation capabilities.

\begin{figure*}[h]
\begin{subfigure}[t]{0.24\textwidth}
\includegraphics[width=\textwidth]{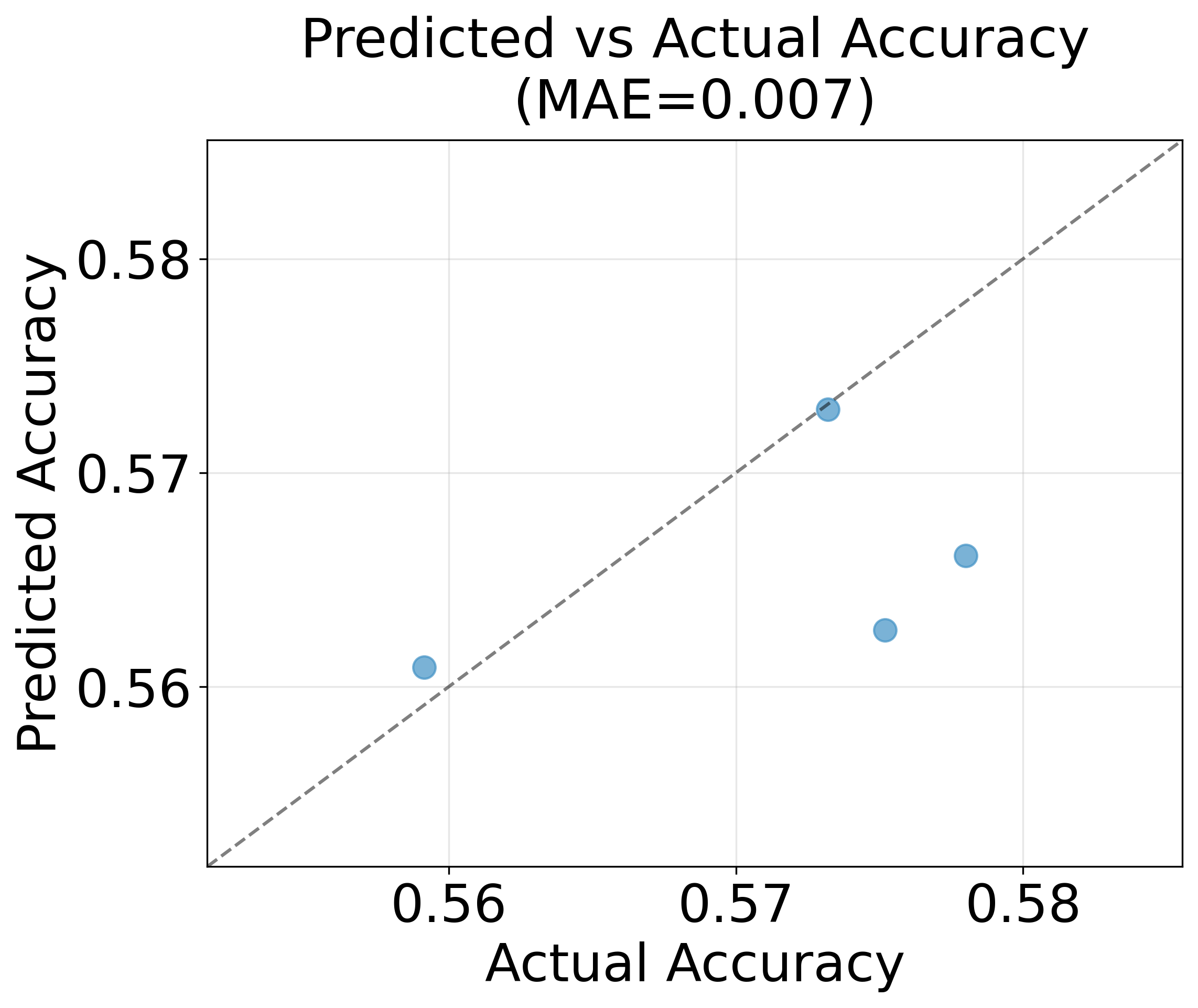}
\caption{\mmlu}
\end{subfigure}
\hfill
\begin{subfigure}[t]{0.24\textwidth}
\includegraphics[width=\textwidth]{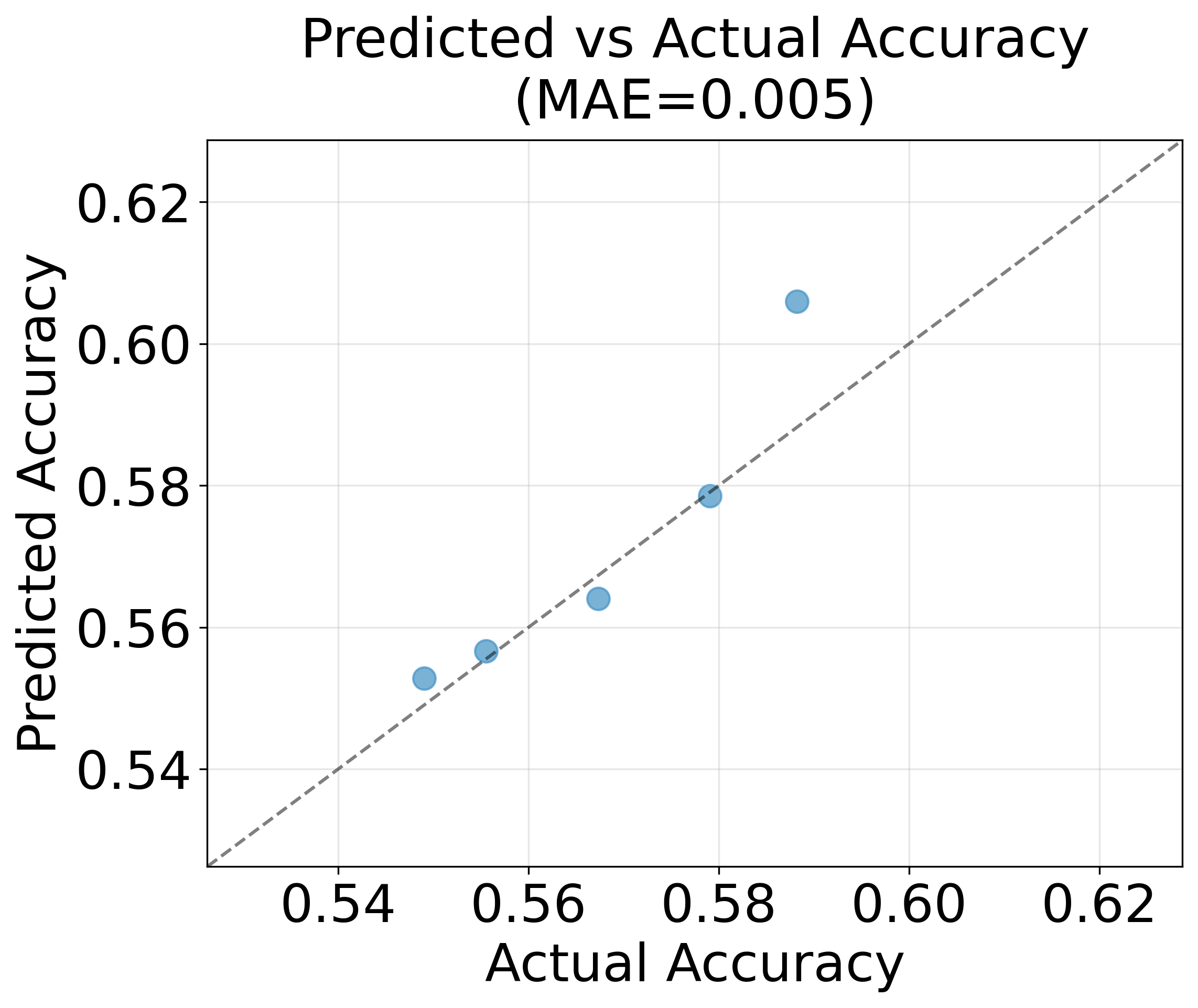}
\caption{\wino}
\end{subfigure}
\hfill
\begin{subfigure}[t]{0.24\textwidth}
\includegraphics[width=\textwidth]{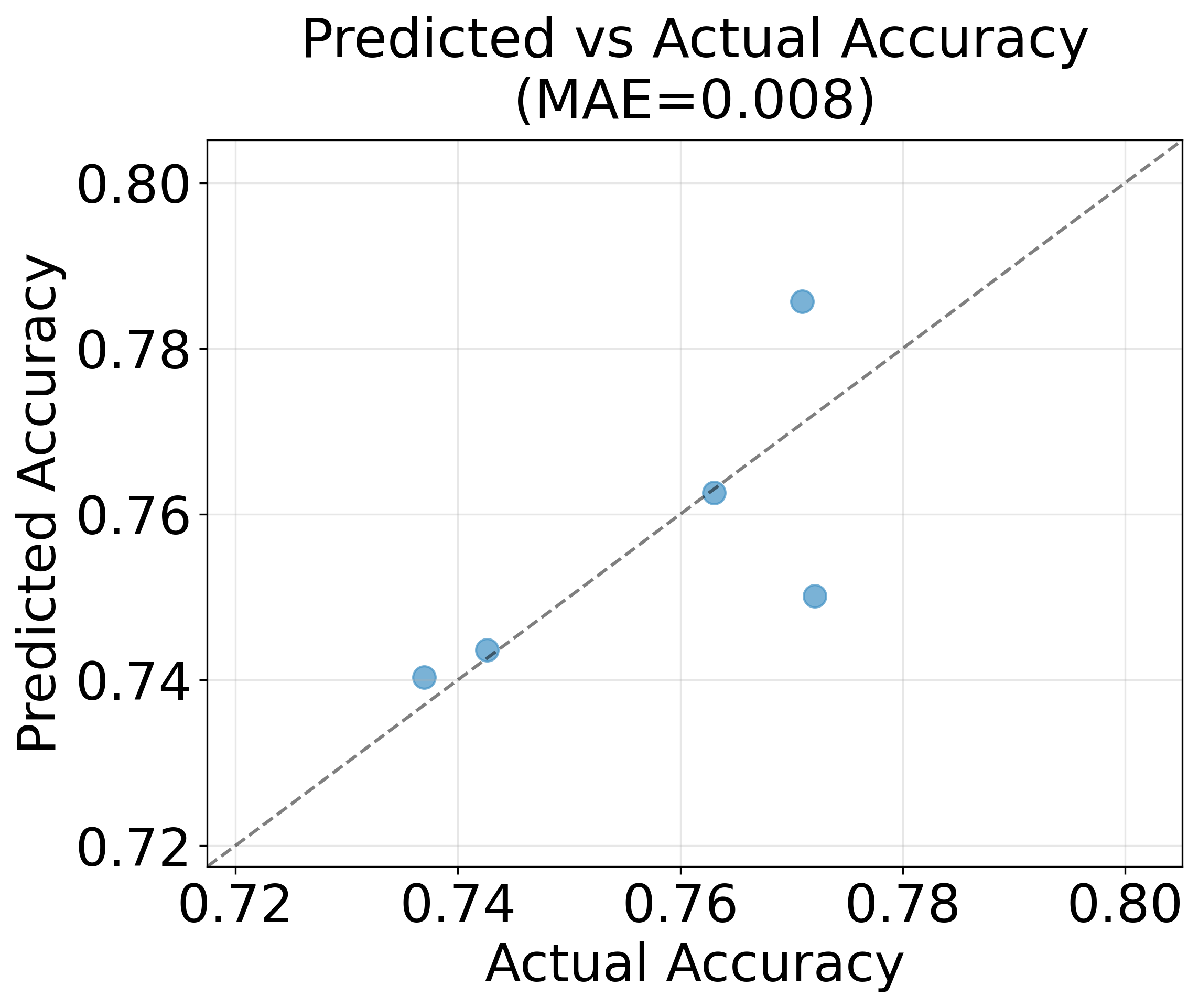}
\caption{\arc}
\end{subfigure}
\hfill
\begin{subfigure}[t]{0.24\textwidth}
\includegraphics[width=\textwidth]{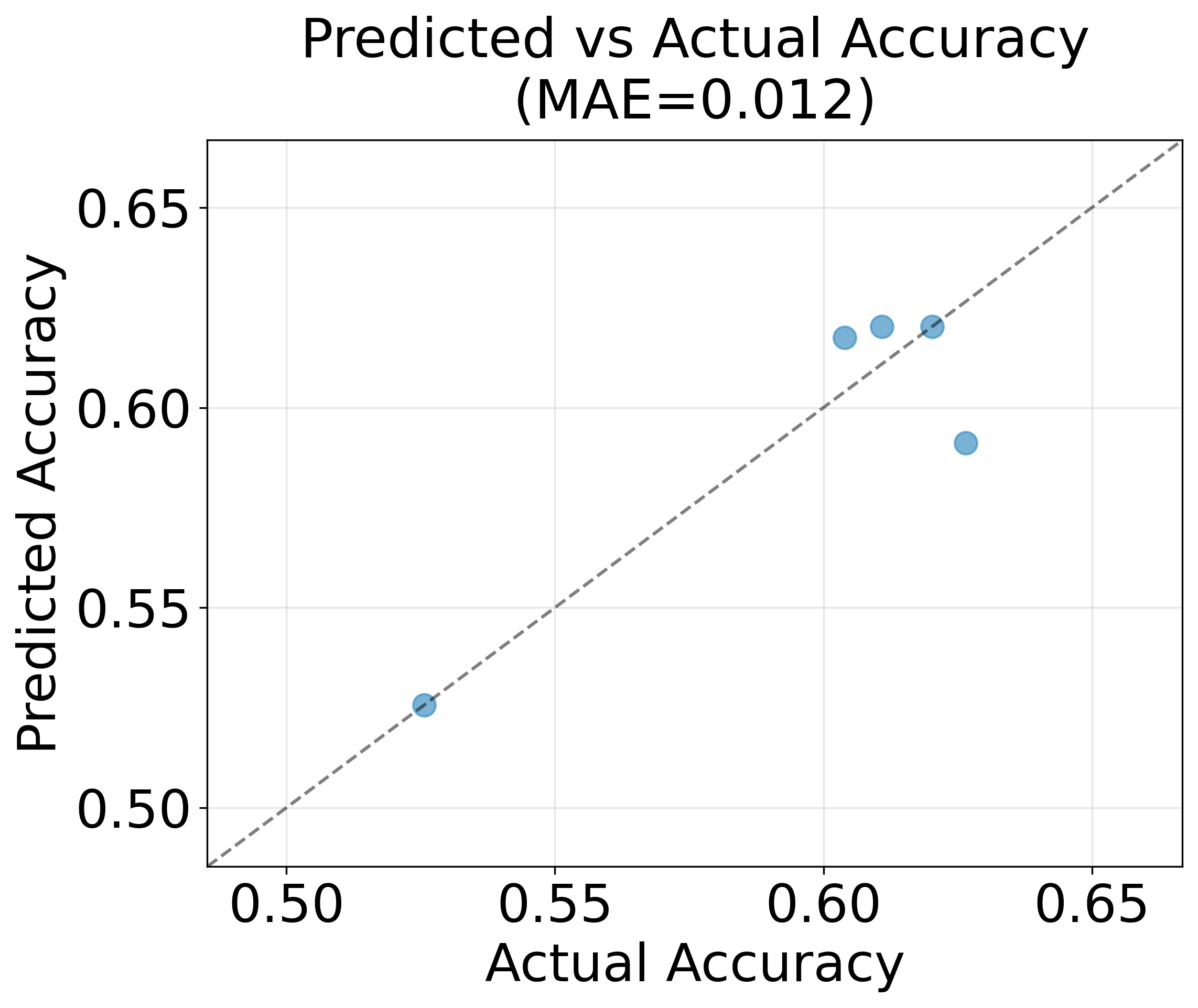}
\caption{\hella}
\end{subfigure}
\\[1em]
\begin{subfigure}[t]{0.24\textwidth}
\includegraphics[width=\textwidth]{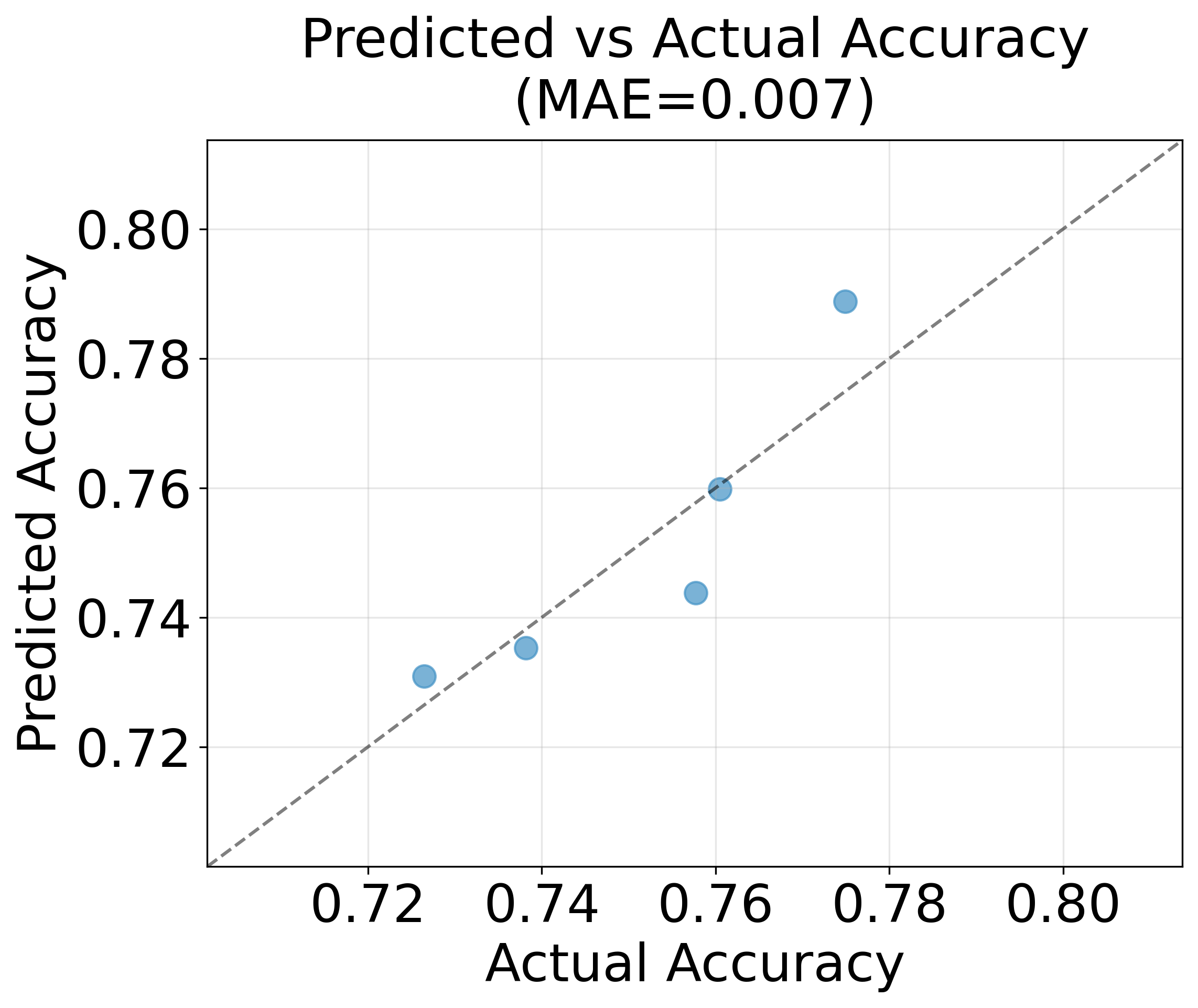}
\caption{\fpb}
\end{subfigure}
\hfill
\begin{subfigure}[t]{0.24\textwidth}
\includegraphics[width=\textwidth]{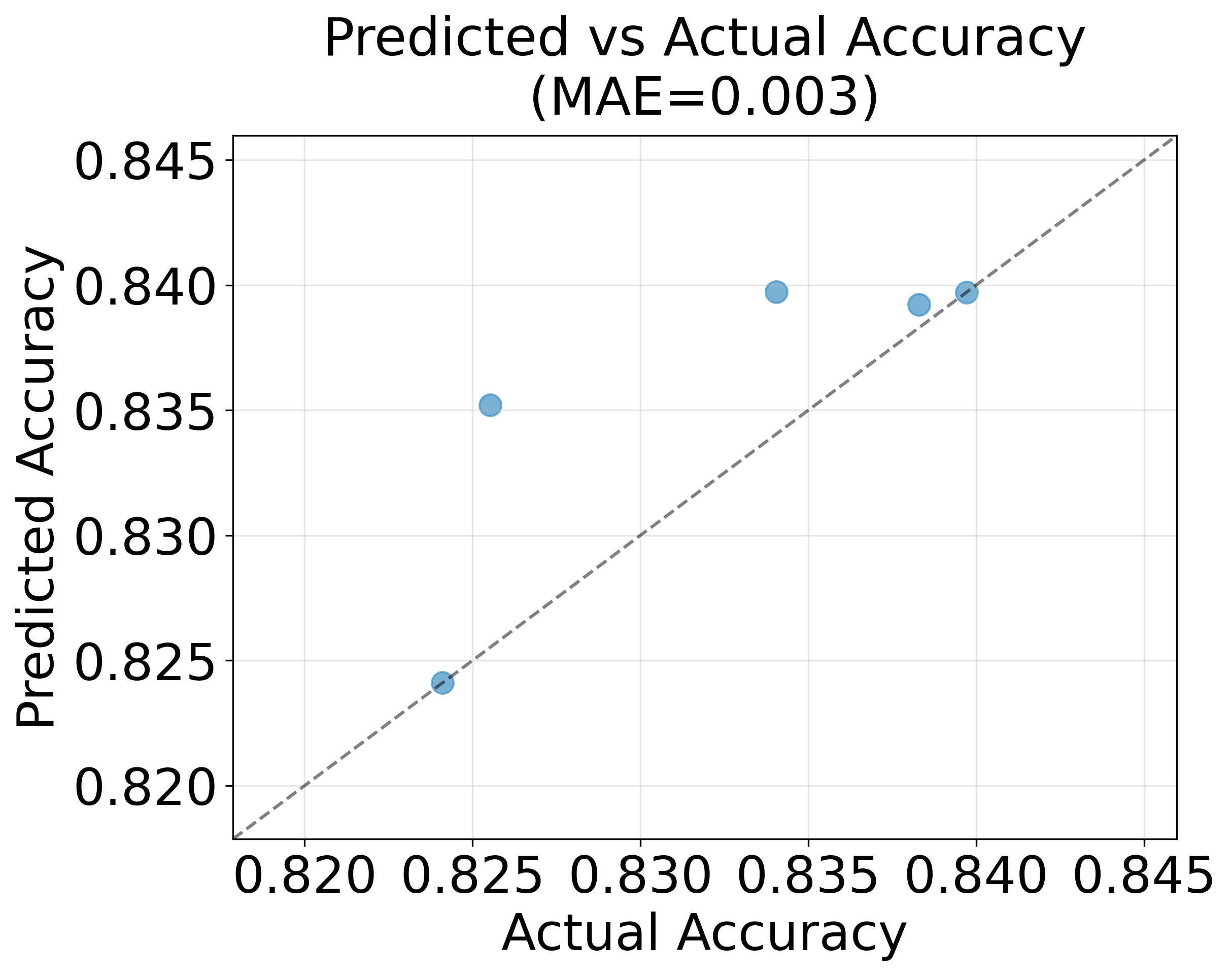}
\caption{\fiqasa}
\end{subfigure}
\hfill
\begin{subfigure}[t]{0.24\textwidth}
\includegraphics[width=\textwidth]{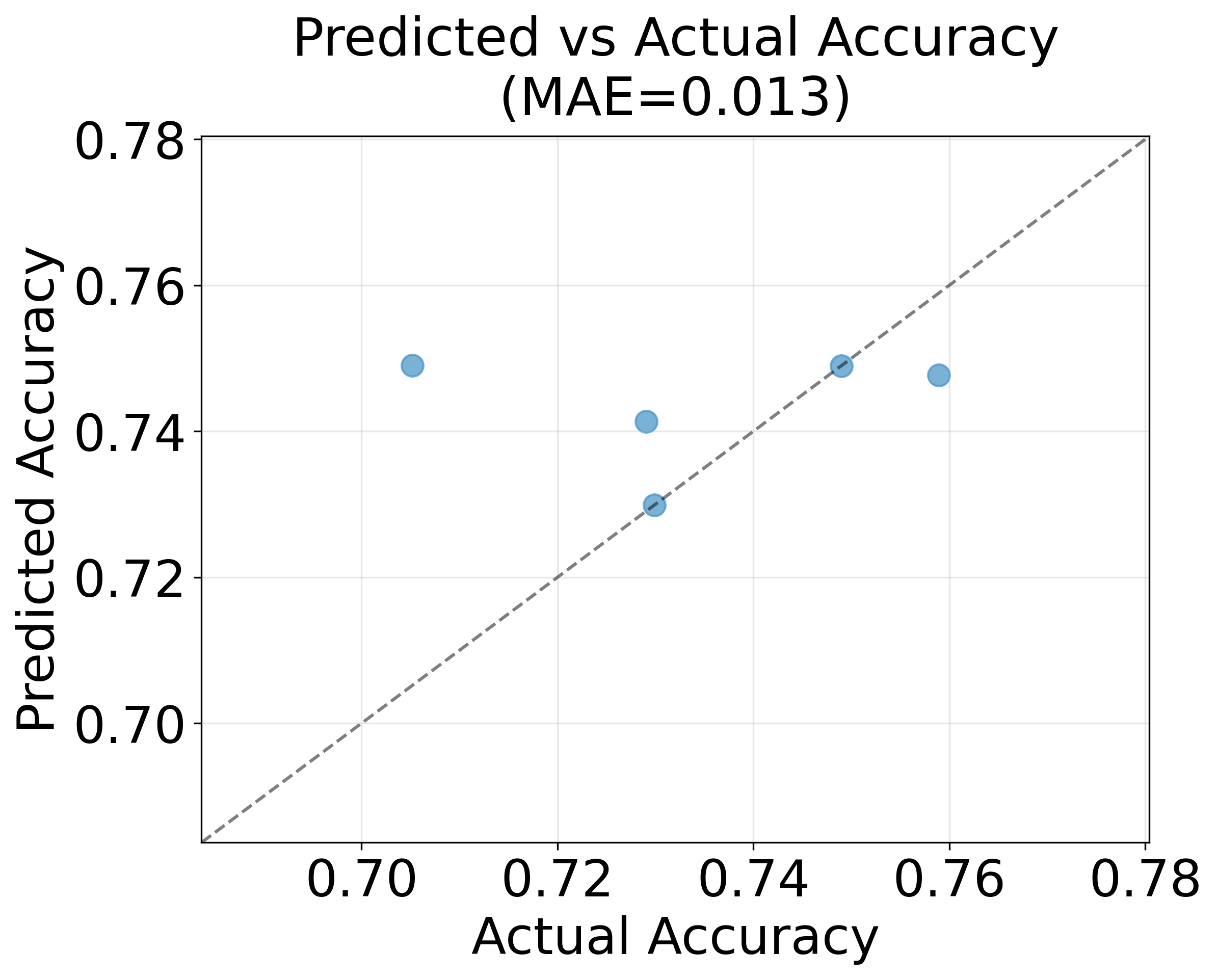}
\caption{\headline}
\end{subfigure}
\hfill
\begin{subfigure}[t]{0.24\textwidth}
\includegraphics[width=\textwidth]{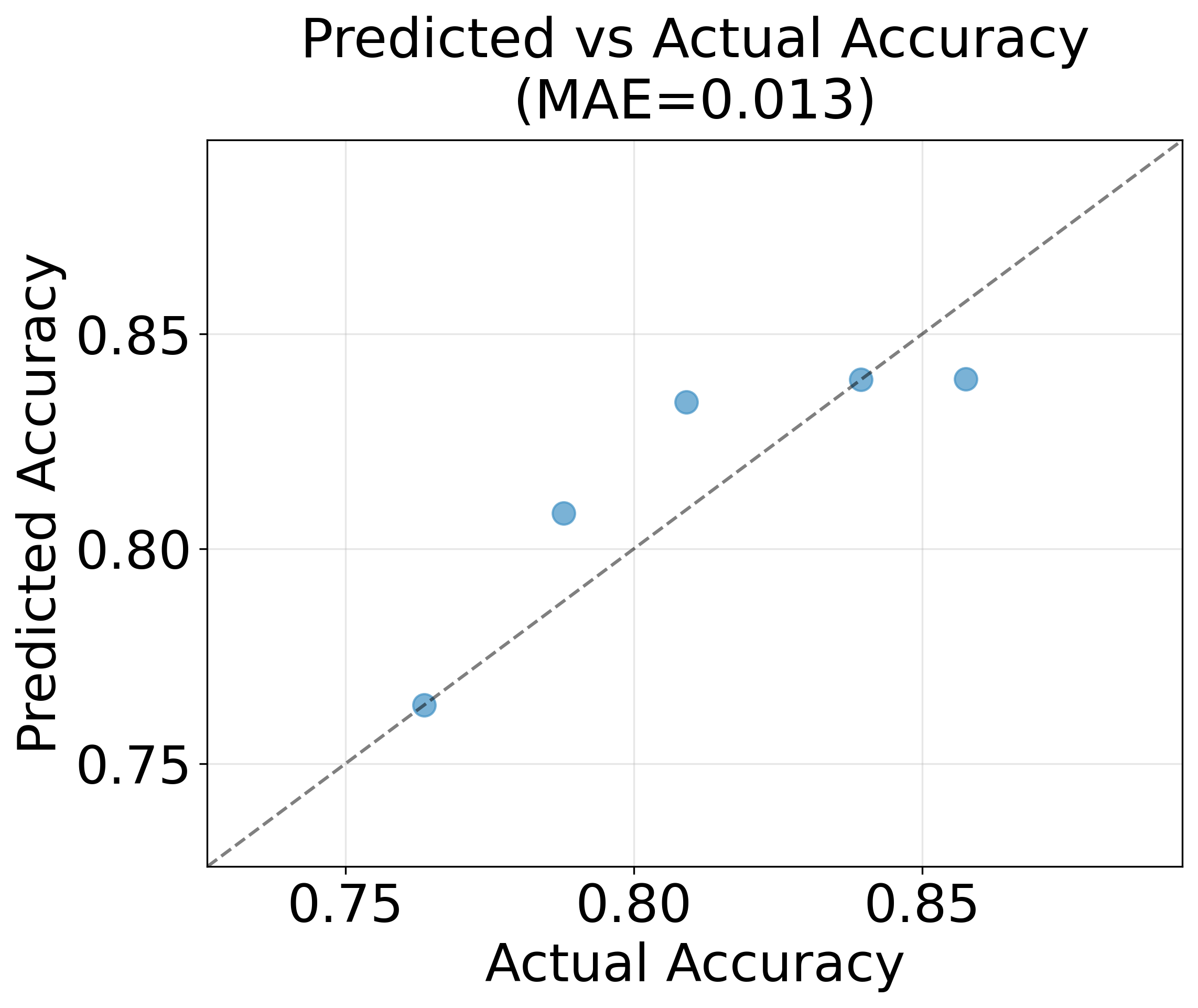}
\caption{\multifin}
\end{subfigure}
\caption{Scatter plots comparing predicted vs. actual accuracy for ICL across eight diverse tasks. Axes are zoomed to highlight fine-grained prediction details, with an average MAE of 0.85\% demonstrating high prediction fidelity. The consistent performance across both general domain (a-d) and financial domain (e-h) tasks validates our framework's robust prediction capabilities.}
\label{fig:detailed_icl_perf}
\end{figure*}

\subsection{A Closer Look at \methodabb's Prediction Ability for Retrieval-Augmented In-Context Learning}\label{app:cap-zoom-in-icl}

\begin{figure*}[h]
\begin{subfigure}[t]{0.24\textwidth}
\includegraphics[width=\textwidth]{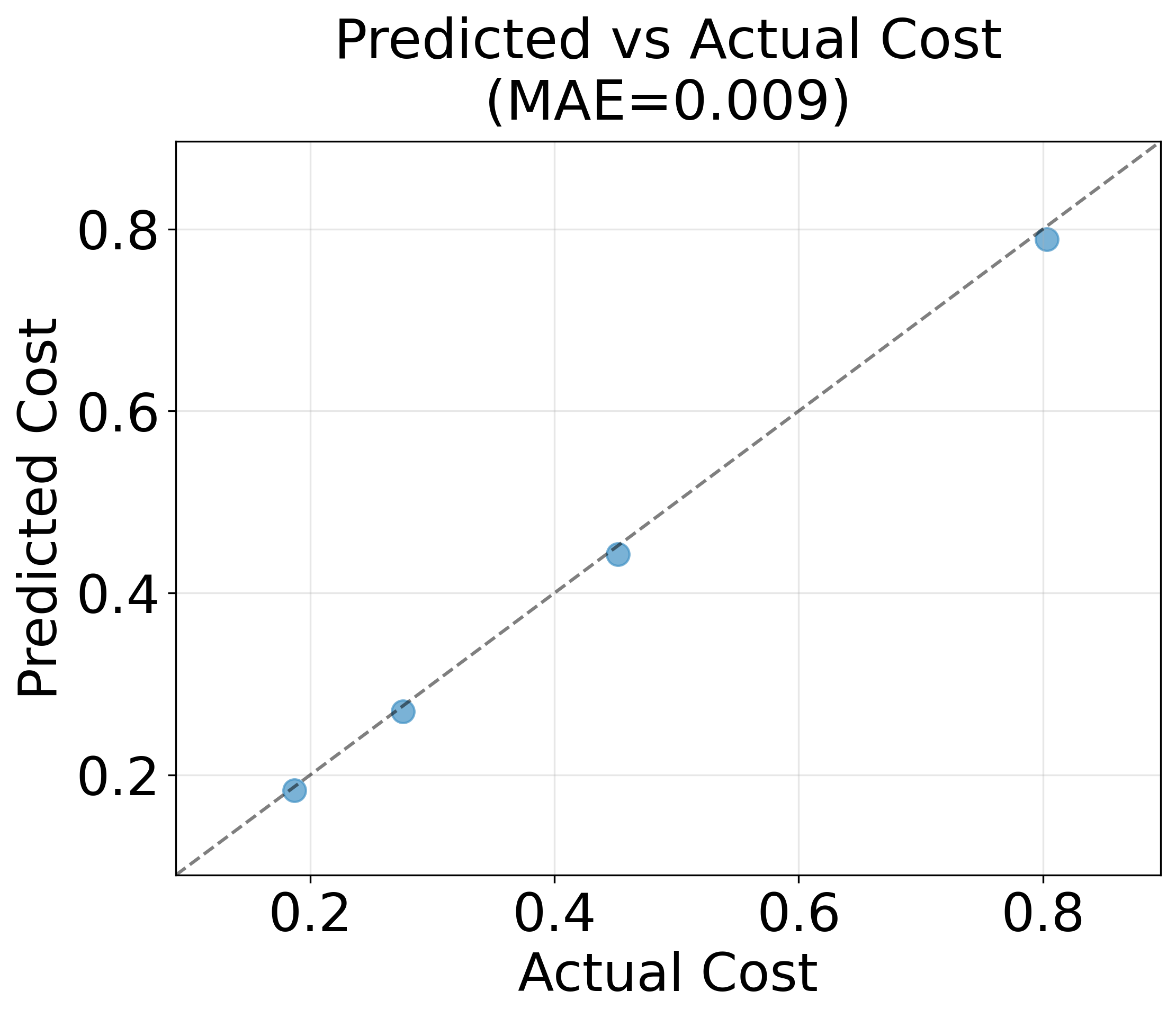}
\caption{\mmlu}
\end{subfigure}
\hfill
\begin{subfigure}[t]{0.24\textwidth}
\includegraphics[width=\textwidth]{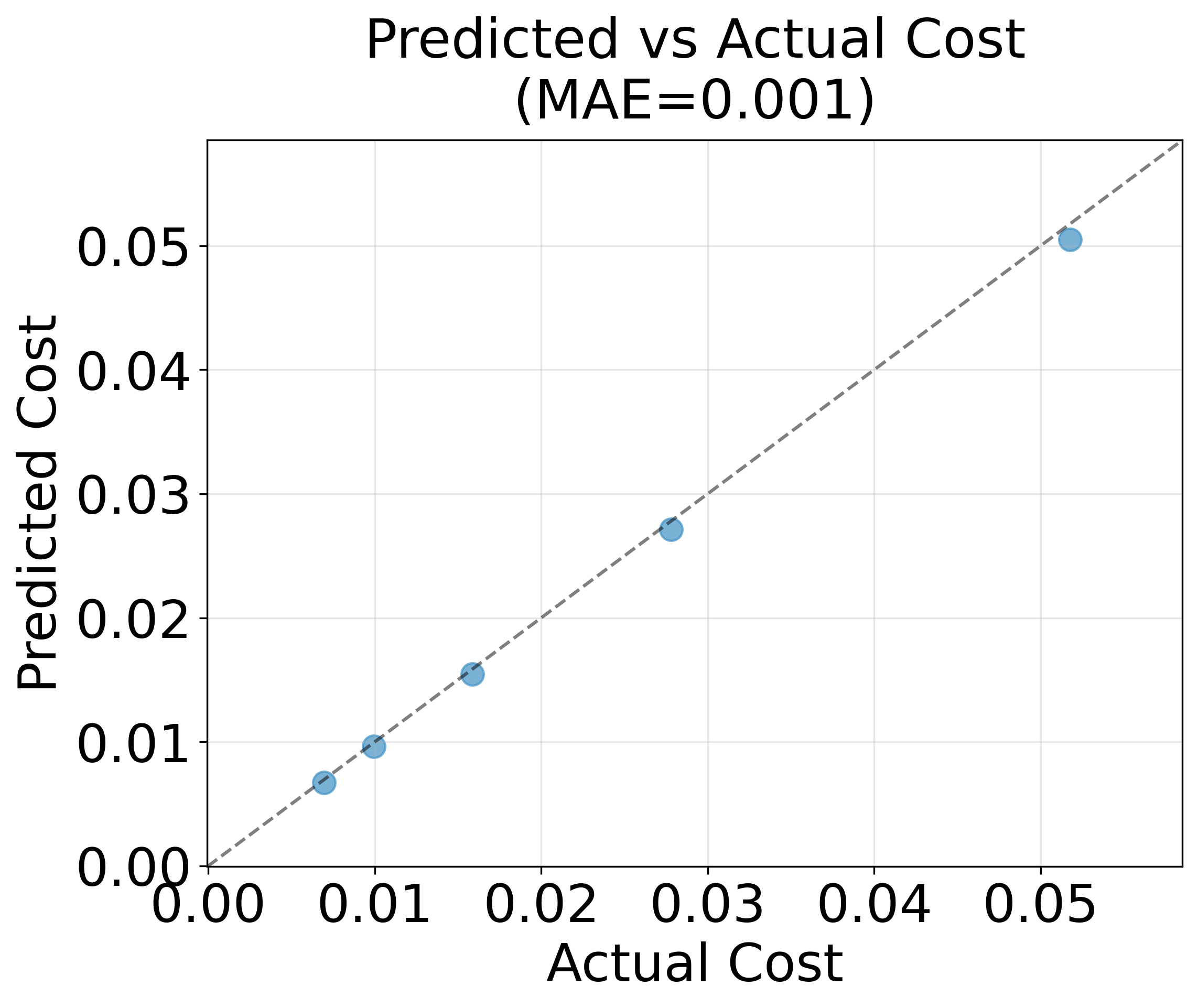}
\caption{\wino}
\end{subfigure}
\hfill
\begin{subfigure}[t]{0.24\textwidth}
\includegraphics[width=\textwidth]{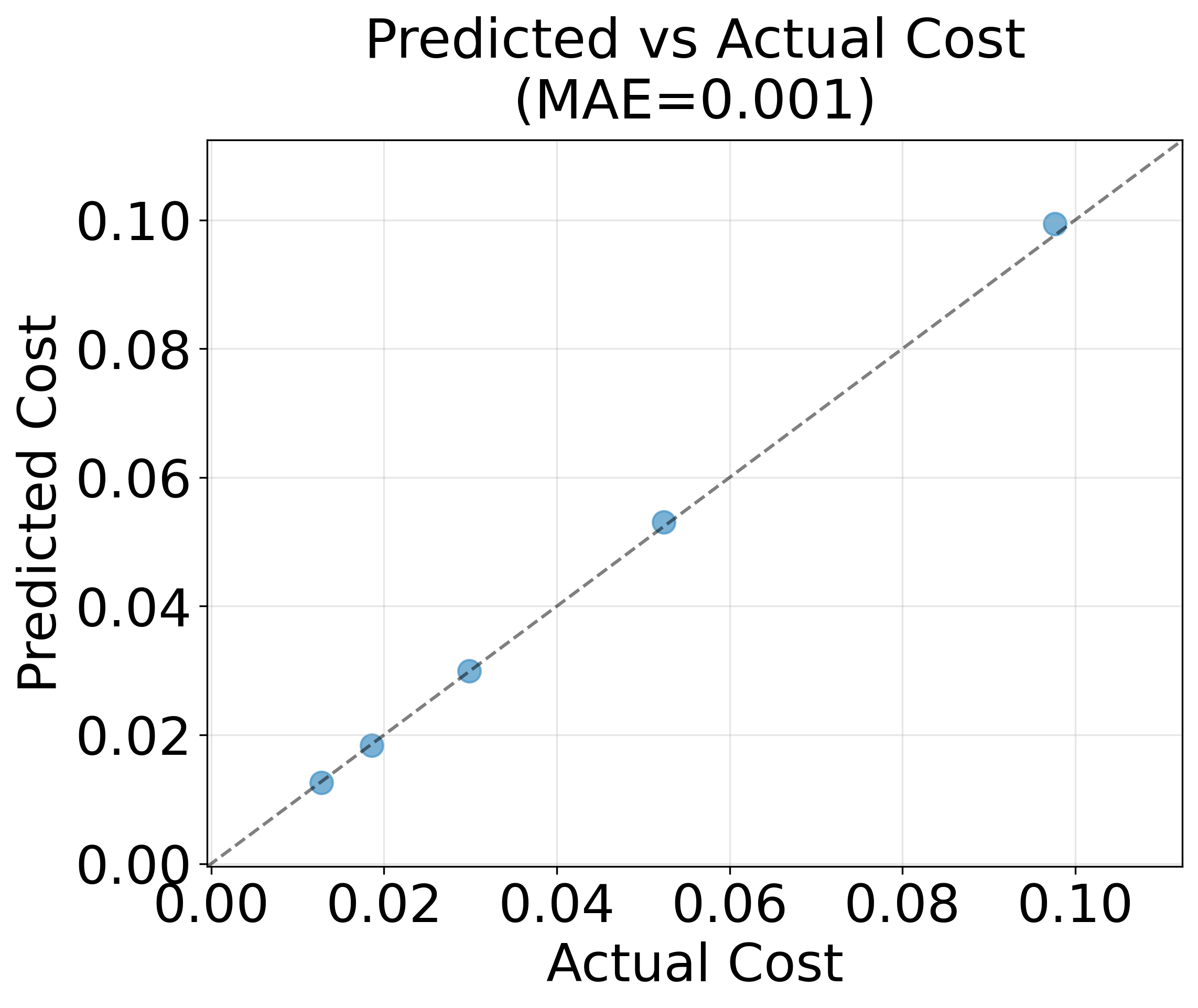}
\caption{\arc}
\end{subfigure}
\hfill
\begin{subfigure}[t]{0.24\textwidth}
\includegraphics[width=\textwidth]{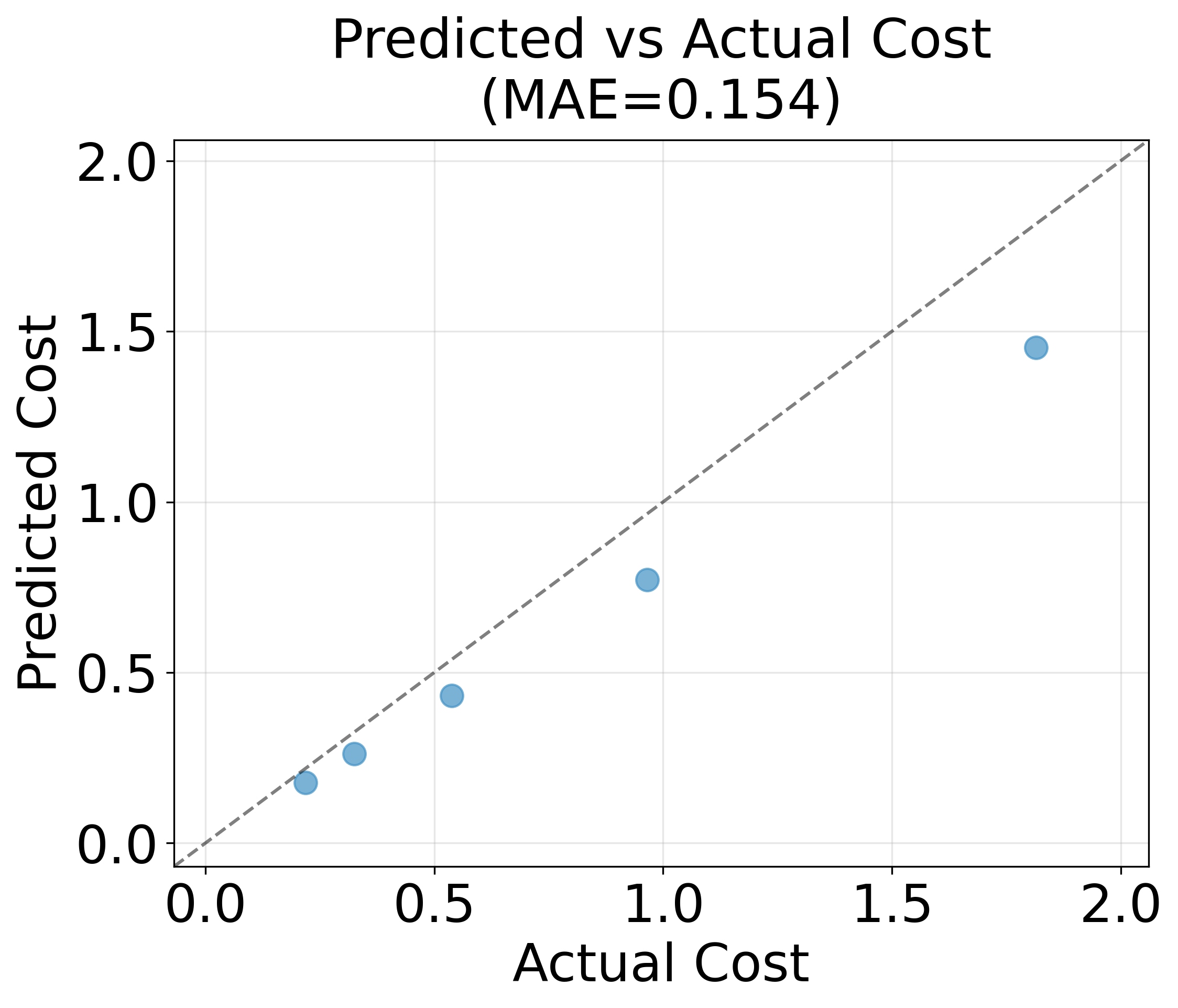}
\caption{\hella}
\end{subfigure}
\\[1em]
\begin{subfigure}[t]{0.24\textwidth}
\includegraphics[width=\textwidth]{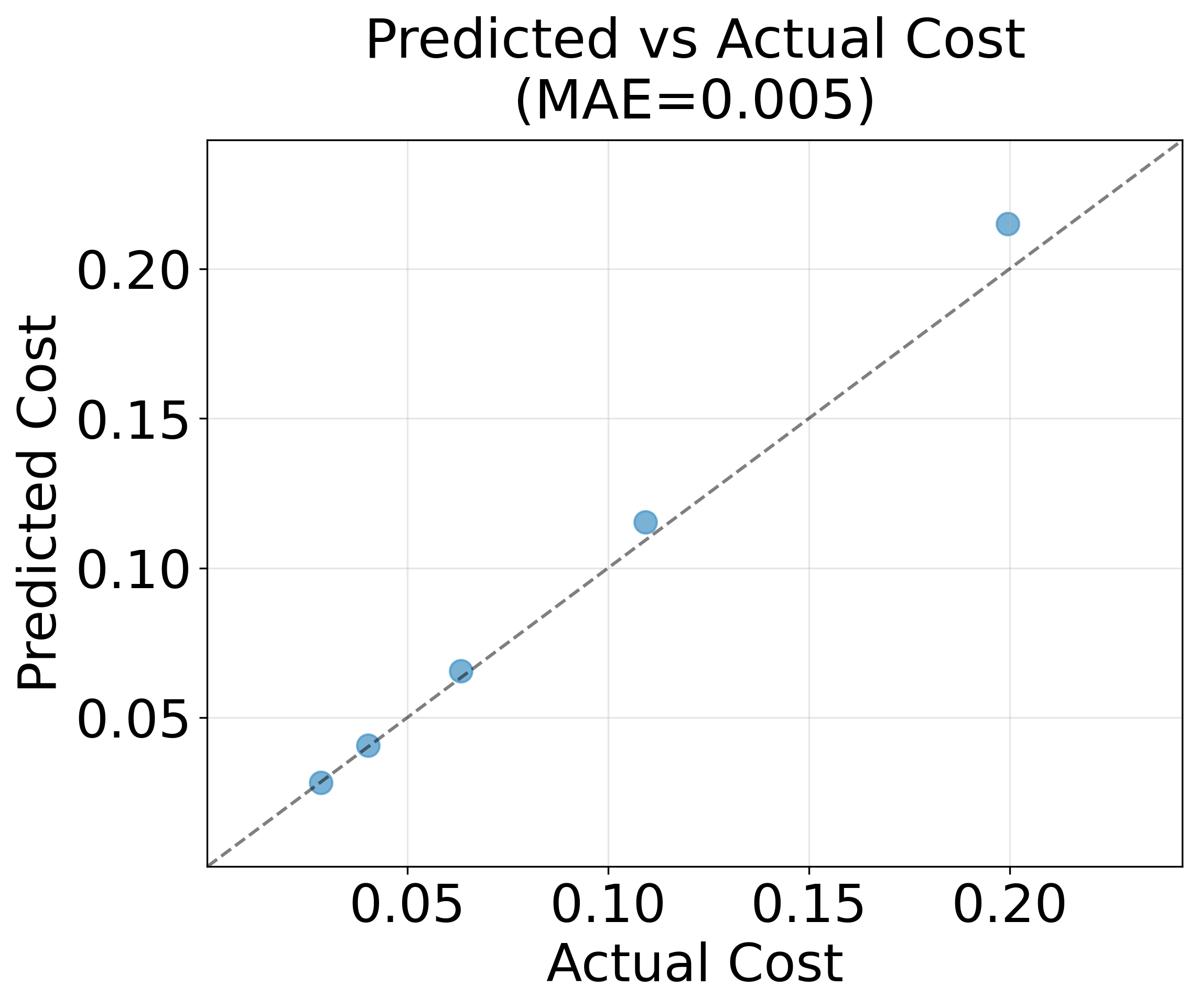}
\caption{\fpb}
\end{subfigure}
\hfill
\begin{subfigure}[t]{0.24\textwidth}
\includegraphics[width=\textwidth]{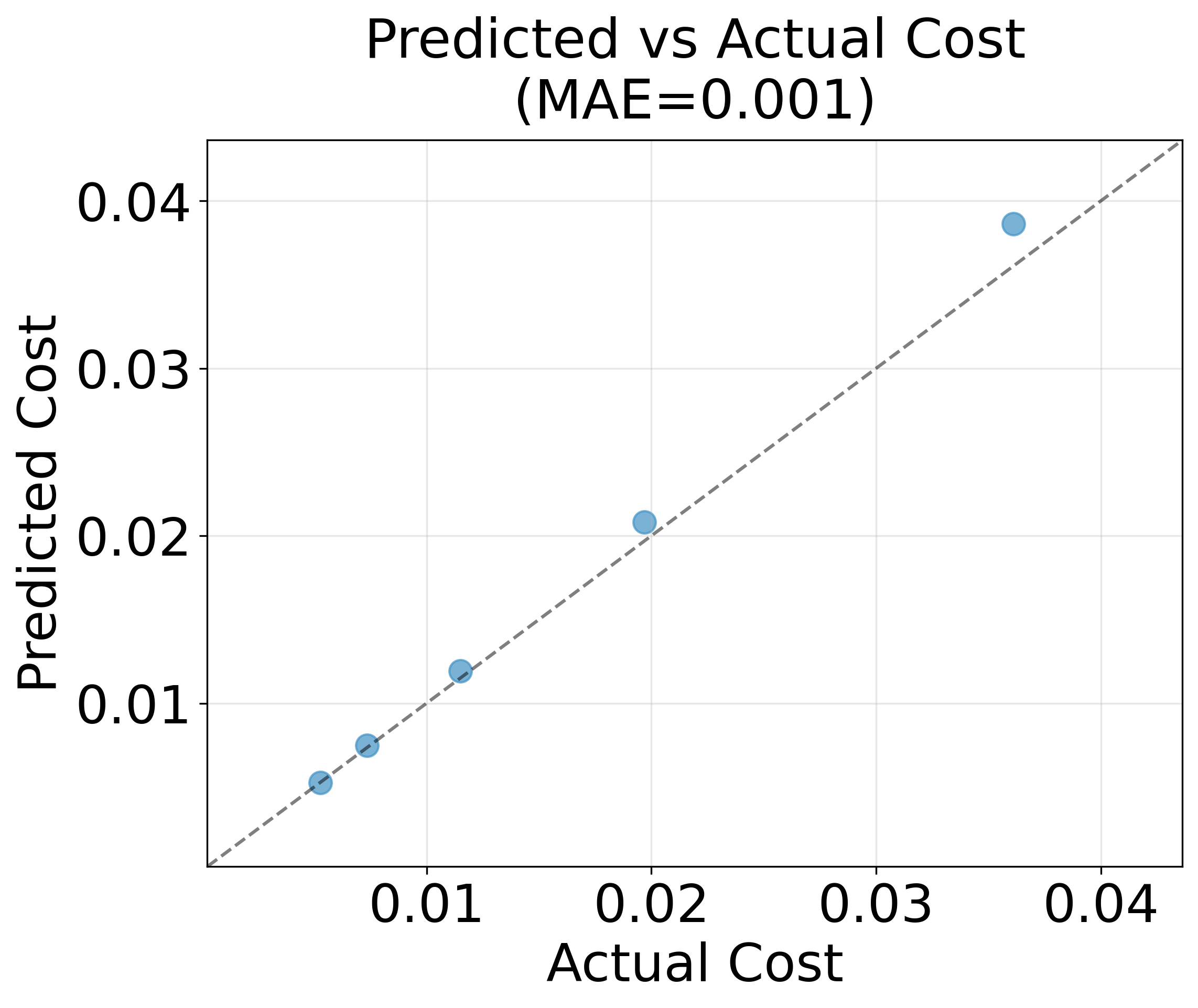}
\caption{\fiqasa}
\end{subfigure}
\hfill
\begin{subfigure}[t]{0.24\textwidth}
\includegraphics[width=\textwidth]{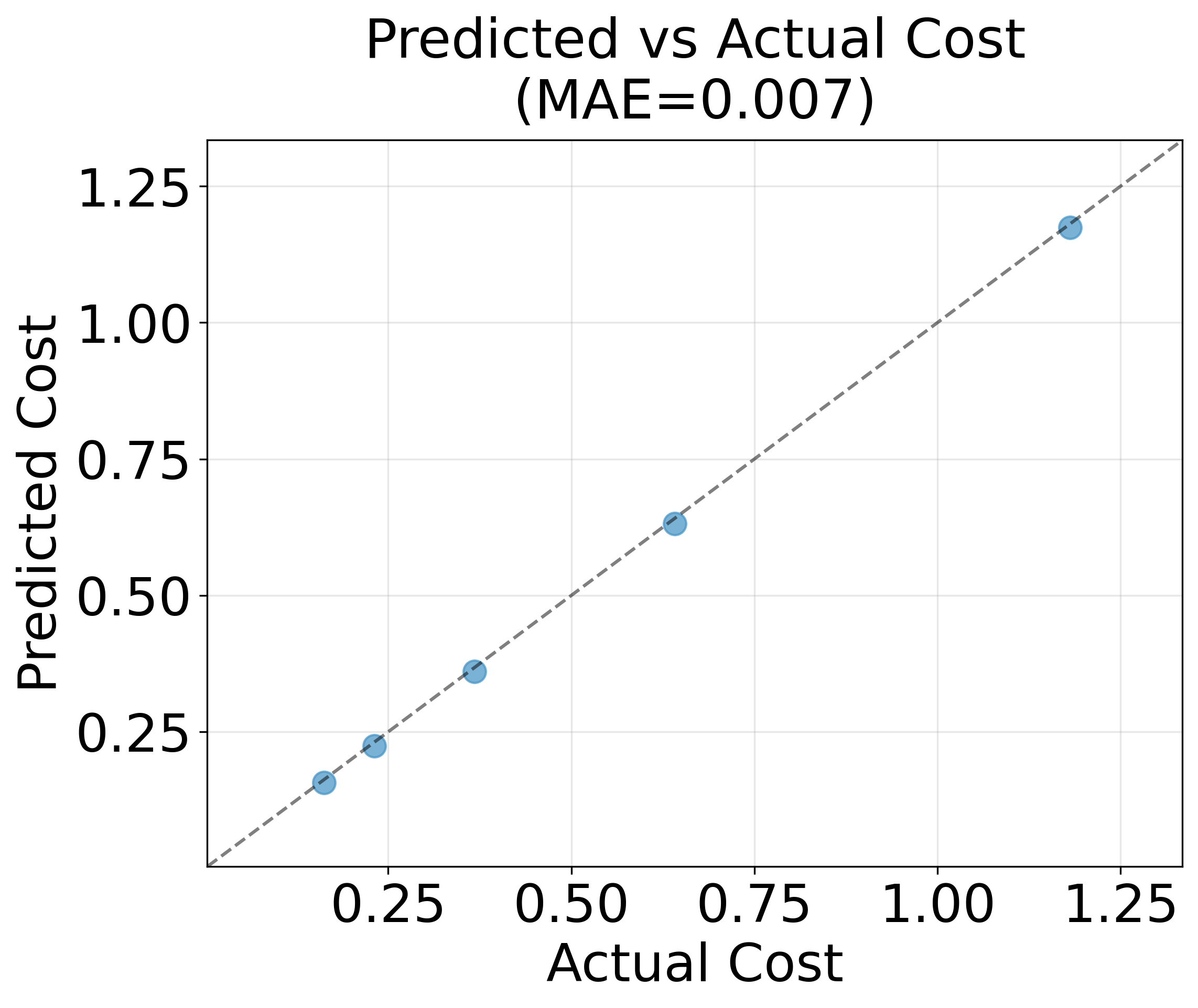}
\caption{\headline}
\end{subfigure}
\hfill
\begin{subfigure}[t]{0.24\textwidth}
\includegraphics[width=\textwidth]{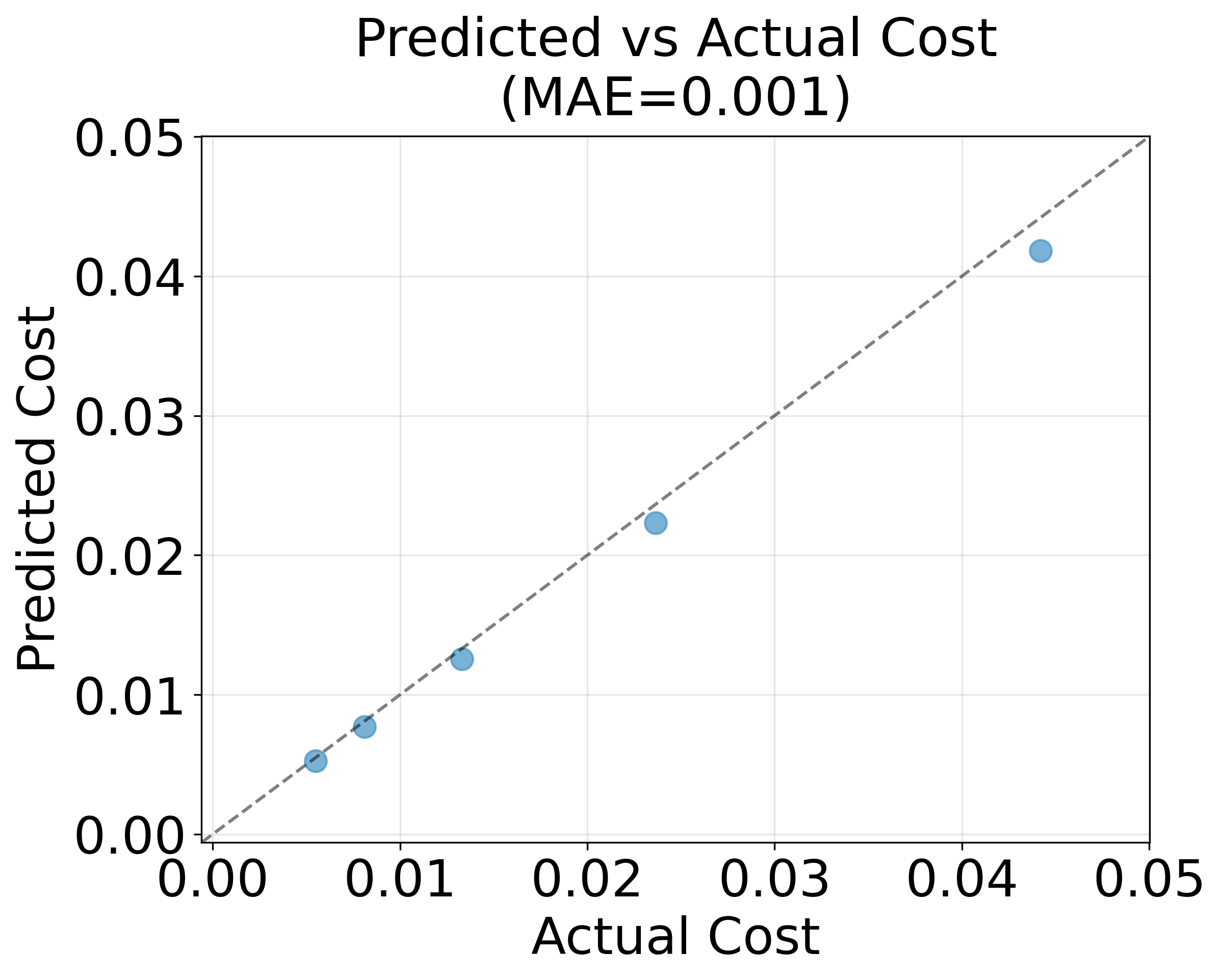}
\caption{\multifin}
\end{subfigure}
\caption{Scatter plots of predicted vs. actual cost for \icl across eight diverse tasks.}
\label{fig:detailed_icl_cost}
\end{figure*}

\paragraph{Performance prediction.}
Figure~\ref{fig:detailed_icl_perf} demonstrates our framework's prediction accuracy for retrieval-augmented ICL across eight tasks. With deliberately zoomed axes to highlight prediction granularity, the scatter plots reveal strong performance with MAE ranging from 0.003 (\fiqasa) to 0.013 (Multifin EN and Headline). The framework maintains consistent accuracy across both general-domain tasks (MMLU: 0.007, Winogrande: 0.005, ARC-Challenge: 0.008, HellaSwag: 0.012) and financial tasks (FPB: 0.007, \fiqasa: 0.003, Headline: 0.013, Multifin EN: 0.013), with an average MAE of 0.0085. This low average deviation of 0.85\% from actual performance validates our framework's robust prediction capabilities across diverse domains.

\paragraph{Cost prediction.}
Figure~\ref{fig:detailed_icl_cost} demonstrates our framework's cost prediction capabilities for ICL across eight tasks. Most tasks show excellent prediction accuracy with MAE ranging from 0.001 (Winogrande, ARC-Challenge, FiQA-SA, Multifin EN) to 0.009 (MMLU), with HellaSwag being the only outlier (MAE=0.154). 
This higher deviation in HellaSwag stems from our design choice to use training set averages for sequence length estimation instead of observed samples during performance model fitting, prioritizing efficiency over perfect accuracy. While this approximation is typically sufficient for cost estimation, prediction accuracy could be trivially improved by using observed sample lengths when higher precision is needed.

\section{Full Results for Combining Training- and Test-time Strategies}\label{app:combine-train-test}

\begin{figure*}[h]
\begin{subfigure}[t]{0.24\textwidth}
\includegraphics[width=\textwidth]{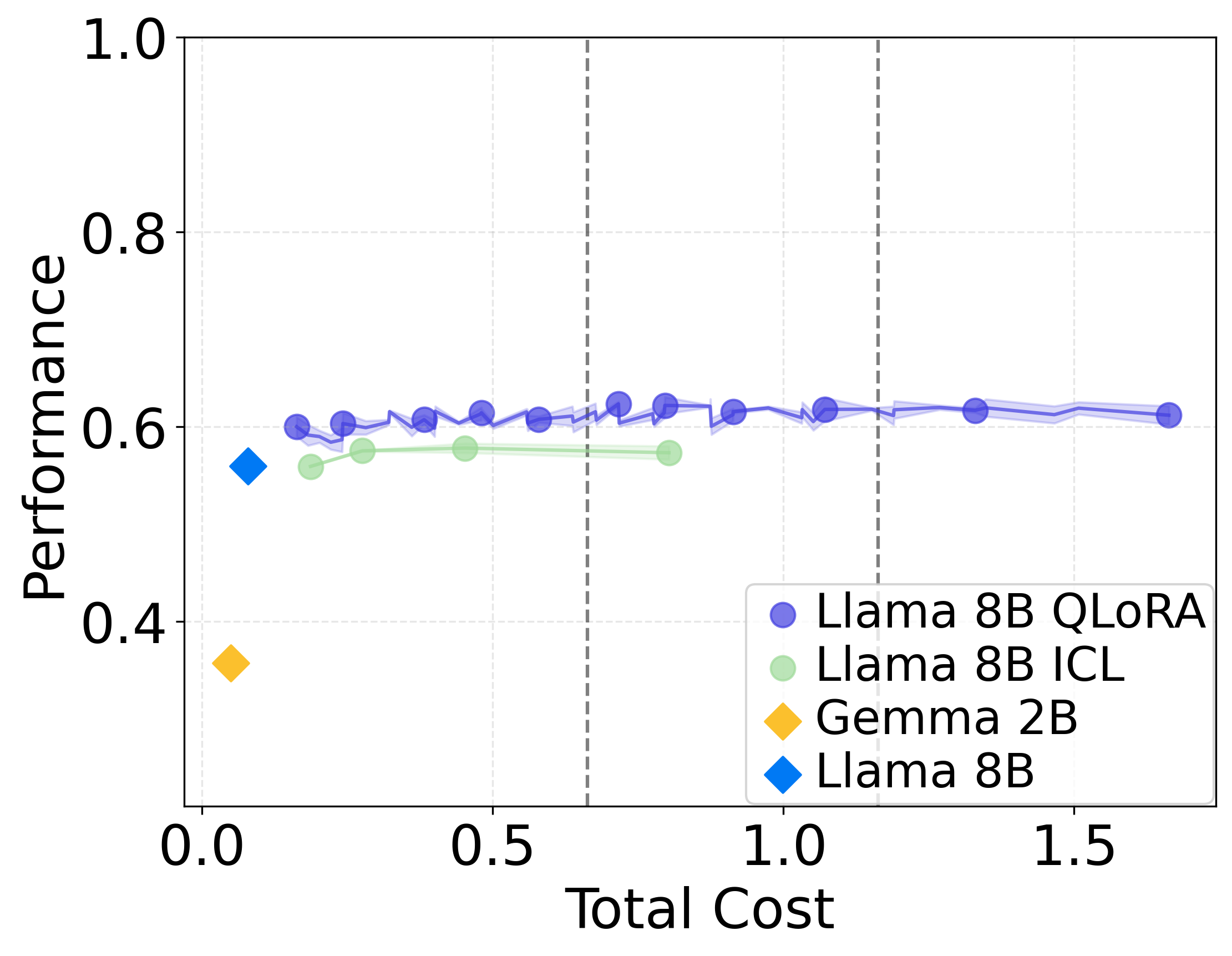}
\caption{\mmlu}
\end{subfigure}
\hfill
\begin{subfigure}[t]{0.24\textwidth}
\includegraphics[width=\textwidth]{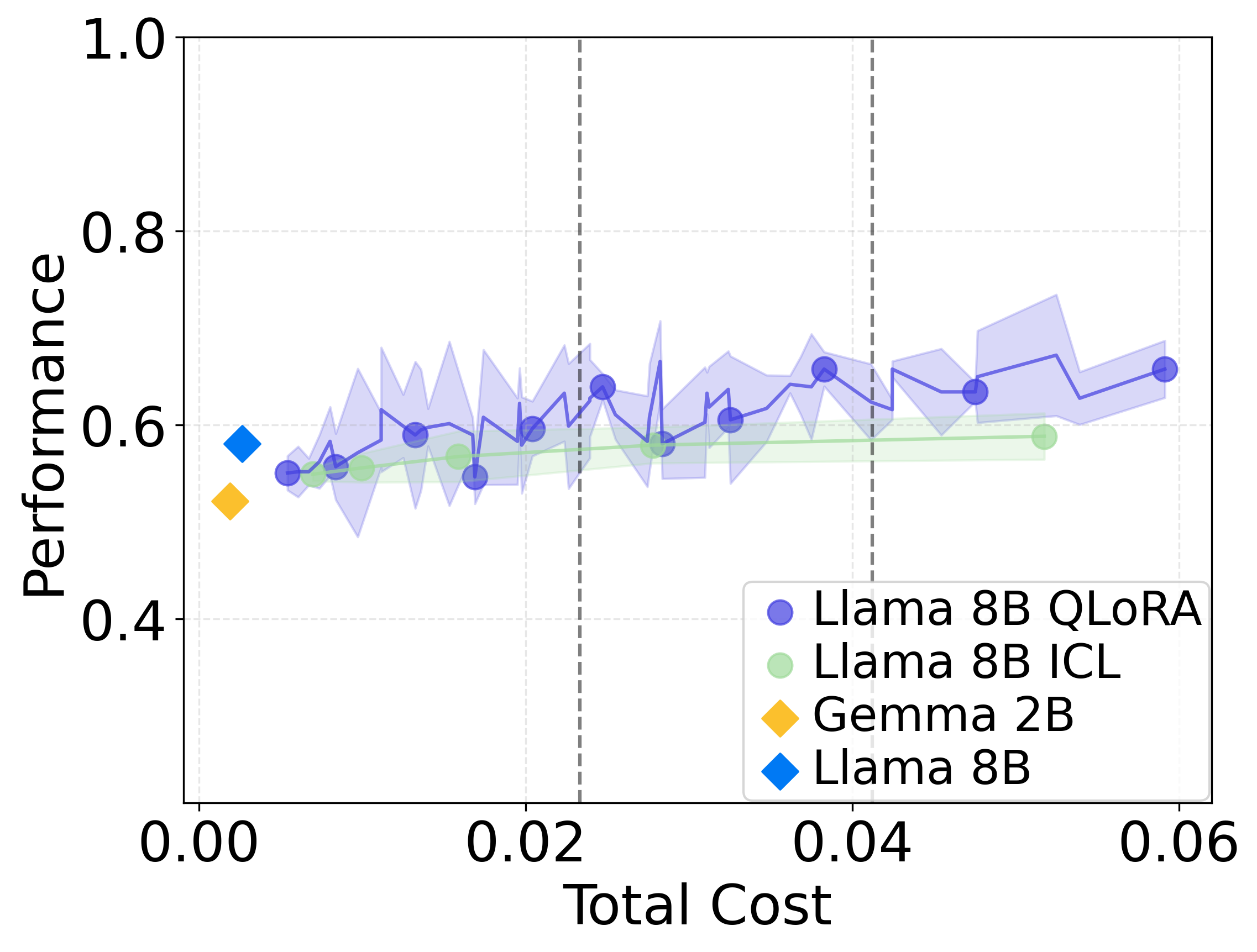}
\caption{\wino}
\end{subfigure}
\hfill
\begin{subfigure}[t]{0.24\textwidth}
\includegraphics[width=\textwidth]{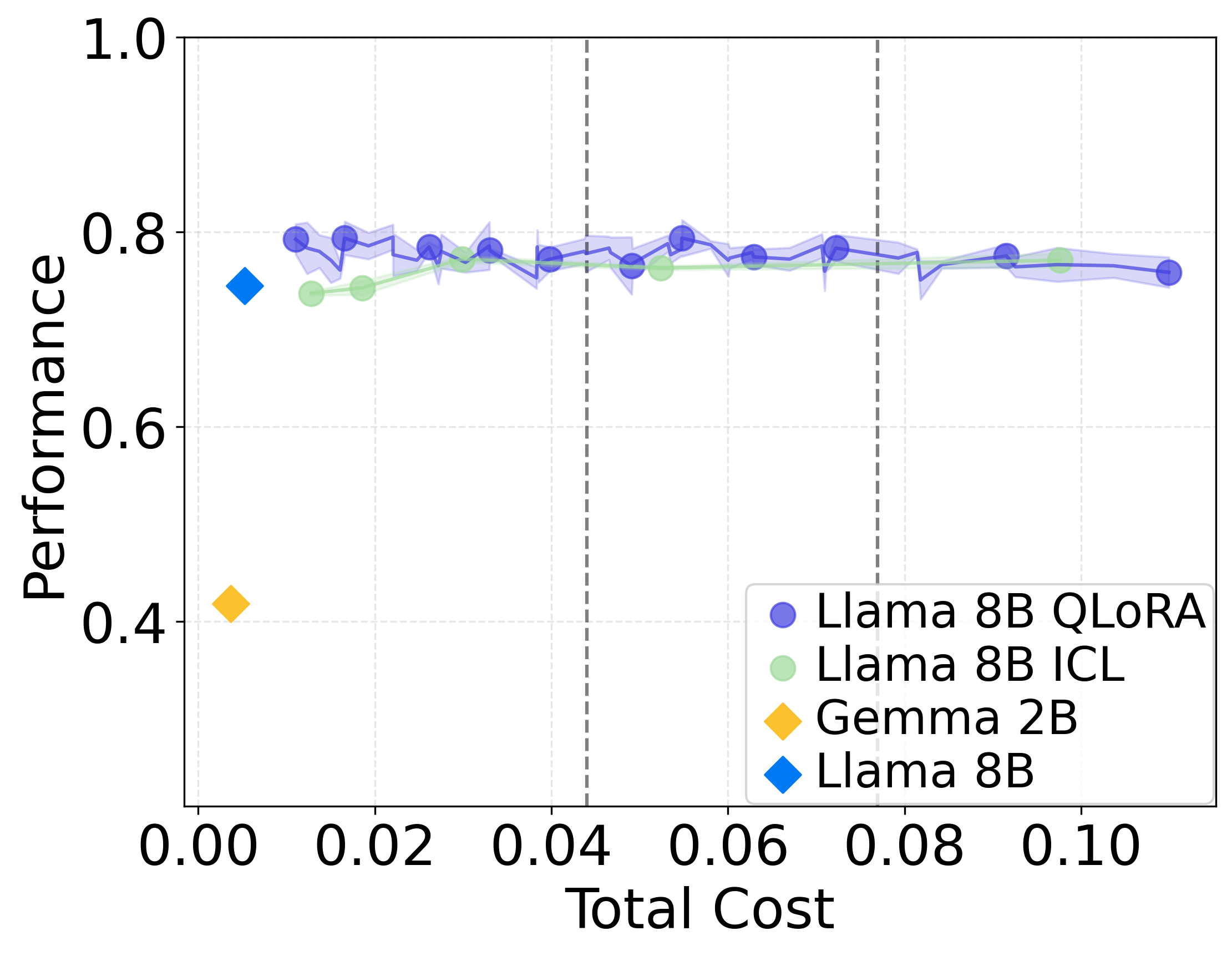}
\caption{\arc}
\end{subfigure}
\hfill
\begin{subfigure}[t]{0.24\textwidth}
\includegraphics[width=\textwidth]{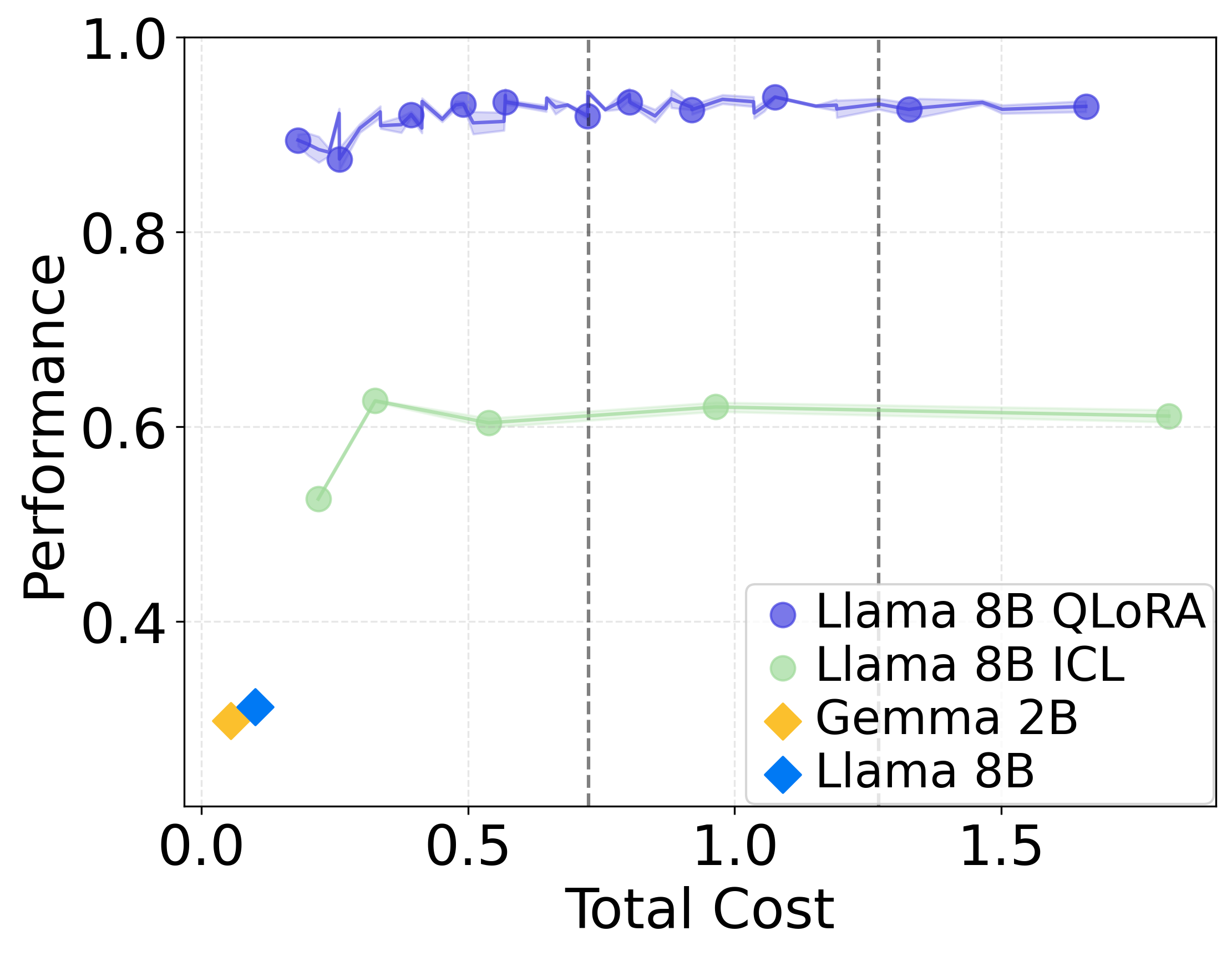}
\caption{\hella}
\end{subfigure}
\\[1em]
\begin{subfigure}[t]{0.24\textwidth}
\includegraphics[width=\textwidth]{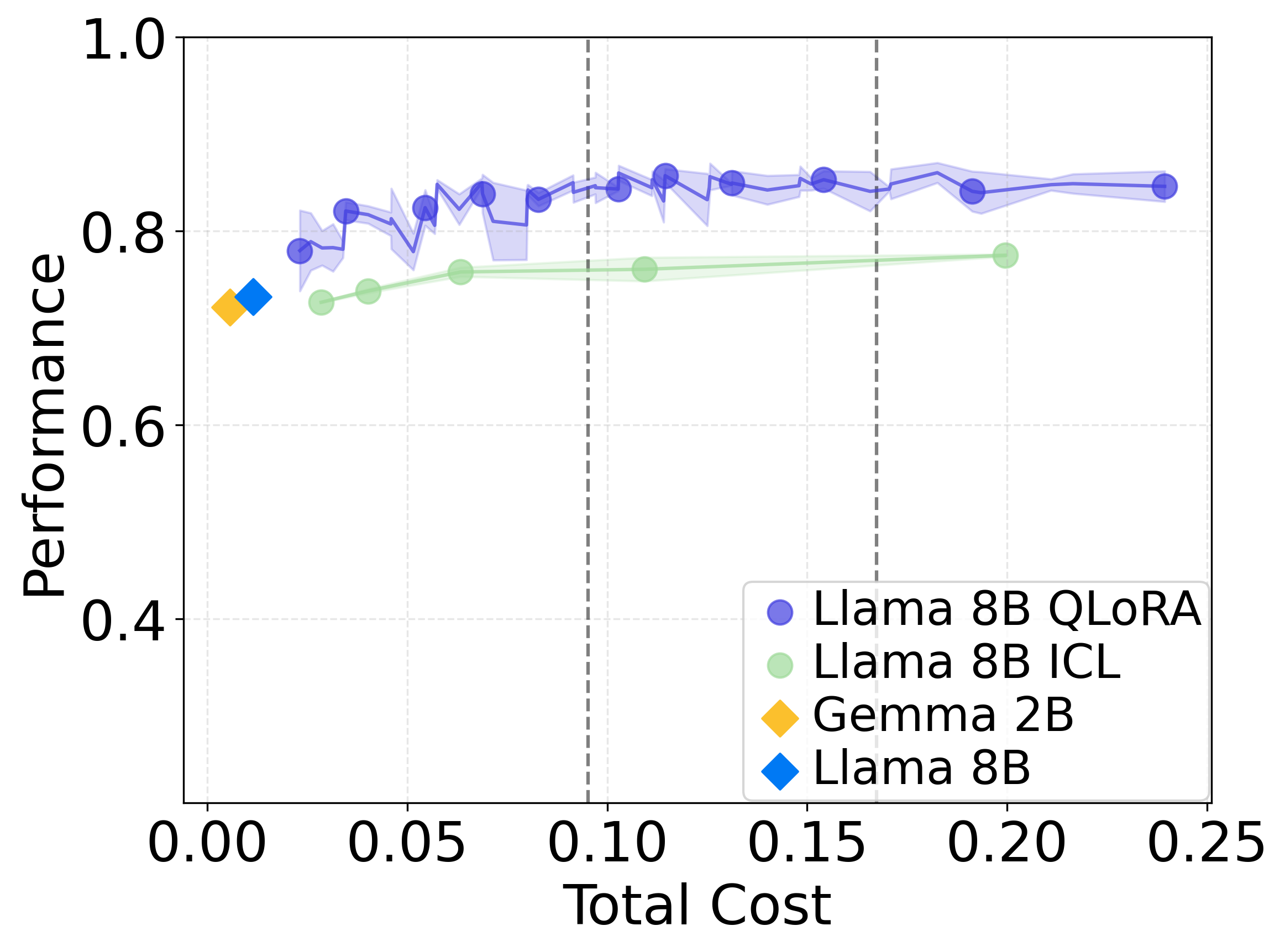}
\caption{\fpb}
\end{subfigure}
\hfill
\begin{subfigure}[t]{0.24\textwidth}
\includegraphics[width=\textwidth]{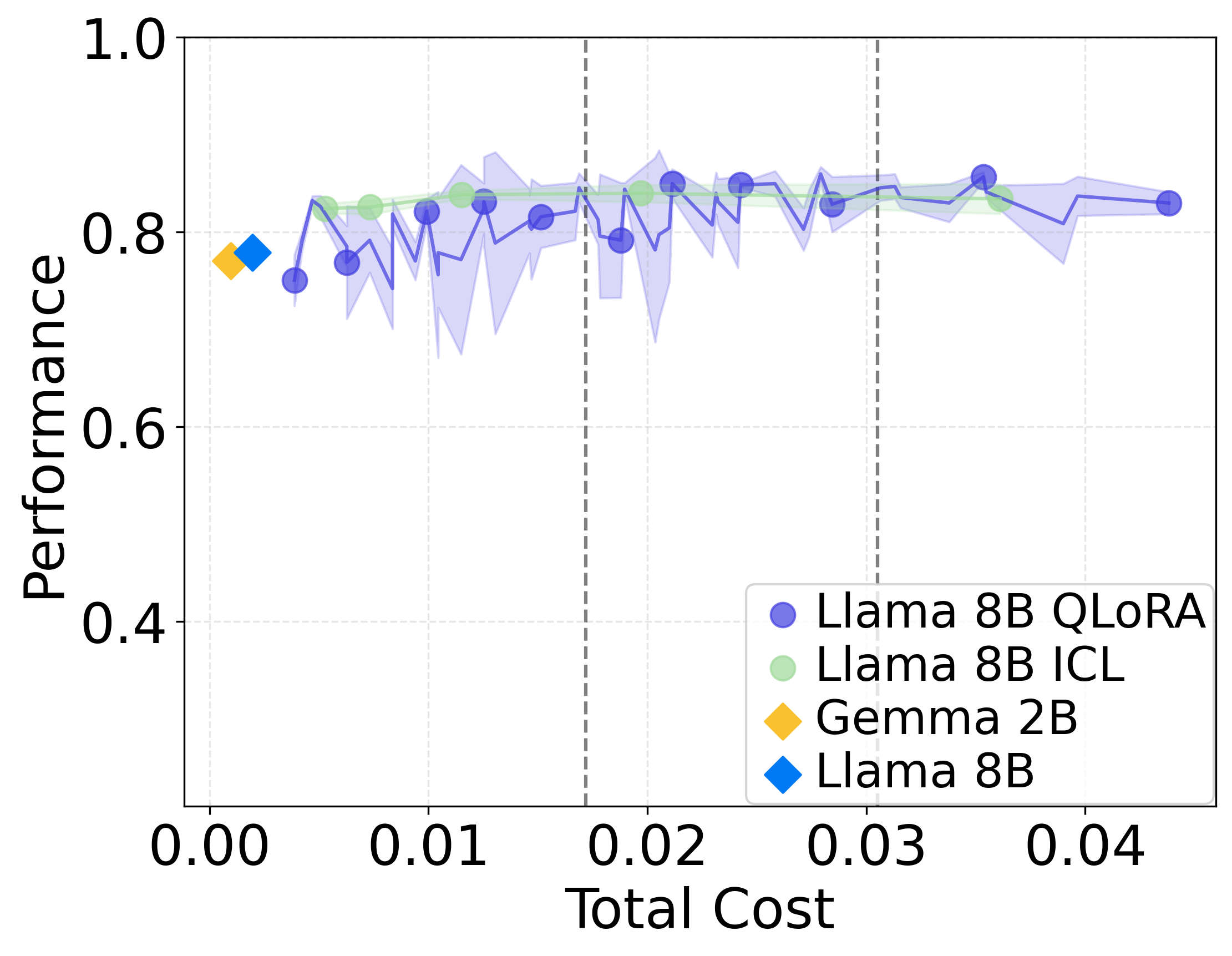}
\caption{\fiqasa}
\end{subfigure}
\hfill
\begin{subfigure}[t]{0.24\textwidth}
\includegraphics[width=\textwidth]{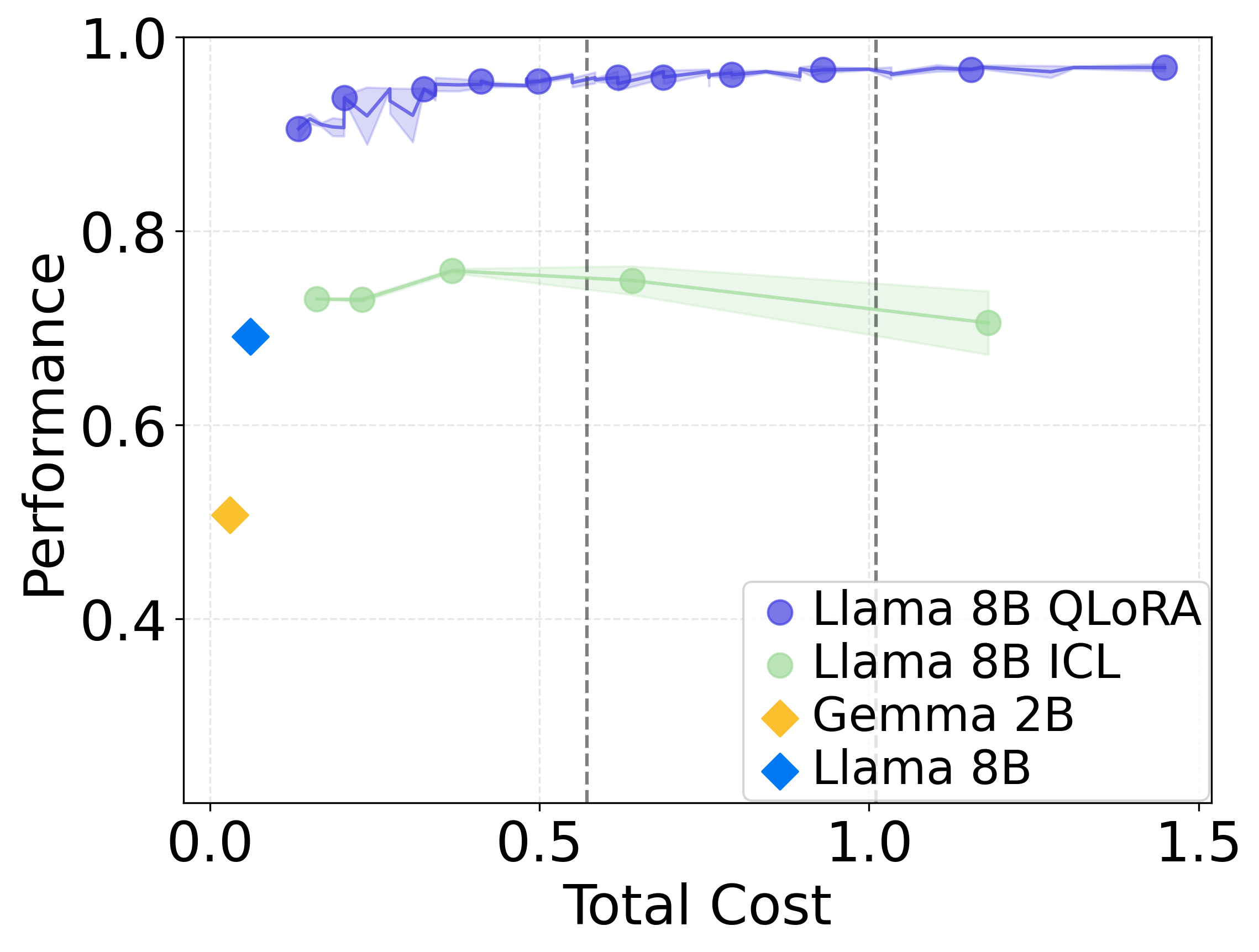}
\caption{\headline}
\end{subfigure}
\hfill
\begin{subfigure}[t]{0.24\textwidth}
\includegraphics[width=\textwidth]{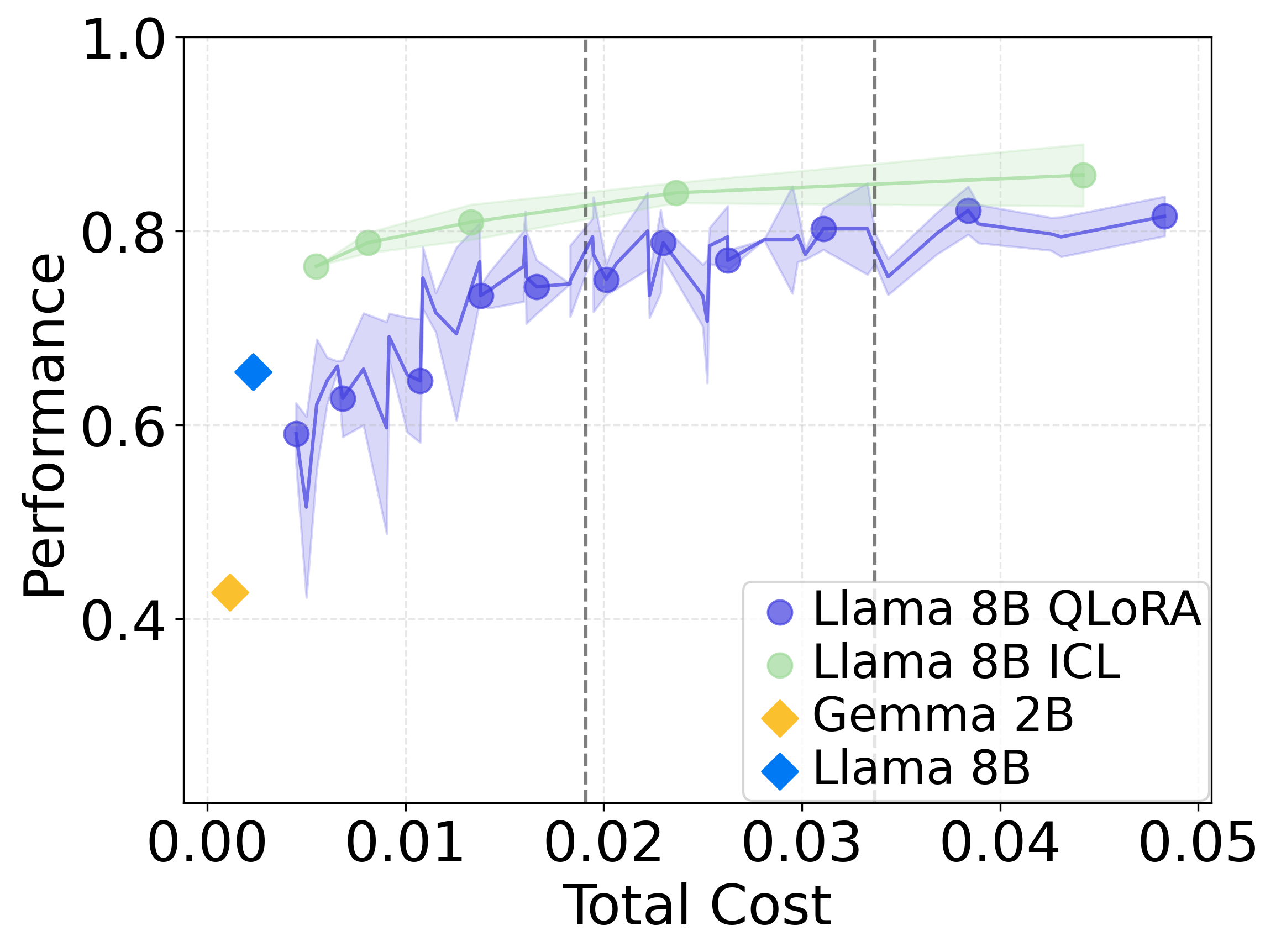}
\caption{\multifin}
\end{subfigure}
\caption{Actual \qlora vs. \icl performance-cost trajectories across eight diverse tasks. Each plot presents the performance-cost curves for \qlora (\textcolor{actualQloraColor}{$\bullet$}) and \icl (\textcolor{actualIclColor}{$\bullet$}) on \llama, with \gemma ($\gemmadiamond$), and \llama ($\llamadiamond$) serving as baselines.
Vertical dashed lines demarcate low, medium, and high-cost thresholds, determined by the minimum and maximum costs of both adaptation strategies. 
The shaded regions represent the standard deviation across 3 seeds for each configuration.}
\label{fig:actual-qlora-icl-performance-cost-full}
\end{figure*}
Following the four representative tasks shown in Section~\ref{sec5.3:scaling-behavior}, we present in Figure~\ref{fig:actual-qlora-icl-performance-cost-full} the full scaling behavior results across all eight tasks.

\section{Extended Potential Implication}\label{app:implication}
Fine-tuning large language models (LLMs) at scale presents significant computational and financial challenges. 
We analyze these costs through a practical case study of fine-tuning GPT-4o, using OpenAI's current pricing structure of \$25 per million training tokens\footnote{\url{https://openai.com/api/pricing/}}. Following the training protocol established in the Llama 3 report~\citep{dubey2024llama}, a typical fine-tuning run requires an average of 8.75K steps with sequences of 8,192 tokens each, amounting to approximately 71M tokens per complete pass through the training data.
To identify optimal fine-tuning parameters, practitioners typically need to explore various hyperparameter combinations. Consider a systematic exploration of epochs (ranging from 1 to 10) and data mixing strategies (10 options), resulting in 100 total trials. 
Each trial with a single epoch over the full dataset costs \$1,775 in training compute alone, with costs scaling linearly with the number of epochs. 
Assume the validation phase requires processing queries totaling 1M tokens, with model outputs averaging 3x the input length due to multi-step reasoning. 
Given OpenAI's pricing of \$2.5 per million input tokens and \$10 per million output tokens, each validation run incurs an additional cost of \$32.5. Across all trials, the validation cost sums to \$3,250, while the total training cost reaches \$976,250 (55 total epochs × 10 data mixing strategies × \$1,775). This brings the total adaptation cost to approximately \$979,500.
Our method, \methodabb, substantially reduces these costs through accurate performance prediction using a small amount of data.
In resource-intensive scenarios, our approach achieves a cost reduction of 95.95\% compared to the total cost, enabling practitioners to estimate the performance-cost trade-offs of different configurations while only running a small subset of experiments. This reduces the total cost to approximately \$39,670--a 24.7x reduction. This dramatic cost reduction enables practitioners to make informed decisions about the performance-cost trade-offs that best suit their specific requirements and constraints.

\section{Expanded Model Evaluation: Diverse Families and Scales}\label{app:expandmodel}

\begin{table}[h]
\centering
\resizebox{0.9\textwidth}{!}{
\begin{tabular}{llcccccc}
\toprule
\textbf{Tasks} & \textbf{Cost Level} & \textbf{Pred. Acc (\%)} & \textbf{Act. Acc (\%)} & \textbf{MAE $\downarrow$ (\%)} & \textbf{Act. Total Cost (\$)} & \textbf{Ours Cost (\$)} & \textbf{CRR $\uparrow$ (\%)} \\
\midrule
\multirow{3}{*}{HellaSwag} 
                      & Low & 94.87 & 94.87 & 0.00 & 27.042 & 2.354 & 91.30 \\
                      & Medium & 94.71 & 95.64 & 0.93 & 62.838 & 2.470 & 96.07 \\
                      & High & 94.94 & 95.84 & 0.90 & 124.414 & 2.759 & 97.78 \\
\midrule
\multirow{3}{*}{Multifin EN} 
                      & Low & 84.09 & 85.76 & 1.67 & 0.748 & 0.271 & 63.72 \\
                      & Medium & 85.91 & 86.36 & 0.45 & 1.751 & 0.265 & 84.88 \\
                      & High & 86.82 & 88.18 & 1.36 & 3.399 & 0.243 & 92.85 \\
\bottomrule
\end{tabular}
}
\vspace{0.1cm} 
\caption{COSMOS demonstrates robust performance across \textit{275} model--strategy--configuration combinations, spanning diverse architectures (Gemma, Llama, Qwen, Mistral) and scales (2B-9B), maintaining high prediction accuracy (average MAE of 0.61\% for HellaSwag and 1.16\% for Multifin EN) while drastically reducing computational costs.}
\label{tab:hellaswag-multifin-results}
\end{table}

To establish the generalizability of COSMOS, we significantly expand our evaluation framework beyond the initial Gemma 2B and Llama 3 8B models to encompass diverse model families and scales. This comprehensive evaluation now incorporates models from multiple architectures (Mistral~\cite{jiang2023mistral7b} and Qwen~\cite{2024qwen2}), newer generations (Gemma 2~\cite{2024gemma2}), and various parameter scales (7B to 9B). Following the experimental protocol established in Section~\ref{sec:experiments}, our expanded evaluation framework includes:
\begin{itemize}
    \item Model pool: 5 models including Gemma 2B, Llama 3 8B, Gemma 2 9B, Qwen 2 7B, Mistral 7B v0.3
    \item Strategies: QLoRA and retrieval-augmented ICL
    \item QloRA configurations: Training iterations $\in$ $\{4,5,6,7,8\}$ and data portions $\in$ $\{0.1, 0.2,..., 1.0\}$ at 0.1 increments
    \item ICL configurations: retrieved number of demonstrations $\in$ $\{1,2,4,8,16\}$
\end{itemize}

This systematic design yields 275 model--strategy--configuration combinations (5 models × (50 QLoRA + 5 ICL)) per task, with all results averaged across three random seeds to ensure statistical robustness.

The results in Table~\ref{tab:hellaswag-multifin-results} demonstrate COSMOS's consistent effectiveness across this expanded evaluation space even though we adopt the same training protocols for all models. 
For the general domain HellaSwag benchmark, COSMOS achieves an average Mean Absolute Error (MAE) of merely 0.61\% across all three budget constraints, with perfect prediction accuracy (0\% MAE) in the low-cost setting while reducing total evaluation costs by an average of 95.05\%. 
Even for the domain-specific Multifin EN benchmark, which presents high variance scenarios (38-380 training samples), our method maintains strong performance with an MAE of 1.16\%, demonstrating its robustness to data scarcity and domain shift.

\section{Is COSMOS robust with limited data access?}\label{app:limiteddata}

\begin{table}[!ht]
\centering
\resizebox{1\textwidth}{!}{
\begin{tabular}{llcccccccccc}
\toprule
\multirow{2}{*}{\textbf{Task}} & \multirow{2}{*}{\textbf{Cost Level}} & \multirow{2}{*}{\textbf{Act. Acc (\%)}} & \textbf{Pred. Acc (\%)} & \textbf{MAE$\downarrow$} & \textbf{Pred. Acc (\%)} & \textbf{MAE$\downarrow$} & \multirow{2}{*}{\textbf{Act. Cost (\$)}} & \textbf{Ours cost (\$)} & \textbf{CRR$\uparrow$ (\%)} & \textbf{Ours cost (\$)} & \textbf{CRR$\uparrow$ (\%)} \\
\cmidrule(lr){4-4}
\cmidrule(lr){5-5} \cmidrule(lr){6-6} \cmidrule(lr){7-7} \cmidrule(lr){9-9} \cmidrule(lr){10-10} \cmidrule(lr){11-11} \cmidrule(lr){12-12}
 &  &  & \textbf{(100\% data)} & \textbf{(100\% data)} & \textbf{(10\% data)} & \textbf{(10\% data)} &  & \textbf{(100\% data)} & \textbf{(100\% data)} & \textbf{(10\% data)} & \textbf{(10\% data)} \\
\midrule
\multirow{3}{*}{HellaSwag} & Low & 94.38 & 94.38 & 0.00 & 94.38 & 0.00 & 13.30 & 0.67 & 94.99 & 0.60 & 95.51 \\
 & Medium & 94.11 & 93.68 & 0.43 & 94.11 & 0.00 & 17.28 & 0.52 & 96.99 & 0.30 & 98.25 \\
 & High & 93.31 & 93.15 & 0.16 & 92.58 & 0.73 & 10.39 & 0.22 & 97.90 & 0.16 & 98.44 \\
\cmidrule(lr){1-12}
\textbf{Average} &  &  &  & \textbf{0.20} &  & \textbf{0.24} &  &  & \textbf{96.63} &  & \textbf{97.40} \\
\bottomrule
\end{tabular}
}
\caption{COSMOS maintains high prediction accuracy with limited data: Using only 10\% of training data achieves comparable MAE (0.24\%) to full data access (0.20\%) while further improving cost reduction (97.40\% vs. 96.63\%) across all budget constraints on HellaSwag.}
\label{tab:hellaswag-limited}
\end{table}

\begin{table}[!ht]
\centering
\resizebox{1\textwidth}{!}{
\begin{tabular}{llcccccccccc}
\toprule
\multirow{2}{*}{\textbf{Task}} & \multirow{2}{*}{\textbf{Cost Level}} & \multirow{2}{*}{\textbf{Act. Acc (\%)}} & \textbf{Pred. Acc (\%)} & \textbf{MAE$\downarrow$} & \textbf{Pred. Acc (\%)} & \textbf{MAE$\downarrow$} & \multirow{2}{*}{\textbf{Act. Cost (\$)}} & \textbf{Ours cost (\$)} & \textbf{CRR$\uparrow$ (\%)} & \textbf{Ours cost (\$)} & \textbf{CRR$\uparrow$ (\%)} \\
\cmidrule(lr){4-4}
\cmidrule(lr){5-5} \cmidrule(lr){6-6} \cmidrule(lr){7-7} \cmidrule(lr){9-9} \cmidrule(lr){10-10} \cmidrule(lr){11-11} \cmidrule(lr){12-12}
 &  &  & \textbf{(100\% data)} & \textbf{(100\% data)} & \textbf{(10\% data)} & \textbf{(10\% data)} &  & \textbf{(100\% data)} & \textbf{(100\% data)} & \textbf{(10\% data)} & \textbf{(10\% data)} \\
\midrule
\multirow{3}{*}{Multifin EN} & Low & 80.91 & 80.91 & 0.00 & 80.91 & 0.00 & 0.287 & 0.083 & 70.92 & 0.082 & 71.41 \\
 & Medium & 83.94 & 83.94 & 0.00 & 83.94 & 0.00 & 0.504 & 0.051 & 89.84 & 0.045 & 91.05 \\
 & High & 85.76 & 85.76 & 0.00 & 85.76 & 0.00 & 0.360 & 0.033 & 90.95 & 0.030 & 91.67 \\
\cmidrule(lr){1-12}
\textbf{Average} &  &  &  & \textbf{0.00} &  & \textbf{0.00} &  &  & \textbf{83.91} &  & \textbf{84.71} \\
\bottomrule
\end{tabular}
}
\caption{COSMOS demonstrates perfect prediction accuracy (0\% MAE) on Multifin EN in both full and limited data scenarios, while the 10\% data setting yields enhanced cost savings (84.71\% vs. 83.91\%), highlighting the method's robustness to data constraints in domain-specific tasks.}
\label{tab:multifin-limited}
\end{table}

A practical question for deployment scenarios concerns whether COSMOS requires full access to the training dataset during performance prediction. 
While our main experiments utilized complete datasets for comprehensive evaluation, this design choice was motivated by convenience rather than necessity---our lightweight linear model for QLoRA trains efficiently regardless of data volume. 
Importantly, COSMOS's data portion is a configurable parameter that can be adjusted based on resource constraints in real-world applications.

To assess the data efficiency of our approach, we conduct a comparative analysis examining prediction performance when accessing 100\% versus only 10\% of the dataset during the prediction phase. 
Following our experimental protocol from Section~\ref{sec:experiments},
we evaluate three distinct cost constraints with a performance-prioritizing score function across a configuration space covering training iterations $\in$ \{4, 5, 6, 7, 8\} and data portions $\in$ \{0.1, ..., 1.0\}, with all results averaged over three random seeds for statistical robustness. 

Tables~\ref{tab:hellaswag-limited} and~\ref{tab:multifin-limited} show that COSMOS's predictors are robust to limited data. 
For the general domain HellaSwag task, our method achieves an average MAE of just 0.24\% across three budget levels when using only 10\% of the data, with two perfect predictions (0\% MAE). 
This approach further improved cost savings from 96.63\% (with 100\% data) to 97.40\% (with 10\% data). 
For the domain-specific Multifin EN challenge, our method perfectly predicts (0\% MAE) the optimal strategies across all three cost levels while enhancing cost savings from 83.91\% to 84.71\%.

\section{A Concrete Example of Strategy Selection Based on Predicted Metrics}\label{app:concrete-example}

\begin{table}[!ht]
\centering
\resizebox{0.7\textwidth}{!}{
\begin{tabular}{llccccccc}
\toprule
\textbf{Cost Level} & \textbf{Strategy} & \textbf{Data Portion} & \textbf{Iter} & \textbf{\# shots} & \textbf{Pred. Acc} & \textbf{Act. Acc} & \textbf{Predicted Cost (\$)} & \textbf{Act. Cost (\$)} \\
\midrule
\multirow{30}{*}{Low} & \multirow{27}{*}{QLoRA} & 0.1 & 4 & - & 0.894 & 0.894 & 0.181 & 0.181 \\
 &  & 0.1 & 5 & - & 0.890 & 0.890 & 0.201 & 0.200 \\
 &  & 0.1 & 6 & - & 0.885 & 0.885 & 0.221 & 0.220 \\
 &  & 0.1 & 7 & - & 0.882 & 0.882 & 0.240 & 0.239 \\
 &  & 0.1 & 8 & - & 0.875 & 0.875 & 0.260 & 0.259 \\
 &  & 0.2 & 4 & - & 0.898 & 0.922 & 0.259 & 0.258 \\
 &  & 0.2 & 5 & - & 0.893 & 0.907 & 0.298 & 0.297 \\
 &  & 0.2 & 6 & - & 0.888 & 0.909 & 0.337 & 0.336 \\
 &  & 0.2 & 7 & - & 0.885 & 0.910 & 0.376 & 0.375 \\
 &  & 0.2 & 8 & - & 0.878 & 0.907 & 0.415 & 0.413 \\
 &  & 0.3 & 4 & - & 0.905 & 0.923 & 0.337 & 0.335 \\
 &  & 0.3 & 5 & - & 0.900 & 0.921 & 0.395 & 0.393 \\
 &  & 0.3 & 6 & - & 0.895 & 0.916 & 0.454 & 0.452 \\
 &  & 0.3 & 7 & - & 0.892 & 0.912 & 0.512 & 0.510 \\
 &  & 0.3 & 8 & - & 0.885 & 0.914 & 0.571 & 0.567 \\
 &  & 0.4 & 4 & - & 0.908 & 0.934 & 0.415 & 0.414 \\
 &  & 0.4 & 5 & - & 0.903 & 0.931 & 0.494 & 0.491 \\
 &  & 0.4 & 6 & - & 0.898 & 0.927 & 0.572 & 0.569 \\
 &  & 0.4 & 7 & - & 0.895 & 0.927 & 0.650 & 0.647 \\
 &  & 0.4 & 8 & - & 0.888 & 0.919 & 0.728 & 0.724 \\
 &  & 0.5 & 4 & - & 0.910 & 0.930 & 0.494 & 0.477 \\
 &  & 0.5 & 5 & - & 0.906 & 0.933 & 0.592 & 0.570 \\
 &  & 0.5 & 6 & - & 0.901 & 0.928 & 0.690 & 0.664 \\
 &  & 0.6 & 4 & - & 0.915 & 0.940 & 0.573 & 0.570 \\
 &  & 0.6 & 5 & - & 0.911 & 0.930 & 0.690 & 0.687 \\
 &  & 0.7 & 4 & - & 0.921 & 0.937 & 0.651 & 0.648 \\
 &  & \textbf{0.8} & \textbf{4} & \textbf{-} & \textbf{0.921} & \textbf{0.944} & \textbf{0.728} & \textbf{0.725} \\
 \cmidrule(l{2pt}r{2pt}){2-9}
 & \multirow{3}{*}{ICL} & - & - & 1 & 0.526 & 0.526 & 0.177 & 0.219 \\
 &  & - & - & 2 & 0.591 & 0.627 & 0.262 & 0.326 \\
 &  & - & - & 4 & 0.617 & 0.604 & 0.432 & 0.538 \\
\bottomrule
\end{tabular}
}
\caption{COSMOS accurately predicts that QLoRA finetuning with 0.8 portion of data for 4 iterations yields optimal performance within the low cost budget. This strategy achieves the highest predicted accuracy (0.921) among all 30 available strategies, and validation confirms it indeed delivers the best actual performance (0.944). COSMOS predicts this strategy will cost \$0.728, closely matching the actual cost of \$0.725.}
\label{tab:hellaswag-breakdowns-low}
\end{table}

\begin{table}[!ht]
\centering
\resizebox{0.7\textwidth}{!}{
\begin{tabular}{llccccccc}
\toprule
\textbf{Cost Level} & \textbf{Strategy} & \textbf{Data Portion} & \textbf{Iter} & \textbf{\# shots} & \textbf{Pred. Acc} & \textbf{Act. Acc} & \textbf{Predicted Cost (\$)} & \textbf{Act. Cost (\$)} \\
\midrule
\multirow{18}{*}{Medium} & \multirow{17}{*}{QLoRA} & 0.5 & 7 & - & 0.898 & 0.926 & 0.787 & 0.758 \\
 &  & 0.5 & 8 & - & 0.891 & 0.919 & 0.885 & 0.851 \\
 &  & 0.6 & 6 & - & 0.906 & 0.933 & 0.807 & 0.803 \\
 &  & 0.6 & 7 & - & 0.903 & 0.926 & 0.925 & 0.920 \\
 &  & 0.6 & 8 & - & 0.896 & 0.922 & 1.042 & 1.037 \\
 &  & 0.7 & 5 & - & 0.917 & 0.933 & 0.788 & 0.784 \\
 &  & 0.7 & 6 & - & 0.911 & 0.928 & 0.924 & 0.920 \\
 &  & 0.7 & 7 & - & 0.908 & 0.929 & 1.061 & 1.056 \\
 &  & 0.7 & 8 & - & 0.901 & 0.926 & 1.198 & 1.192 \\
 &  & 0.8 & 5 & - & 0.917 & 0.936 & 0.884 & 0.880 \\
 &  & 0.8 & 6 & - & 0.912 & 0.934 & 1.041 & 1.036 \\
 &  & 0.8 & 7 & - & 0.909 & 0.930 & 1.197 & 1.191 \\
 &  & 0.9 & 4 & - & 0.925 & 0.941 & 0.807 & 0.803 \\
 &  & 0.9 & 5 & - & 0.921 & 0.936 & 0.983 & 0.978 \\
 &  & 0.9 & 6 & - & 0.915 & 0.929 & 1.158 & 1.153 \\
 &  & 1.0 & 4 & - & 0.931 & 0.937 & 0.886 & 0.881 \\
 &  & 1.0 & 5 & - & 0.928 & 0.939 & 1.081 & 1.076 \\
\cmidrule(l{2pt}r{2pt}){2-9}
 & ICL & - & - & 8 & 0.620 & 0.620 & 0.772 & 0.965 \\
\bottomrule
\end{tabular}
}
\caption{Predicted performance and cost given by COSMOS and actual performance and cost corresponding to each strategy within the medium cost level.}
\label{tab:cosmos-prediction-medium}
\end{table}

\begin{table}[!ht]
\centering
\resizebox{0.7\textwidth}{!}{
\begin{tabular}{llccccccc}
\toprule
\textbf{Cost Level} & \textbf{Strategy} & \textbf{Data Portion} & \textbf{Iter} & \textbf{\# shots} & \textbf{Pred. Acc} & \textbf{Act. Acc} & \textbf{Predicted Cost (\$)} & \textbf{Act. Cost (\$)} \\
\midrule
\multirow{7}{*}{High} & \multirow{6}{*}{QLoRA} & 0.8 & 8 & - & 0.902 & 0.927 & 1.353 & 1.346 \\
 &  & 0.9 & 7 & - & 0.913 & 0.926 & 1.334 & 1.327 \\
 &  & 0.9 & 8 & - & 0.906 & 0.926 & 1.510 & 1.502 \\
 &  & 1.0 & 6 & - & 0.920 & 0.931 & 1.277 & 1.270 \\
 &  & 1.0 & 7 & - & 0.917 & 0.933 & 1.472 & 1.465 \\
 &  & 1.0 & 8 & - & 0.910 & 0.929 & 1.668 & 1.659 \\
\cmidrule(l{2pt}r{2pt}){2-9}
 & ICL & - & - & 16 & 0.620 & 0.611 & 1.452 & 1.815 \\
\bottomrule
\end{tabular}
}
\caption{Predicted performance and cost given by COSMOS and actual performance and cost corresponding to each strategy within the high cost level.}
\label{tab:cosmos-prediction-high}
\end{table}

In Tables~\ref{tab:hellaswag-breakdowns-low},~\ref{tab:cosmos-prediction-medium} and~\ref{tab:cosmos-prediction-high}, we provide a concrete example illustrating how practitioners can select the optimal strategy based on COSMOS's predicted metrics. 
Using the HellaSwag dataset across three cost regimes with a performance-prioritizing function (Same setting in Section~\ref{sec5.1-overall-perf}), we demonstrate the decision-making process. 
In the low-cost regime (30 candidate strategies), our predicted values identify QLoRA with 0.8 data portion for 4 iterations as the optimal choice, and this is the actual optimal strategy. 
For medium-cost scenarios (18 candidates), COSMOS guides selection toward 1.0 data portion with 4 iterations (0.937 accuracy), which closely approximates the ground-truth optimal strategy of 0.9 data portion with 4 iterations (0.941 accuracy, MAE: 0.004). 
Similarly, in high-cost settings (7 candidates), our predicted best strategy (1.0 data portion, 6 iterations, 0.931 accuracy) nearly matches the optimal strategy (1.0 data portion, 7 iterations, 0.933 accuracy, MAE: 0.002). 
Note that while we present these examples with our performance-prioritizing objective, practitioners can define custom trade-off functions between performance and cost as described in Section~\ref{sec:general-prob-formulation}.

\end{document}